\pdfoutput=1
\documentclass[preprint]{imsart}

\usepackage{amsthm,amsmath,amssymb}
\usepackage[authoryear]{natbib}
\usepackage{graphicx}
\usepackage{color}
\usepackage{soul}

\RequirePackage[colorlinks, citecolor=blue, urlcolor=blue]{hyperref}

\arxiv{1708.06716}

\startlocaldefs

\newcommand{\Do}{\operatorname{do}}
\newcommand{\mip}{\operatorname{MIP}}
\newcommand{\m}{\textrm{max}}
\newcommand{\C}{\mathcal{C}}
\newcommand{\A}{\mathcal{A}}
\newcommand{\transition}{v_{t-1} \prec v_t}
\DeclareMathOperator*{\argmin}{arg\,min}

\newtheorem{theorem}{Theorem}[section]

\theoremstyle{definition}
\newtheorem{definition}{Definition}[section]

\endlocaldefs

\begin{document}

\begin{frontmatter}

\thankstext{t1}{Corresponding authors: albantakis@wisc.edu, gtononi@wisc.edu}

\thankstext{T1}{We thank Matteo Mainetti for early discussions concerning the extension of IIT to actual causation. This work has been supported by the Templeton World Charities Foundation (Grant \#TWCF0067/AB41). L.A. receives funding from the Templeton World Charities Foundation (Grant \#TWCF0196).}

\title{What caused what? A quantitative account of actual causation using dynamical causal networks. \thanksref{T1}}
\runtitle{What caused what?}


\begin{aug}
\author{\fnms{Larissa} \snm{Albantakis}\corref{}},
\author{\fnms{William} \snm{Marshall}},
\author{\fnms{Erik} \snm{Hoel}}
\and
\author{\fnms{Giulio} \snm{Tononi}}
\runauthor{Albantakis et al.}
\affiliation{Department of Psychiatry, Wisconsin Institute for Sleep and Consciousness, University of Wisconsin-Madison, WI, USA\thanksmark{m1}}
\affiliation{Allen Discovery Center, Tufts University, Medford, MA, USA\thanksmark{m2}}
\end{aug}

\begin{abstract}
Actual causation is concerned with the question ``what caused what?" Consider a transition between two states within a system of interacting elements, such as an artificial neural network, or a biological brain circuit. Which combination of synapses caused the neuron to fire? Which image features caused the classifier to misinterpret the picture? Even detailed knowledge of the system's causal network, its elements, their states, connectivity, and dynamics does not automatically provide a straightforward answer to the {``what caused what?"} question. Counterfactual accounts of actual causation based on graphical models, paired with system interventions, have demonstrated initial success in addressing specific problem cases {in line with intuitive causal judgments}. Here, we start from a set of basic requirements for causation (realization, composition, information, integration, and exclusion) and develop a rigorous, quantitative account of actual causation {that is generally} applicable to discrete dynamical systems. We present a formal framework to evaluate these causal requirements that is based on system interventions and partitions, and considers all counterfactuals of a state transition. This framework is used to provide a complete causal account of the transition by identifying and quantifying the strength of all actual causes and effects linking the two {consecutive system states}. Finally, we examine several exemplary cases and paradoxes of causation and show that they can be illuminated by the proposed framework for quantifying actual causation.
\end{abstract}



\end{frontmatter}

\section{Introduction}

The nature of cause and effect has been much debated in both philosophy and the sciences. To date, there is no single widely accepted account of causation, and the various sciences focus on different aspects of the issue \citep{Illari2011}. In physics, no formal notion of causation seems even required to describe the dynamical evolution of a system by a set of mathematical equations. At most, the notion of causation is reduced to the basic requirement that causes must precede and be able to influence their effects---no further constraints are imposed as to ``what caused what". 

However, a detailed record of ``what happened" prior to a particular occurrence\footnote{{A formal definition of the term ``occurrence" is provided below in the theory section, where it denotes a system (sub)state, i.e., a set of random variables in a particular state at a particular time. This corresponds to the general usage of the term ``event" in the computer science and probability literature. The term ``occurrence" was chosen instead to avoid philosophical baggage associated with the term ``event".}} rarely provides a satisfactory explanation for \textit{why} it occurred in causal, mechanistic terms. As an example, take AlphaGo, the deep neural network that repeatedly defeated human champions in the game Go \citep{Silver2016}. Understanding why AlphaGo chose a particular move is a non-trivial problem \citep{Metz2016}, even though all its network parameters and its state evolution can be recorded in detail. Identifying ``what caused what" becomes particularly difficult in complex systems with a distributed, recurrent architecture and wide-ranging interactions such as the brain \citep{Sporns2000,Wolff2018}. 

Our interest here lies in the principled analysis of \textit{actual causation} in discrete distributed dynamical systems, such as artificial neural networks, computers made of logic gates, or cellular automata, but also biological brain circuits or gene regulatory networks. By contrast to \textit{general} (or \textit{type}) \textit{causation} which addresses the question whether the type of occurrence $A$ generally ``brings about" the type of occurrence $B$, the underlying notion of \textit{actual} (or \textit{token}) \textit{causation} addresses the question ``what caused what" given a specific occurrence $A$ followed by a specific occurrence $B$. For example, {what part of the board's particular pattern caused AlphaGo to decide on this particular move?\footnote{A question regarding general causation in the context of AlphaGo would be, e.g., whether an opponents ``moyo'' (framework for establishing territory) typically causes AlphaGo to perform an invasion.} As highlighted by the AlphaGo example,} even with detailed knowledge of all circumstances, the prior system state, and the outcome, there often is no straightforward answer to the ``what caused what" question. This has also been demonstrated by a long list of controversial examples conceived, analyzed, and debated primarily by philosophers (e.g., \cite{Lewis1986, Pearl2000, Woodward2003,Hitchcock2007, Paul2013, Weslake2015-WESAPT, Halpern2016}).

During the last decades, a number of attempts to operationalize the notion of causation and to give it a formal description have been developed, most notably in computer science, probability theory, statistics \citep{Good1961, Suppes1970,Spirtes1993, Pearl1988, Pearl2000}, the law \citep{Wright1985}, and neuroscience, (\textit{e.g.}, \cite{Tononi1999}). 
Graphical methods paired with system interventions \citep{Pearl2000} have proven especially valuable for developing causal explanations. Given a causal network that represents how the state of each variable depends on other system variables via a ``structural equation" \citep{Pearl2000}, it is possible to evaluate the effects of interventions imposed from outside the network by setting certain variables to a specific value. This operation has been formalized by Pearl, who introduced the ``do-operator", $\Do(X=x)$, which signifies that a subset of system variables $X$ has been actively set into state $x$ rather than being passively observed in this state \citep{Pearl2000}. Because statistical dependence does not imply causal dependence, the conditional probability of occurrence $B$ after observing occurrence $A$, $p(B \mid A)$ may differ from the probability of occurrence $B$ after enforcing $A$, $p(B \mid \Do(A))$. Causal networks are a specific subset of ``Bayesian" networks that explicitly represent \textit{causal} dependencies consistent with interventional probabilities.

The causal networks approach has also been applied to the case of \textit{actual causation} \citep{Pearl2000,Hitchcock2001, Woodward2003, Halpern2005, Weslake2015-WESAPT, Halpern2015}.
There, system interventions can be used to evaluate whether and to what extent an occurrence was necessary or sufficient for a subsequent occurrence by assessing counterfactuals---alternative occurrences ``counter to fact''\footnote{Note that counterfactuals here strictly refer to possible states within the system's state space other than the actual one and not to abstract notions such as other ``possible worlds'' as in \citep{Lewis1973}, (see also \citep{Pearl2000} Chapter 7).} \citep{Lewis1973, Pearl2000, Woodward2004}---{within a given causal model. The objective is to define ``what it means for $A$ to be a cause of $B$ \textit{in model $M$}" \citep{Halpern2016}.} 
While promising results have been obtained in specific cases, no single proposal to date has characterized actual causation in a universally satisfying manner \citep{Paul2013, Halpern2016}. {One concern about existing measures of actual causation is the incremental manner in which they progress; a definition is proposed that satisfies existing examples in the literature, until a new problematic example is discovered, at which point the definition is updated to address the new example \citep{Weslake2015-WESAPT, Beckers2018}. While valuable, the problem with such an approach is that one cannot be confident in applying the framework beyond the scope of examples already tested. For example, while the methods are well explored in simple binary examples, there is less evidence that the methods conform with intuition when we consider the much larger space of non-binary examples (see \ref{S1}). This is especially critical when moving beyond intuitive toy examples to scientific problems where intuition is lacking, such as understanding actual causation in biological or artificial neural networks.}

Our goal is to provide a robust framework for assessing actual causation that is based on general causal principles, and can thus be expected to naturally extend beyond simple, binary, and deterministic example cases.
Below we present a formal account of actual causation that is generally applicable to discrete Markovian dynamical systems that are constituted of interacting elements (Fig. \ref{fig1}). 
The proposed framework is based on five causal principles identified in the context of integrated information theory (IIT)---namely existence (here: realization), composition, information, integration, and exclusion \citep{Oizumi2014, Albantakis2015}). Originally developed as a theory of consciousness \citep{Tononi2015, Tononi2016}, IIT provides the tools to characterize \textit{potential causation}---the causal constraints exerted by a mechanism in a given state.

In particular, our objective is to provide a complete, quantitative causal account of ``what caused what" within a transition between consecutive system states. Our approach differs from previous accounts of actual causation in what constitutes a complete causal account. Unlike most accounts of actual causation (\textit{e.g.}, \cite{Pearl2000, Paul2013, Halpern2016}), but see \citep{Chajewska1997}, causal links within a transition are considered from the perspective of \textit{both} causes and effects. Additionally, we not only evaluate actual causes and effects of individual variables, but also actual causes and effects of high-order occurrences comprising multiple variables. While some existing accounts of actual causation include the notion of being ``part of a cause" \citep{Halpern2015, Halpern2016}, the possibility of multi-variate causes and effects is rarely addressed, or even outright excluded \citep{Weslake2015-WESAPT}. 

Despite the differences in what constitutes a complete causal account, our approach remains compatible with the traditional view of actual causation, which considers only actual causes of individual variables (no high-order causation, and no actual effects). In this context, the main difference between our proposed framework and existing ``contingency'' based definitions is that we simultaneously consider \textit{all} counterfactual states of the transition, rather than a single contingency (e.g., \cite{Hitchcock2001,Yablo2002,Woodward2003,Halpern2005,Hall2007,Halpern2015,Weslake2015-WESAPT}, see \ref{S1} for a detailed comparison). This allows us to express the causal analysis in probabilistic, informational terms \citep{Ay2008, Korb2011, Janzing2013, Oizumi2014}{, which has the additional benefit that our framework naturally extends from deterministic to probabilistic causal networks, and also from binary to multi-valued variables. Finally, it allows us to quantify the strength of all causal links between occurrences and their causes and effects within the transition.}

In the following, we will first formally describe the proposed causal framework of actual causation. We then demonstrate its utility on a set of examples, which illustrate the benefits of characterizing both causes and effects, the fact that causation can be compositional, and the importance of identifying irreducible causes and effects for obtaining a complete causal account. Finally, we illustrate several prominent paradoxical cases from the actual causation literature, including overdetermination and prevention, {as well as a toy-model of an image classifier based on an artificial neural network}.

\section{Theory}

\begin{figure}
\includegraphics[width = 4.5 in]{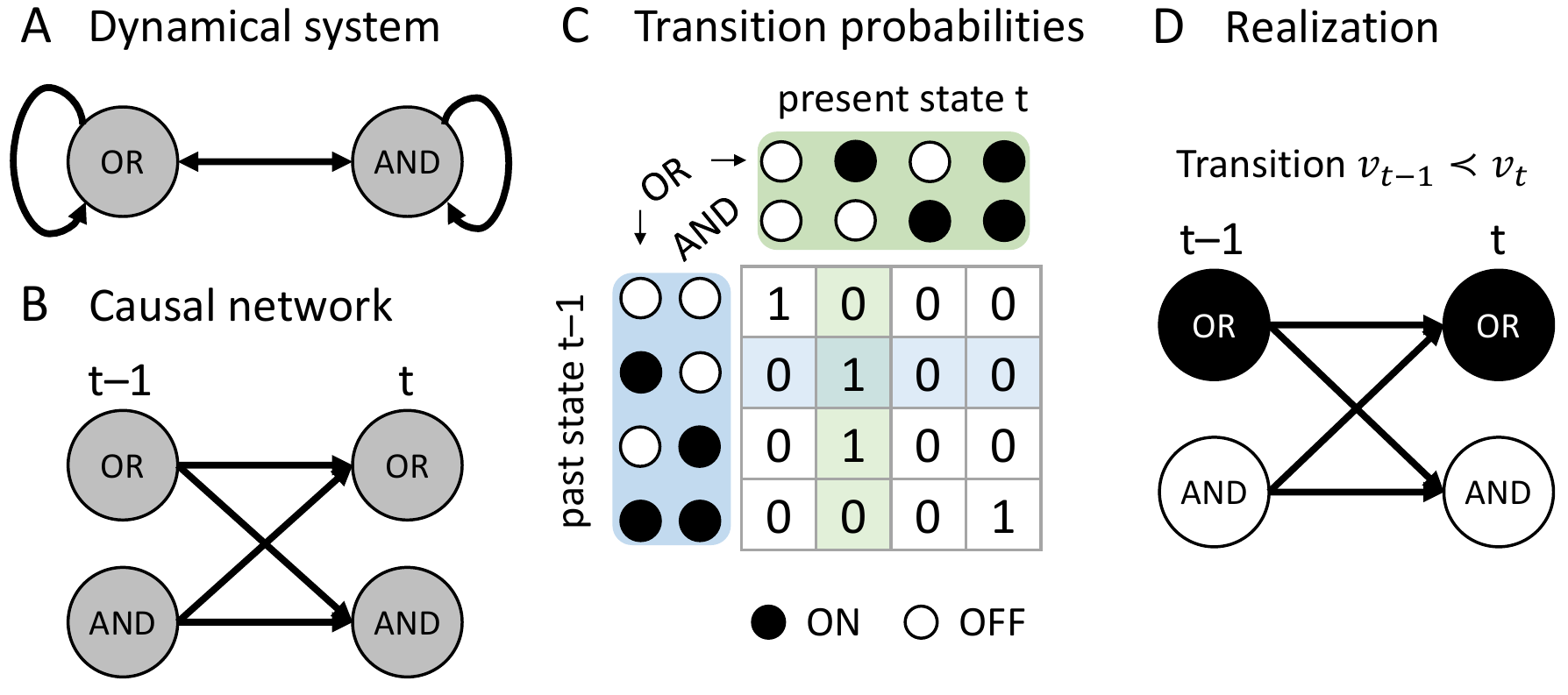}
\caption{\textbf{Realization: dynamical causal network and transition.} (A) A discrete dynamical system constituted of 2 interacting elements: an OR- and AND-logic gate, which are updated synchronously at every time step according to their input-output functions. Arrows denote connections between the elements. (B) The same system can be represented as a dynamical causal network over consecutive time steps. (C) The system described by its entire set of transition probabilities. Since this particular system is deterministic all transitions have a probability of either $p = 0$ or $p = 1$. (D) A realization of a system transient over two time steps, consistent with the system's transition probabilities: $\{\text{(OR, AND)}_{t-1} = 10\} \prec \{\text{(OR, AND)}_t = 10\}$.}
\label{fig1}
\end{figure}

Integrated information theory is concerned with the \textit{intrinsic cause-effect power} of a physical system (\textit{intrinsic existence}). The IIT formalism \citep{Oizumi2014, Tononi2015} starts from a discrete distributed dynamical system in its current state and asks how the system's elements, alone and in combination (\textit{composition}), constrain the \textit{potential} past and future states of the system (\textit{information}), and whether they do so above and beyond their parts (\textit{integration}). 
The potential causes and effects of a system subset correspond to the set of elements over which the constraints are maximally informative and integrated (\textit{exclusion}). 
In the following we aim to translate IIT's account of potential causation into a principled, quantitative framework for \textit{actual} causation that allows evaluating all actual causes and effects within a state transition of a dynamical system of interacting elements, such as a biological or artificial neural network (Fig. \ref{fig1}). For maximal generality, we will formulate our account of actual causation in the context of dynamical causal networks \citep{Ay2008, Janzing2013,Biehl2016}.

\subsection{Dynamical Causal Networks}
\label{DCN}
Our starting point is a dynamical causal network---a directed acyclic graph (DAG) $G_u = (V, E)$ with edges $E$ that indicate the causal connections among a set of nodes $V$ and a given set of background conditions (state of exogenous variables) $U = u$ (Fig. \ref{fig1}B). The nodes in $G_u$ represent a set of associated random variables (which we also denote $V$) with state space $\Omega$ and probability function $p(v | u), \, v \in \Omega$. For any node $V_i \in V$, we can define the parents of $V_i$ in $G_u$ as all nodes with an edge leading into $V_i$,
\begin{equation*}
    pa(V_i) = \{V_j \mid e_{ji} \in E\}.
\end{equation*}

A causal network $G_u$ is dynamical in the sense that we can define a partition of its nodes $V$ into $k+1$ temporally ordered ``slices'' $V = \{V_0, V_1, \dots, V_{k}\}$, starting with an initial slice without parents ($pa(V_0) = \varnothing$) and such that the parents of each successive slice are fully contained within the previous slice ($pa(V_t) \subseteq V_{t-1}, \, t = 1, \dots, k$). This definition is similar to one proposed in \cite{Ay2008}, but is stricter,  requiring that there are no within-slice causal interactions. This restriction prohibits any ``instantaneous causation" between variables (see also \cite{Pearl2000}, Section 1.5) and signifies that $G_u$ fulfills the Markov property. The parts of $V = \{V_0, V_1, \dots, V_{k}\}$ can thus be interpreted as consecutive time steps of a discrete dynamical system of interacting elements (Fig. \ref{fig1}); a particular state $V = v$ then corresponds to a realization of a system transient over $k+1$ time steps.

In a \emph{Bayesian} network, the edges of $G_u$ fully capture the dependency structure between nodes $V$. That is, for a given set of background conditions, each node is conditionally independent of every other node given its parents in $G_u$, and the probability function can be factored as 
\begin{equation*}
    p(v \mid u) = \prod_{i} p(v_i \mid pa(v_i), u), \quad v \in \Omega
\end{equation*}

For a \emph{causal} network, there is the additional requirement that the edges $E$ capture causal dependencies (rather than merely correlations) between nodes. This means that the decomposition of $p(v\mid u)$ holds even if the parent variables are actively set into their state as opposed to passively observed in that state (``Causal Markov Condition", \cite{Spirtes1993, Pearl2000}), 
\begin{equation*}
    p(v \mid u) = \prod_{i} p\big(v_i \mid do(pa(v_i), u) \big), \quad v \in \Omega.
\end{equation*}

Because we assume here that $U$ contains all relevant exogenous variables, any statistical dependencies between $V_{t-1}$ and $V_t$ are in fact causal dependencies, and cannot be explained by latent external variables (``causal sufficiency'', see \cite{Janzing2013}). Moreover, because time is explicit in $G_u$ and we assume that there is no instantaneous causation, there is no question of the direction of causal influences---it must be that the earlier variables ($V_{t-1}$) influence the later variables ($V_t$). By definition, $V_{t-1}$ contains all parents of $V_t$ for $t = 1, \dots, k$. 
Together, these assumptions imply a transition probability function for $V$ such that the nodes at time $t$ are conditionally independent given the state of the nodes at time $t-1$ (Fig. \ref{fig1}C), 
\begin{equation}
\label{eqn1b}
\begin{split}
    p_u(v_t \mid v_{t-1}) &= p(v_t \mid v_{t-1}, u) \\ 
    &= \prod_{i} p\big(v_{i, t} \mid v_{t-1}, u\big) \\
    &= \prod_{i} p\big(v_{i, t} \mid do(v_{t-1}, u)\big), \quad \forall \, (v_{t-1}, v_{t}) \in \Omega.
\end{split}
\end{equation}

To reiterate, a dynamical causal network $G_u$ describes the causal interactions among a set of nodes (the edges in $E$ describe the causal connections between the nodes in $V$) conditional on the state of exogenous variables $U$, and the transition probability function $p_u(v_t \mid v_{t-1})$ (Eqn. \ref{eqn1b}) fully captures the nature of these causal dependencies. 

In sum, we assume that $G_u$ fully and accurately describes the system of interest for a given set of background conditions. In reality, a causal network reflects assumptions about a system's elementary mechanisms. Current scientific knowledge must inform which variables to include, what their relevant states are, and how they are related mechanistically \citep{Pearl2000, Pearl2010}. Here, we are primarily interested in natural and artificial systems, such as neural networks, for which detailed information about the causal network structure and the mechanisms of individual system elements is often available, or can be obtained through exhaustive experiments\footnote{The transition probabilities can, in principle, be determined, by perturbing the system into all possible states while holding the exogenous variables fixed and observing the resulting transitions. Alternatively, the causal network can be constructed by experimentally identifying the input-output function of each element (its structural equation \citep{Pearl2000, Janzing2013}). Merely observing the system without experimental manipulation is insufficient to identify causal relationships in most situations.}. In such systems, counterfactuals can be evaluated by performing experiments or simulations that assess how the system reacts to interventions. Our objective here is to formulate a quantitative account of actual causation applicable to any predetermined, dynamical causal network, independent of practical considerations about model selection \citep{Pearl2010, Halpern2016}. Confounding issues due to incomplete knowledge, such as estimation biases of probabilities from finite sampling, or latent variables, are thus set aside for the present purposes. To what extent and under which conditions the identified actual causes and effects generalize across possible levels of description, or under incomplete knowledge, is an interesting question that we plan to address in future work (see also \cite{Rubenstein2017, Marshall2018}). 

\subsection{Occurrences and transitions}
In general, actual causation can be evaluated over multiple time steps, e.g., considering indirect causal influences. Here, however, we specifically focus on direct causes and effects without intermediary variables or time steps.\footnote{Note that our approach generalizes, in principle, to system transitions across multiple time steps by considering the transition probabilities $p_u(v_t \mid v_{t-k})$ instead of $p_u(v_t \mid v_{t-1})$ in Eqn. \ref{eqn1b}. While this practice would correctly identify counterfactual dependencies between $v_{t-k}$ and $v_{t}$, it ignores the actual states of intermediate time steps $(v_{t-k+1}, \dots, v_{t-1})$. As a consequence, the approach cannot, at present, address certain issues regarding causal transitivity across multiple paths, incomplete causal processes in probabilistic causal networks \citep{Schaffer2001}, or causal dependencies in non-Markovian systems.} For this reason, we only consider causal networks containing nodes from two consecutive time points, $V = \{V_{t-1}, V_{t}\}$ and define a \textit{transition}, denoted $\transition$, as a realization $V = v$ with $v = (v_{t-1}, v_t) \in \Omega$ (Fig. \ref{fig1}D).

Within a dynamical causal network $G_u = (V, E)$ with $V = \{V_{t-1}, V_t\}$, our objective is to determine the actual cause or actual effect of occurrences within a transition $\transition$. Formally, an \textit{occurrence} is defined to be a substate $X_{t-1} = x_{t-1} \subseteq V_{t-1} = v_{t-1}$ or $Y_t = y_t\subseteq V_{t} = v_{t}$, corresponding to a subset of elements at a particular time and in a particular state. 

\subsection{Cause and effect repertoires} 
\label{cer}
\begin{figure}
\includegraphics[width = 4.5in]{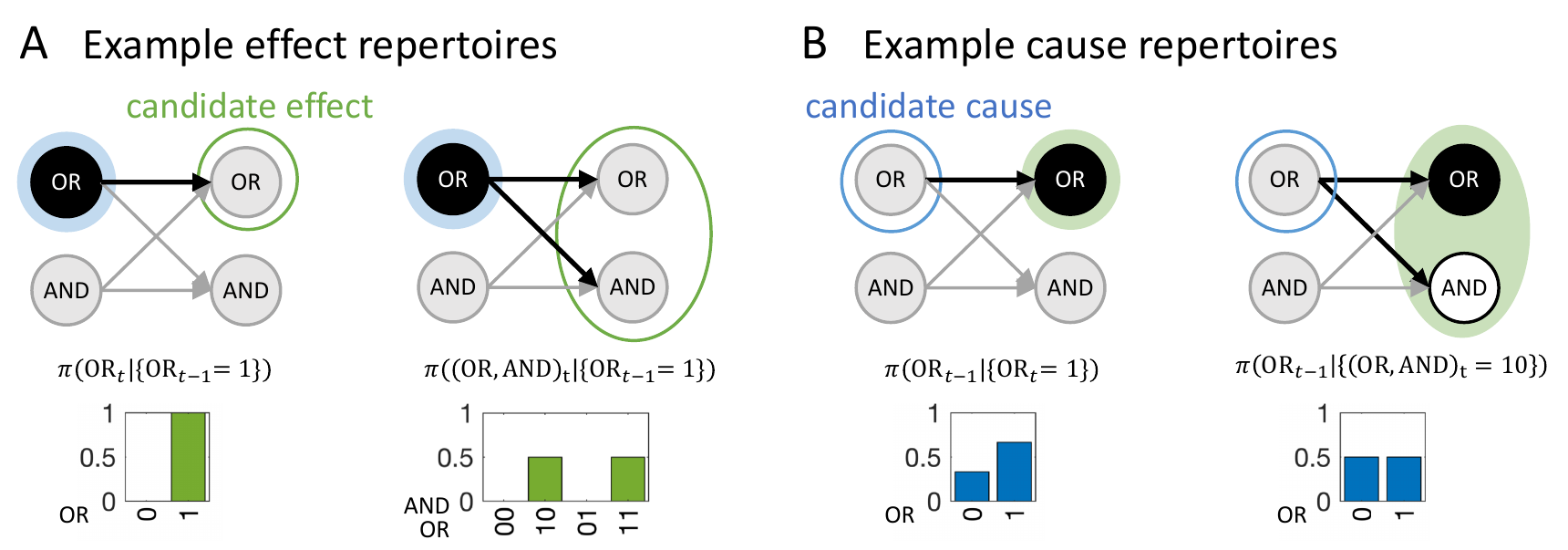}
\caption{\textbf{Assessing cause and effect repertoires.} (A) Example effect repertoires indicating how the occurrence  $\{\text{OR}_{t-1} = 1\}$ constrains the future states of $\text{OR}_t$ (left) and $(\text{OR}, \text{AND})_t$ (right) in the causal network shown in Fig. 1. (C) Example cause repertoires indicating how the occurrences $\{\text{OR}_t = 1\}$ (left) and $\{(\text{OR}, \text{AND})_t = 10\}$ (right) constrain the past states of $\text{OR}_{t-1}$. Throughout the manuscript, filled circles denote occurrences, while open circles denote candidate causes and effects. Green shading is used for $t$, blue for $t-1$. Nodes that are not included in the occurrence or candidate cause/effect are causally marginalized.}
\label{fig2}
\end{figure}

Before defining the actual cause or actual effect of an occurrence, we first introduce two definitions from IIT that are useful for characterizing the causal powers of occurrences in a causal network: cause/effect repertoires and partitioned cause/effect repertoires. In IIT, a cause (or effect) repertoire is a conditional probability distribution that describes how an occurrence (set of elements in a state) constrains the potential past (or future) states of other elements in a system \citep{Oizumi2014, Albantakis2015}, see also \citep{Tononi2015, Marshall2016} for a general mathematical definition. In the present context of a transition $\transition$, an effect repertoire specifies how an occurrence $x_{t-1}\subseteq v_{t-1}$ constrains the potential future states of a set of nodes $Y_t \subseteq V_t$. Likewise, a cause repertoire specifies how an occurrence $y_t\subseteq v_t$ constrains the potential past states of a set of nodes $X_{t-1} \subset V_{t-1}$ (Fig. \ref{fig2}).

The effect and cause repertoire can be derived from the system's transition probabilities (Eqn. \ref{eqn1b}) by conditioning on the state of the occurrence and \emph{causally marginalizing} the variables outside the occurrence $V_{t-1}\setminus X_{t-1}$ and $V_t\setminus Y_t$ (see Discussion \ref{D1} and Fig. 
\ref{fig10B}). Causal marginalization serves to remove any contributions to the repertoire from variables outside the occurrence by averaging over all their possible states. Explicitly, for a single node $Y_{i,t}$ the effect repertoire is:
\begin{equation}
\label{cmarg}
\pi(Y_{i, t} \mid x_{t-1}) = \frac{1}{|\Omega_{W}|} \sum_{w \in \Omega_{W}} p_u\left(Y_{i, t} \mid \Do\left(x_{t-1}, W = w\right)\right),
\end{equation}
where $W = V_{t-1}\setminus X_{t-1}$ with state space $\Omega_{W}$. Note that for causal marginalization, each possible state $W = w \in \Omega_{W}$ is given the same weight $|\Omega_{W}|^{-1}$ in the average. This ensures that the repertoire captures the constraints due to the occurrence \textit{per se}, and not to whatever external factors might bias the variables in $W$ to one state or another (this is discussed in more detail in Section \ref{D1}). 

The complementary cause repertoire of a singleton occurrence $y_{i, t}$, using Bayes' rule, is:
\begin{equation*}
\pi(X_{t-1}\mid y_{i, t}) = \sum_{w \in \Omega_{W}} \frac{p_u\left(y_{i, t} \mid \Do\left(X_{t-1}, W = w\right)\right)}{\sum_{z \in \Omega_{V_{t-1}}} p_u\left(y_{i, t} \mid \Do\left(V_{t-1} = z\right)\right)}.
\end{equation*}

In the general case of a multi-variate $Y_t$ (or $y_t$), the transition probability function $p_u(Y_t \mid x_{t-1})$ not only contains dependencies of $Y_t$ on $x_{t-1}$, but also correlations between variables in $Y_t$ due to common inputs from nodes in $W_{t-1} = V_{t-1}\setminus X_{t-1}$, which should not be counted as constraints due to $x_{t-1}$. To discount such correlations, we define the effect repertoire over a set of variables $Y_t$ as the product of the effect repertoires over individual nodes\footnote{In general, $\pi(Y_t \mid x_{t-1}) \neq p(Y_t \mid x_{t-1})$. However, $\pi(Y_t \mid x_{t-1})$ is equivalent to $p(Y_t \mid x_{t-1})$ in the special case that all variables $Y_{i,t} \in Y_{t}$ are conditionally independent given $x_{t-1}$ (see also \cite{Janzing2013}, Remark 1). This is the case, for example, if $X_{t-1}$ already includes all inputs (all parents) of $Y_t$, or determines $Y_t$ completely.} (see also \cite{Janzing2013}):
\begin{equation}
\label{eqn2}
\pi(Y_t \mid x_{t-1}) = \prod_i \pi(Y_{i, t} \mid x_{t-1}).
\end{equation}
In the same manner, we define the cause repertoire of a general occurrence $y_t$ over a set of variables $X_{t-1}$ as:
\begin{equation}
\label{eqn3}
\pi(X_{t-1} \mid y_t) = \frac{\prod_i \pi(X_{t-1} \mid y_{i, t})}{\sum_{x \in \Omega_{X_{t-1}}} \prod_i \pi(X_{t-1} = x \mid y_{i, t})}.
\end{equation}

We can also define \textit{unconstrained} cause and effect repertoires, a special case of cause or effect repertoires, where the occurrence that we condition on is the empty set. In this case, the repertoire describes the causal constraints on a set of the nodes due to the structure of the causal network, under maximum uncertainty about the states of variables within the network. With the convention that $\pi(\varnothing) = 1$, we can derive these unconstrained repertoires directly from the formulas for the cause and effect repertoires, Eqn \ref{eqn2} and \ref{eqn3}. The unconstrained cause repertoire simplifies to a uniform distribution, representing the fact that the causal network itself imposes no constraint on the possible states of variables in $V_{t-1}$,  
\begin{equation}
\label{eqnUCC}
    \pi(X_{t-1})= |\Omega_{X_{t-1}}|^{-1}.
\end{equation}
The unconstrained effect repertoire is shaped by the update function of each individual node $Y_{i,t} \in Y_t$ under maximum uncertainty about the state of its parents,
\begin{equation}
\label{eqnUCE}
    \pi(Y_t) = \prod_i \pi(Y_{i,t}) = \prod_i |\Omega_{W}|^{-1}\sum_{w \in \Omega_{W}} p_u(Y_{i, t} \mid \Do(W = w)),
\end{equation}
where $W = V_{t-1}\setminus X_{t-1} = V_{t-1}$, since $X_{t-1} = \varnothing$.

In summary, the effect and cause repertoires $\pi(Y_t \mid x_{t-1})$ and $\pi(X_{t-1} \mid y_t)$, respectively, are conditional probability distributions that specify the causal constraints due to an occurrence on the \textit{potential} past and future states of variables in a causal network $G_u$. The cause and effect repertoires discount constraints that are not specific to the occurrence of interest; possible constraints due to the state of variables outside of the occurrence are causally marginalized from the distribution, and constraints due to common inputs from other nodes are avoided by treating each node in the occurrence independently.

An objective of IIT is to evaluate whether the causal constraints of an occurrence on a set of nodes are ``integrated", or ``irreducible", that is, whether the individual variables in the occurrence work together to constrain the past or future states of the set of nodes  in a way that is not accounted for by the variables taken independently \citep{Balduzzi2008,Oizumi2014}. To this end, the occurrence (together with the set of nodes it constrains) is partitioned into independent parts, by rendering the connection between the parts causally ineffective \citep{Balduzzi2008, Janzing2013, Oizumi2014, Albantakis2015}. The \textit{partitioned} cause and effect repertoires describe the residual constraints under the partition. Comparing the partitioned cause and effect repertoires to the intact cause and effect repertoires reveals what is lost or changed by the partition. 

\begin{figure}
\includegraphics[width = \textwidth]{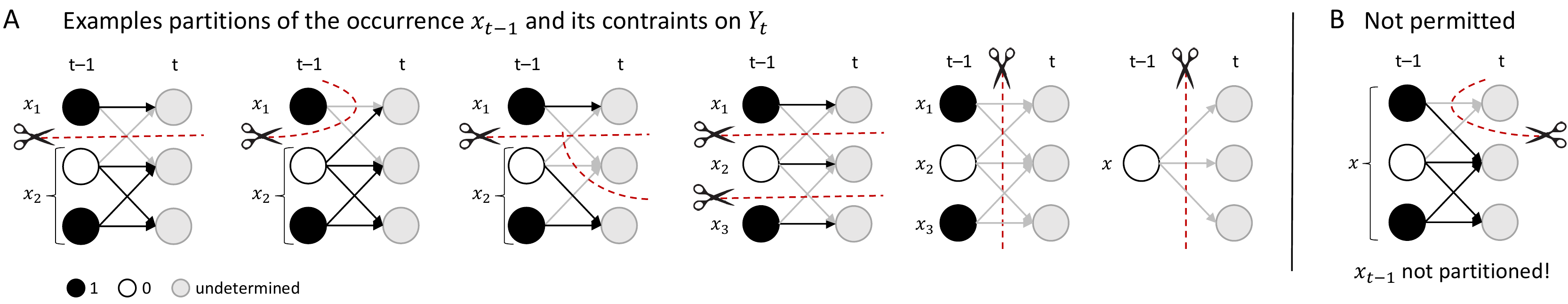}
\caption{\textbf{Partitioning the repertoire $\pi(Y_t \mid x_{t-1})$.} (A) The set of all possible partitions of an occurrence, $\Psi(x_{t-1}, Y_t)$, includes all partitions of $x_{t-1}$ into $2 \leq m \leq |x_{t-1}|$ parts according to Eqn. \ref{eqn:Pe}, as well as the special case $\psi = \{(x_{t-1}, \varnothing)\}$. Considering this special case a potential partition has the added benefit that it allows us to treat singleton occurrences and multi-variate occurrences in a common framework. (B) Except for the special case when the occurrence is completely cut from the nodes it constrains, we generally do not consider cases with $m = 1$ as partitions of the occurrence. The partition must eliminate the possibility of joint constraints of $x_{t-1}$ onto $Y_t$. The set of all partitions $\Psi(X_{t-1}, y_t)$ of a cause repertoire $\pi(X_{t-1} \mid y_{t})$ includes all partitions of $y_t$ into $2 \leq m \leq |y_t|$ parts according to Eqn. \ref{eqn:Pc} and again the special case of $\psi = \{(\varnothing, y_t)\}$ for $m=1$.}
\label{Fig2B}
\end{figure}

A partition $\psi$ of the occurrence $x_{t-1}$ (and the nodes it constrains, $Y_t$) into $m$ parts is defined as: 
\begin{equation}
\label{eqn:Pe}
    \psi(x_{t-1}, Y_t) = \{(x_{1, t-1}, Y_{1, t}), (x_{2, t-1}, Y_{2, t}), \ldots, (x_{m, t-1}, Y_{m, t})\},
\end{equation} 
such that $\{x_{j, t-1}\}_{j=1}^m$ is a partition of $x_{t-1}$ and $Y_{j, t} \subseteq Y_t$ with $Y_{j, t} \cap Y_{k, t} = \varnothing,\, j \neq k$. Note that this includes the possibility that any $Y_{j,t} = \varnothing$, which may leave a set of nodes $Y_t \setminus \bigcup_{j=1}^m Y_{j,t}$ completely unconstrained (see Fig. \ref{Fig2B} for examples and details). 

The partitioned effect repertoire of an occurrence $x_{t-1}$ over a set of nodes $Y_t$ under a partition $\psi$ is defined as:
\begin{equation}
\label{eqn6}
\pi(Y_t \mid x_{t-1})_\psi = \prod_{j = 1}^m \pi(Y_{j, t} \mid x_{j, t-1}) \times \pi\left(Y_t \setminus \bigcup_{j=1}^m Y_{j,t}\right).
\end{equation}
It is the product of the corresponding $m$ effect repertoires, multiplied by the unconstrained effect repertoire of the remaining set of nodes $Y_t \setminus \bigcup_{j=1}^m Y_{j,t}$, as these nodes are no longer constrained by any part of $x_{t-1}$ under the partition. 

In the same way, a partition $\psi$ of the occurrence $y_t$ (and the nodes it constrains $X_{t-1}$) into $m$ parts is defined as:
\begin{equation}
\label{eqn:Pc}
    \psi(X_{t-1}, y_t) = \{(X_{1, t-1}, y_{1, t}), (X_{2, t-1}, y_{2, t}), \ldots, (X_{m, t-1}, y_{m, t})\},
\end{equation} 
such that $\{y_{i, t}\}_{i=1}^m$ is a partition of $y_{t}$ and $X_{j, t-1} \subseteq X_{t-1}$ with $X_{j, t-1} \cap X_{k, t-1} = \varnothing,\, j \neq k$. The partitioned cause repertoire of an occurrence $y_{t}$ over a set of nodes $X_{t-1}$ under a partition $\psi$ is defined as:
\begin{equation}
\label{eqn7}
\pi(X_{t-1} \mid y_t)_\psi = \prod_{j = 1}^m \pi(X_{j, t-1} \mid y_{j,t}) \times \pi\left(X_{t-1} \setminus \bigcup_{j = 1}^m X_{j, t-1}\right). 
\end{equation}

\subsection{Actual causes and actual effects}
\label{ACAE}
The objective of this section is to introduce the notion of a causal account for a transition of interest $\transition$ in $G_u$ as the set of all causal links between occurrences within the transition.  There is a causal link between occurrences $x_{t-1}$ and $y_t$ if $y_t$ is the actual effect of $x_{t-1}$, or if $x_{t-1}$ is the actual cause of $y_t$. Below, we define \textit{causal link}, \textit{actual cause}, \textit{actual effect}, and \textit{causal account} following five causal principles: realization, composition, information, integration, and exclusion. 	

\textbf{Realization.} A transition $\transition$ must be consistent with the transition probability function of a dynamical causal network $G_u$,
\begin{equation*}
    p_u(v_t | v_{t-1}) > 0.
\end{equation*}
Only occurrences within a transition $\transition$ may have, or be, an actual cause or actual effect.\footnote{This requirement corresponds to the first clause (``AC1") of the Halpern and Pearl account of actual causation \citep{Halpern2005, Halpern2015}, that for $C=c$ to be an actual cause of $E=e$ both must actually happen in the first place.}
As a first example, we consider the transition $\{(\text{OR}, \text{AND})_{t-1} = 10\} \prec \{(\text{OR}, \text{AND})_{t} = 10\}$ shown in Fig. \ref{fig1}D. The transition is consistent with the conditional transition probabilities of the system shown in Fig. \ref{fig1}C. 

\begin{figure}
\includegraphics[width = \textwidth]{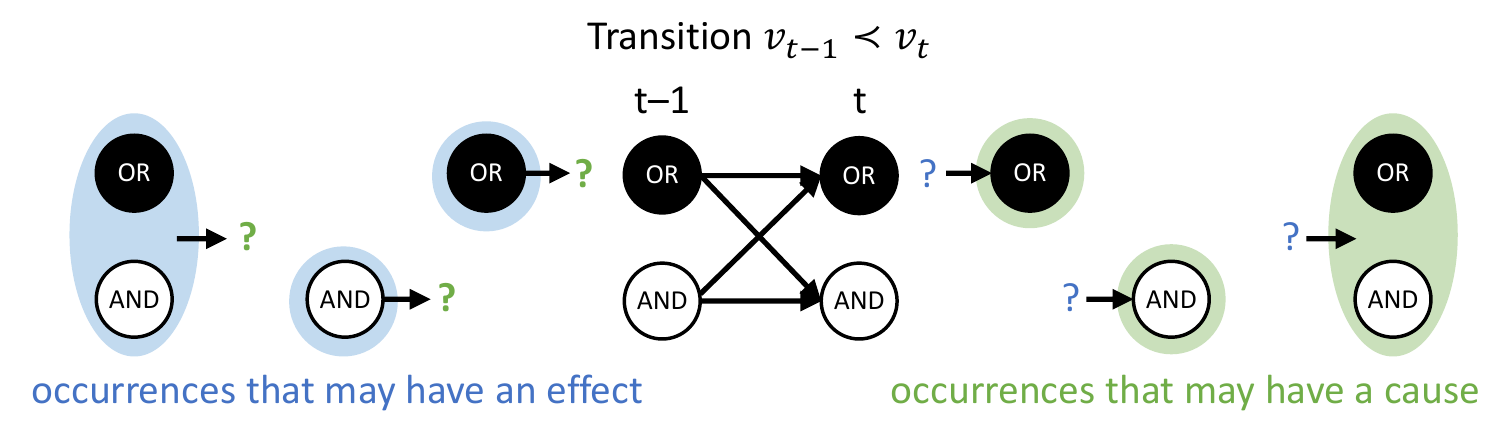}
\caption{\textbf{Considering the power set of occurrences.} All subsets $x_{t-1} \subseteq v_{t-1}$ and $y_t \subseteq v_t$ are considered as occurrences that may have an actual effect or an actual cause.}
\label{fig2C}
\end{figure}

\begin{figure}
\includegraphics[width = \textwidth]{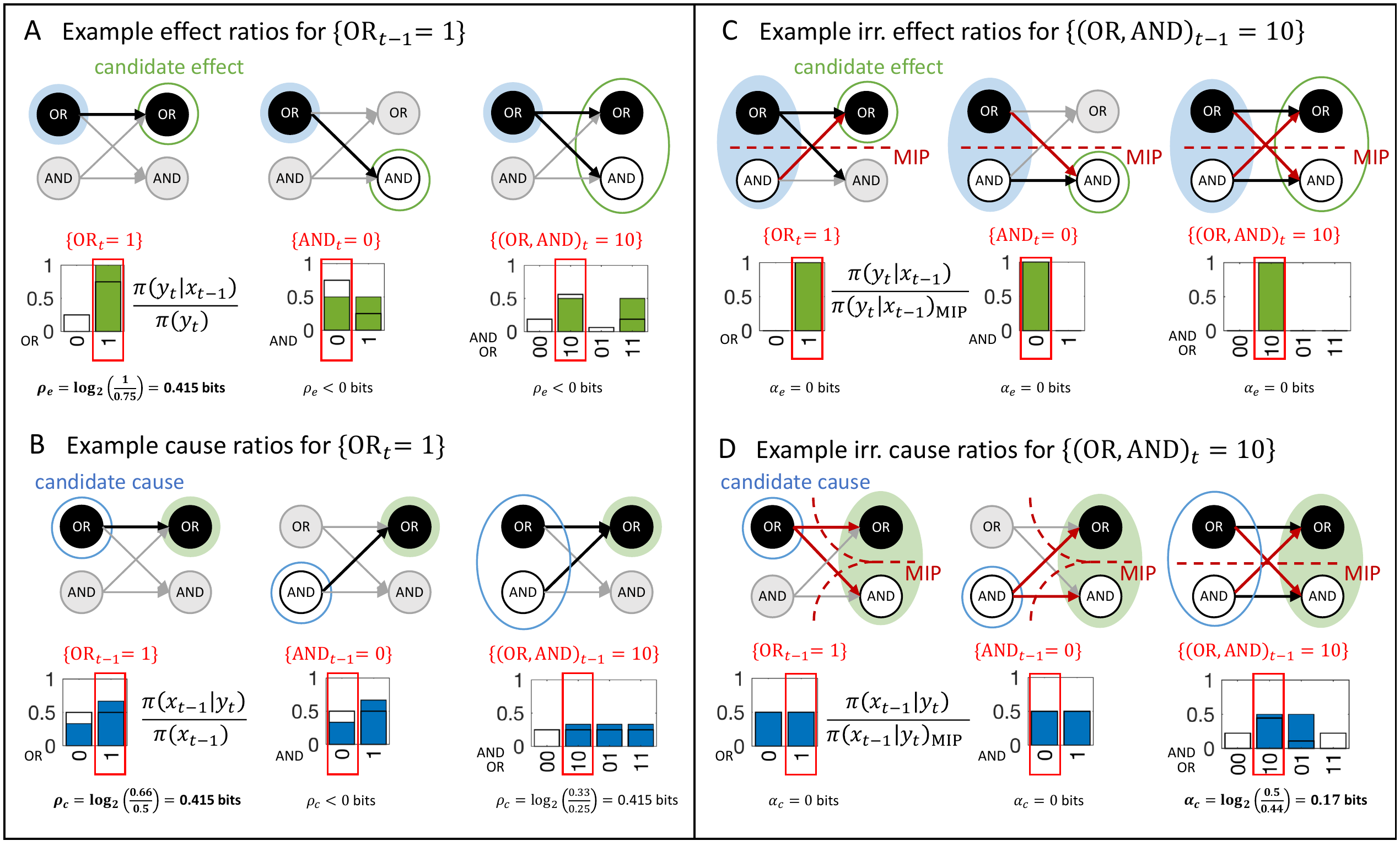}
\caption{\textbf{Assessing cause and effect ratios (information), their irreducibility (integration), and the maximum cause/effect (exclusion).} (A, B) Example effect and cause ratios. The state that actually occurred is selected from the effect or cause repertoire (green is used for effects, blue for causes). Its probability is compared to the probability of the same state when unconstrained (overlaid distributions without fill). All repertoires are based on product probabilities $\pi$ (Eqn. \ref{eqn2} and \ref{eqn3}) that discount correlations due to common inputs when variables are causally marginalized. For example, $\pi(\{\left(\text{OR, AND}\right)_{t} = 01\}) > 0$ in (A, right panel), although $p(\{\left(\text{OR, AND}\right)_{t} = 01\}) = 0$. (C, D) Irreducible effect and cause ratios. The probability of the actual state in the effect or cause repertoire is compared against its probability in the partitioned effect or cause repertoire (overlaid distributions without fill). Of all second-order occurrences shown, only $\{\left(\text{OR, AND}\right)_t = 10\}$ irreducibly constrains $\{\left(\text{OR, AND}\right)_{t-1} = 10\}$. For first-order occurrences $\alpha_{c/e} = \rho_{c/e}$ (see text). Maximum values are highlighted in bold. If, as in panel (B), a superset of a candidate cause or effect specifies the same maximum value, it is excluded by a minimality condition.}
\label{fig3}
\end{figure}

\textbf{Composition.} Occurrences and their actual causes and effects can be uni- or multi-variate. For a complete causal account of the transition $\transition$, \emph{all} causal links between occurrences $x_{t-1} \subseteq v_{t-1}$ and $y_t \subseteq v_t$ should be considered. For this reason, we evaluate every subset of $x_{t-1} \subseteq v_{t-1}$  as occurrences that may have actual effects and every subset $y_t \subseteq v_t$ as occurrences that may have actual causes (Fig. \ref{fig2C}). For a particular occurrence $x_{t-1}$, all subsets $y_t \subseteq v_t$ are considered as candidate effects (Fig. \ref{fig3}A). For a particular occurrence $y_t$, all subsets $x_{t-1} \subseteq v_{t-1}$ are considered as candidate causes (Fig. \ref{fig3}B). In what follows we refer to occurrences consisting of a single variable as ``first-order'' occurrences and to multi-variate occurrences as ``high-order'' occurrences, and, likewise, to ``first-order" and ``high-order" causes and effects.

In the example transition shown in Fig. \ref{fig2C}, $\{\text{OR}_{t-1} = 1\}$ and $\{\text{AND}_t = 0\}$ are first-order occurrences that could have an actual effect in $v_t$, and $\{\text{(OR, AND)}_{t-1} = 10\}$ is a high-order occurrence that could also have its own actual effect in $v_t$. On the other side, $\{\text{OR}_t = 1\}$, $\{\text{AND}_t = 0\}$ and $\{\text{(OR, AND)}_t = 10\}$ are occurrences (two first-order and one high-order) that could have an actual cause in $v_{t-1}$. To identify the respective actual cause (or effect) of any of these occurrences, we evaluate all possible sets $\{\text{OR} = 1\}$, $\{\text{AND} = 0\}$, and $\{(\text{OR, AND}) = 10\}$ at time $t-1$ (or $t$). Note that, in principle, we also consider the empty set, again using the convention that $\pi(\varnothing) = 1$ (see ``exclusion" below).

\textbf{Information.} An occurrence must provide information about its actual cause or effect. This means that it should increase the probability of its actual cause or effect compared to its probability if the occurrence is unspecified. To evaluate this, we compare the probability of a candidate effect $y_t$ in the effect repertoire of the occurrence $x_{t-1}$ (Eqn. \ref{eqn2}) to its corresponding probability in the unconstrained repertoire (Eqn. \ref{eqnUCE}). Specifically, we define an effect ratio $\rho_e$ for the occurrence $x_{t-1}$ and a subsequent occurrence $y_t$ (the candidate effect) as:
\begin{equation}
\label{eqn4}
\rho_e(x_{t-1}, y_t) = \log_2\left(\frac{\pi(y_t \mid x_{t-1})}{\pi(y_t)}\right),
\end{equation}
In words, the effect ratio $\rho_e$ is the relative increase in probability of an occurrence at $t$ when constrained by an occurrence at $t-1$, compared to when it is unconstrained. A positive effect ratio $\rho_e(x_{t-1}, y_t) > 0$ means that the occurrence $x_{t-1}$ makes a positive difference in bringing about $y_t$. Similarly, we compare the probability of a candidate cause $x_{t-1}$ in the cause repertoire of the occurrence $y_{t}$ (Eqn. \ref{eqn3}) to its corresponding probability in the unconstrained repertoire (Eqn. \ref{eqnUCC}). Thus, we define the cause ratio $\rho_c$ for the occurrence $y_t$ and a prior occurrence $x_{t-1}$ (the candidate cause) as:
\begin{equation}
\label{eqn5}
\rho_c(x_{t-1}, y_t) = \log_2\left(\frac{\pi(x_{t-1} \mid y_t)}{\pi(x_{t-1})}\right).
\end{equation}
In words, the cause ratio $\rho_c$ is the relative increase in probability of an occurrence at $t-1$ when constrained by an occurrence at $t$, compared to when it is unconstrained. Note that the unconstrained repertoire (Eqn. \ref{eqnUCC} and \ref{eqnUCE}) is an average over all possible states of the occurrence. The cause and effect ratios thus take all possible counterfactual states of the occurrence into account in determining the strength of constraints.

Both $\rho_e$ and $\rho_c$ can be interpreted as the number of bits of information that one occurrence specifies about the other (see \cite{Fano1961}, Chapter 2).\footnote{In an information theoretic context, the formula $\log_2\left({p(x \mid y)}/{p(x)}\right)$ is also known as the ``pointwise mutual information". While the pointwise mutual information is symmetric, the cause and effect ratios for an occurrence pair $(x_{t-1}, y_t)$ are not always identical as they are defined based on the product probabilities in Eqn. \ref{eqn2} and \ref{eqn3}.}\textsuperscript{,}\footnote{In addition to the mutual information, $\rho_{e/c}$ is also related to information theoretic divergences that measure differences in probability distributions, such as the Kullback-Leibler divergence, which would correspond to an average of $\log_2\left({p(x \mid y)}/{p(x)}\right)$ over all states $x \in \Omega_X$ weighted by $p(x \mid y)$. Here, we do not include any such weighting factor, since the transition specifies which states actually occurred.} Note that $\rho_e > 0$ is a necessary, but not sufficient condition for $y_t$ to be an actual effect of $x_{t-1}$ and $\rho_c > 0$ is a necessary, but not sufficient condition for $x_{t-1}$ to be an actual cause of $y_t$. $\rho_{c/e} = 0$ iff conditioning on the occurrence does not change the probability of a potential cause or effect, which includes the case of the empty set.

Occurrences $x_{t-1}$ that lower the probability of a subsequent occurrence $y_t$ have been termed ``preventative causes'' by some \citep{Korb2011}. Rather than counting a negative effect ratio $\rho_e(x_{t-1}, y_t) < 0$ as indicating a possible ``preventative effect'', we take the stance that such an occurrence $x_{t-1}$ has no effect on $y_t$, since it actually predicts other occurrences $Y_t = \neg y_t$ that did not happen. 
By the same logic, a negative cause ratio $\rho_c(x_{t-1}, y_t) < 0$ means that $x_{t-1}$ is no cause of $y_t$ within the transition. Nevertheless, the current framework can in principle quantify the strength of possible ``preventative'' causes and effects.

In Fig. \ref{fig3}A, for example, the occurrence $\{\text{OR}_{t-1} = 1\}$ raises the probability of $\{\text{OR}_t = 1\}$, and vice versa (Fig. \ref{fig3}B), with $\rho_e(\{\text{OR}_{t-1} = 1\}, \{\text{OR}_t = 1\}) = \rho_c(\{\text{OR}_{t} = 1\}, \{\text{OR}_{t-1} = 1\}) = 0.415$ bits. By contrast, the occurrence $\{\text{OR}_{t-1} = 1\}$ lowers the probability of occurrence $\{\text{AND}_t = 0\}$ and also of the second-order occurrence $\{\text{(OR, AND)}_t = 10\}$ compared to their unconstrained probabilities. Thus, neither $\{\text{AND}_t = 0\}$ nor $\{\text{(OR, AND)}_t = 10\}$ can be actual effects of $\{\text{OR}_{t-1} = 1\}$. Likewise, the occurrence $\{\text{OR}_t = 1\}$ lowers the probability of $\{\text{AND}_{t-1} = 0\}$, which can thus not be its actual cause.

\textbf{Integration.} A high-order occurrence must specify more information about its actual cause or effect than when its parts are considered independently. This means that the high-order occurrence must increase the probability of its actual cause or effect beyond the value specified by its parts. 

As outlined in section \ref{cer}, a partitioned cause or effect repertoire specifies the residual constraints of an occurrence after applying a partition $\psi$.
We quantify the amount of information specified by the parts of an occurrence based on partitioned cause/effect repertoires (Eqn. \ref{eqn6} and \ref{eqn7}). We define the partitioned effect ratio
\begin{equation}
\rho_e(x_{t-1}, y_t)_\psi = \log_2\left(\frac{\pi(y_t \mid x_{t-1})_\psi}{\pi(y_t)}\right),
\end{equation}
and the partitioned cause ratio
\begin{equation}
\rho_c(x_{t-1}, y_t)_\psi = \log_2\left(\frac{\pi(x_{t-1} \mid y_t)_\psi}{\pi(x_{t-1})}\right).
\end{equation}

The information a high-order occurrence specifies about its actual cause or effect is irreducible to the extent that it exceeds the information specified under \textit{any} partition $\psi$. Out of all permissible partitions $\Psi(x_{t-1}, Y_t)$ (Eqn. \ref{eqn:Pe}), or $\Psi(X_{t-1}, y_t)$ (Eqn. \ref{eqn:Pc}), the partition that least reduces an effect or cause ratio is denoted the ``minimum information partition" ($\mip$) \citep{Oizumi2014, Albantakis2015}, respectively:

\begin{equation*}
\mip = \argmin_{\psi \in \Psi(x_{t-1} , Y_t)} \left(\rho_e(x_{t-1}, y_t) - \rho_e(x_{t-1}, y_t)_\psi\right)
\end{equation*}
or
\begin{equation*}
\mip = \argmin_{\psi \in \Psi(X_{t-1}, y_t)} \left(\rho_c(x_{t-1}, y_t) - \rho_c(x_{t-1}, y_t)_\psi\right).
\end{equation*}

We can then define the irreducible effect ratio $\alpha_e$ as the difference between the intact ratio and the ratio under the $\mip$:
\begin{equation}
\label{eqn8}
\alpha_e(x_{t-1}, y_t) = \rho_e(x_{t-1}, y_t) - \rho_e(x_{t-1}, y_t)_{\mip} = \log_2\left(\frac{\pi(y_t \mid x_{t-1})}{\pi(y_t \mid x_{t-1})_{\mip}}\right),
\end{equation}
and the irreducible cause ratio $\alpha_c$ as:
\begin{equation}
\label{eqn9}
\alpha_c(x_{t-1}, y_t) = \rho_c(x_{t-1}, y_t) - \rho_c(x_{t-1}, y_t)_{\mip} = \log_2\left(\frac{\pi(x_{t-1} \mid y_t)}{\pi(x_{t-1} \mid y_t)_{\mip}}\right).
\end{equation}
For first-order occurrences $x_{i,t-1}$ or $y_{i,t-1}$ there is only one way to partition the occurrence ($\psi = \{(x_{i, t-1}, \varnothing)\}$ or $\psi = \{(y_{i, t}, \varnothing)\}$) which is necessarily the $\mip$, leading to  $\alpha_e(x_{i, t-1}, y_t) = \rho_e(x_{i, t-1}, y_t)$ or $\alpha_c(x_{t-1}, y_{i,t}) = \rho_c(x_{t-1}, y_{i,t})$, respectively.

A positive irreducible effect ratio ($\alpha_e(x_{t-1}, y_t) > 0$) signifies that the occurrence $x_{t-1}$ has an irreducible effect on $y_t$, which is necessary but not sufficient for $y_t$ to be an actual effect of $x_{t-1}$. Likewise, a positive irreducible cause ratio ($\alpha_c(x_{t-1}, y_t) > 0$) means that $y_t$ has an irreducible cause in $x_{t-1}$, which is a necessary but not sufficient condition for $x_{t-1}$ to be an actual cause of $y_t$. 

In our example transition, the occurrence $\{\text{(OR, AND)}_{t-1} = 10\}$ (Fig. \ref{fig3}C) is reducible. This is because $\{\text{OR}_{t-1} = 1\}$ is sufficient to determine that $\{\text{OR}_t = 1\}$ with probability 1.0 and $\{\text{AND}_{t-1} = 0\}$ is sufficient to determine that $\{\text{AND} = 0\}$ with probability 1.0. Thus, there is nothing to be gained by considering the two nodes together as a second-order occurrence. By contrast, the occurrence $\{\text{(OR, AND)}_t = 10\}$ determines the particular past state $\{\text{(OR, AND)}_{t-1} = 10\}$ with higher probability than the two first-order occurrences $\{\text{OR}_t = 1\}$ and $\{\text{AND}_t = 0\}$ taken separately (Fig. \ref{fig3}D, right). Thus, the second-order occurrence $\{\text{(OR, AND)}_t = 10\}$ is irreducible over the candidate cause $\{\text{(OR, AND)}_{t-1} = 10\}$ with $\alpha_c(\{\text{(OR, AND)}_{t-1} = 10\}, \{\text{(OR, AND)}_t = 10\}) = 0.17$ bits (see Discussion \ref{D4}).

\textbf{Exclusion:} An occurrence should have at most one actual cause and one actual effect (which, however, can be multivariate, that is, a high-order occurrence). In other words, only one occurrence $y_t \subseteq v_t$ can be the actual effect of an occurrence $x_{t-1}$, and only one occurrence $x_{t-1} \subseteq v_{t-1}$ can be the actual cause of an occurrence $y_t$.

It is possible that there are multiple occurrences $y_t \subseteq v_t$ over which $x_{t-1}$ is irreducible, $\alpha_e(x_{t-1}, y_t) > 0$, as well as multiple occurrences $x_{t-1} \subseteq v_{t-1}$ over which $y_t$ is irreducible, $\alpha_c(x_{t-1}, y_t) > 0$. The irreducible effect or cause ratio of an occurrence quantifies the strength of its causal constraint on a candidate effect or cause. When there are multiple candidate causes or effects for which $\alpha_{c/e}(x_{t-1}, y_t) > 0$, we select the strongest of those constraints as its actual cause or effect (that is, the one that maximizes $\alpha$). Note that adding unconstrained variables to a candidate cause (or effect) does not change the value of $\alpha$, as the occurrence still specifies the same irreducible constraints about the state of the extended candidate cause (or effect). For this reason, we include a ``minimality'' condition, such that no subset of an actual cause or effect should have the same irreducible cause or effect ratio.\footnote{The minimality condition between overlapping candidate causes or effects is related to the third clause (``AC3") in the various Halpern-Pearl accounts of actual causation \citep{Halpern2005,Halpern2015}, which states that no subset of an actual cause should also satisfy the conditions for being an actual cause. See \ref{S1}.}\textsuperscript{,}\footnote{Under uncertainty about the causal model, or other practical considerations, the minimality condition could, in principle, be replaced by a more elaborate criterion, similar to, e.g., the Akaike information criterion (AIC) that weighs increases in causal strength as measured here against the number of variables included in the candidate cause or effect.}

We define the irreducibility of an occurrence as its maximum irreducible effect (or cause) ratio over all candidate effects (or causes),
\begin{equation*}
\alpha_e^{\m}(x_{t-1}) = \max_{y_t \subseteq v_t} \alpha_e(x_{t-1}, y_t),
\end{equation*}
and
\begin{equation*}
\alpha_c^{\m}(y_t) = \max_{x_{t-1} \subseteq v_{t-1}} \alpha_c(x_{t-1}, y_t).
\end{equation*}
Considering the empty set as a possible cause or effect guarantees that the minimal value that $\alpha^{\m}$ can take is $0$. Accordingly, if $\alpha^{\m}=0$, then the occurrence is said to be reducible, and it has is no actual cause or effect. 

For the example in Fig. \ref{fig2}A, $\{\text{OR}_t = 1\}$ has two candidate causes with $\alpha_c^{\m}(\{\text{OR}_t = 1\}) = 0.415$ bits, the first-order occurrence $\{\text{OR}_{t-1} = 1\}$ and the second-order occurrence $\{\text{(OR, AND)}_{t-1} = 10\}$. In this case, $ \{\text{OR}_{t-1} = 1\}$ is the actual cause of $\{\text{OR}_t = 1\}$ by the minimality condition across overlapping candidate causes. 

The exclusion principle avoids causal overdetermination which arises from counting multiple causes or effects for a single occurrence. Note, however, that symmetries in $G_u$ can give rise to genuine indeterminism about the actual cause or effect (see Results \ref{R}). This is the case if multiple candidate causes (or effects) are maximally irreducible and they are not simple sub- or supersets of each other. Upholding the causal exclusion principle, such degenerate cases are resolved by stipulating that the \textit{one} actual cause remains undetermined between all minimal candidate causes (or effects). 

To summarize, we formally translate the five causal principles of IIT into the following requirements for actual causation:
\begin{description}
\item [Realization] There is a dynamical causal network $G_u$ and a transition $\transition$, such that $p_u(v_t | v_{t-1}) > 0$.
\item [Composition] All $x_{t-1} \subseteq v_{t-1}$ may have actual effects and be actual causes and all $y_{t} \subseteq v_t$ may have actual causes and be actual effects.
\item [Information] Occurrences must increase the probability of their causes or effects ($\rho(x_{t-1}, y_t) > 0$).
\item [Integration] Moreover, they must do so above and beyond their parts ($\alpha(x_{t-1}, y_t) > 0$).
\item [Exclusion] An occurrence has only one actual cause (or effect) and it is the occurrence that maximizes $\alpha_c$ (or $\alpha_e$). 
\end{description}

Having established the above causal principles, we now formally define the actual cause and the actual effect of an occurrence within a transition $\transition$ of the dynamical causal network $G_u$:
\begin{definition}
\label{def1}
Within a transition $\transition$ of a dynamical causal network $G_u$, the actual cause of an occurrence $y_t \subseteq v_t$ is an occurrence $x_{t-1} \subseteq v_{t-1}$ which satisfies the following conditions: 
\begin{enumerate}
    \item The irreducible cause ratio of $y_t$ over $x_{t-1}$ is maximal
    \[ \alpha_c(x_{t-1}, y_t) = \alpha^{\max}(y_t) \]
    \item No subset of $x_{t-1}$ satisfies condition (1)
    \[\alpha_c(x'_{t-1}, y_t) = \alpha^{\max}(y_t) \Rightarrow x'_{t-1} \not\subset x_{t-1} \]
\end{enumerate}
Define the set of all occurrences that satisfy the above conditions as $x^*(y_t)$. Since an occurrence can have at most one actual cause, there are three potential outcomes:
\begin{enumerate}
    \item if $x^*(y_t) = \{x_{t-1}\}$, then $x_{t-1}$ is the actual cause of $y_t$; 
    \item if $|x^*(y_t)| > 1$ then the actual cause of $y_t$ is indeterminate;
    \item if $x^*(y_t) = \{\varnothing\}$, then $y_t$ has no actual cause.
\end{enumerate}
\end{definition}

\begin{definition}
\label{def2}
Within a transition $\transition$ of a dynamical causal network $G_u$, the actual effect of an occurrence $x_{t-1} \subseteq v_{t-1}$ is an occurrence $y_t \subseteq v_t$ which satisfies the following conditions: 
\begin{enumerate}
    \item The irreducible effect ratio of $x_{t-1}$ over $y_t$ is maximal
    \[ \alpha_e(x_{t-1}, y_t) = \alpha^{\max}(x_{t-1}) \]
    \item No subset of $y_t$ satisfies condition (1)
    \[\alpha_e(x_{t-1}, y'_t) = \alpha^{\max}(x_{t-1}) \Rightarrow y'_t \not\subset y_t \]
\end{enumerate}
Define the set of all occurrences that satisfy the above conditions as $y^*(x_{t-1})$. Since an occurrence can have at most one actual effect, there are three potential outcomes:
\begin{enumerate}
    \item if $y^*(x_{t-1}) = \{y_t\}$, then $y_t$ is the actual effect of $x_{t-1}$;  
    \item if $|y^*(x_{t-1})| > 1$ then the actual effect of $x_{t-1}$ is indeterminate;
    \item if $y^*(x_{t-1}) = \{\varnothing\}$, then $x_{t-1}$ has no actual effect.
\end{enumerate}
\end{definition}

Based on Definitions \ref{def1} and \ref{def2}:
\begin{definition}
\label{def3}
Within a transition $\transition$ of a dynamical causal network $G_u$, a \textit{causal link} is an occurrence $x_{t-1} \subseteq v_{t-1}$ with $\alpha_e^{\m}(x_{t-1}) > 0$ and its actual effect $y^*(x_{t-1})$,
\begin{equation*}
x_{t-1} \rightarrow y^*(x_{t-1}),
\end{equation*}
or an occurrence $y_t \subseteq v_t$ with $\alpha_c^{\m}(y_t) >0$ and its actual cause $x^*(y_t)$,
\begin{equation*}
x^*(y_t) \leftarrow y_t 
\end{equation*}
\end{definition}

An irreducible occurrence defines a single causal link, regardless of whether the actual cause (or effect) is unique or indeterminate. When the actual cause (or effect) is unique, we sometimes refer to the actual cause (or effect) explicitly in the causal link, $x_{t-1} \leftarrow y_t$  (or $x_{t-1} \rightarrow y_t$). The \textit{strength} of a causal link is determined by its $\alpha_e^{\m}$ or $\alpha_c^{\m}$ value. Reducible occurrences ($\alpha^{\m} = 0$) cannot form a causal link.  

\begin{definition}
\label{def4}
For a transition $\transition$ of a dynamical causal network $G_u$, the causal account $\C(\transition)$ is the set of all causal links $x_{t-1} \rightarrow y^*(x_{t-1})$ and $x^*(y_t) \leftarrow y_t$ within the transition. 
\end{definition}

Under this definition, all actual causes and actual effects contribute to the causal account $\C(\transition)$. Notably, the fact that there is a causal link $x_{t-1} \rightarrow y_t$ does not necessarily imply that the reverse causal link $x_{t-1}\leftarrow y_t$ is also present, and vice versa. In other words, just because $y_t$ is the actual effect of $x_{t-1}$, the occurrence $x_{t-1}$ does not have to be the actual cause of $y_t$. It is therefore not redundant to include both directions in $\C(\transition)$, as illustrated by examples of overdetermination and prevention in the Results section (see also Discussion \ref{D2}).

\begin{figure}
\includegraphics[width = 3.0in]{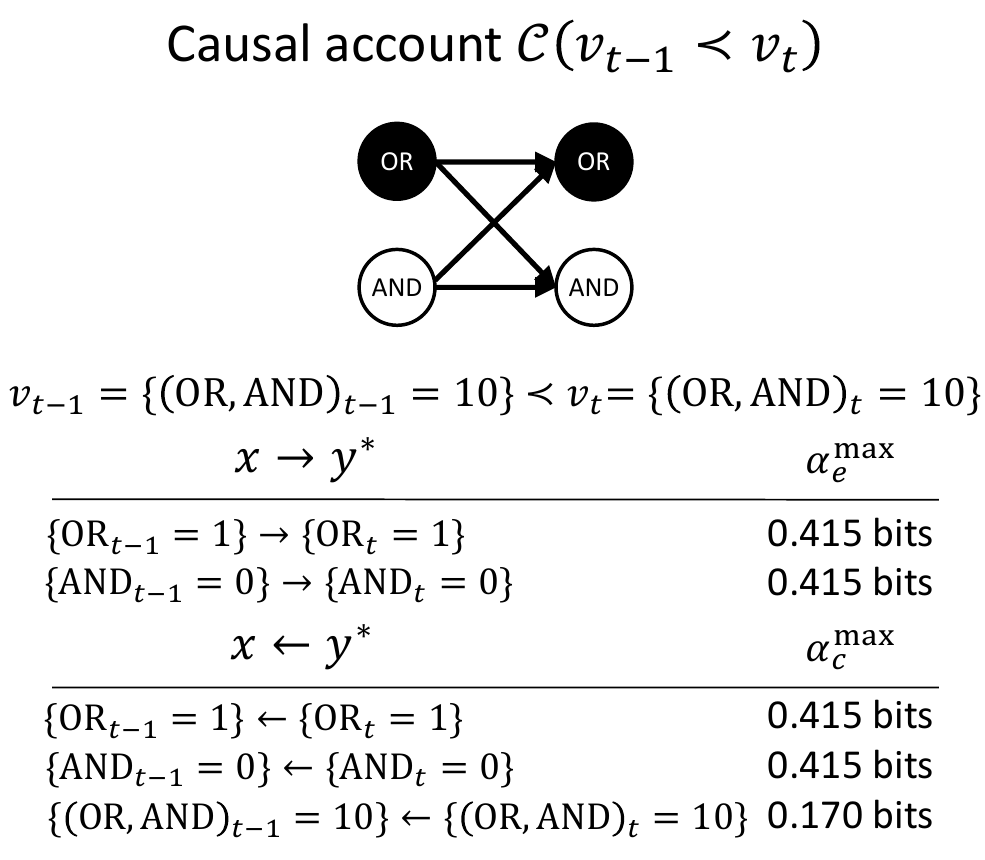}
\caption{\textbf{Causal Account.} There are two first-order occurrences with actual effects and actual causes. In addition, the second-order occurrence $\{\text{(OR, AND)}_t = 10\}$ has an actual cause $\{\text{(OR, AND)}_{t-1} = 10\}$.}
\label{fig4}
\end{figure}

Fig. \ref{fig4} shows the entire causal account of our example transition. Intuitively, in this simple example, $\{\text{OR}_{t-1} = 1\}$ has the actual effect $\{\text{OR}_t = 1\}$ and is also the actual cause of $\{\text{OR}_t = 1\}$, and the same for $\{\text{AND}_{t-1} = 0\}$ and  $\{\text{AND} = 0\}$. Nevertheless, there is also a causal link between the second-order occurrence $\{\text{(OR, AND)}_t = 10\}$ and its actual cause $\{\text{(OR, AND)}_{t-1} = 10\}$, which is irreducible to its parts, as shown in Fig. \ref{fig3}D (right). However, there is no complementary link from $\{\text{(OR, AND)}_t = 10\}$ to  $\{\text{(OR, AND)}_{t-1} = 10\}$, as it is reducible (Fig. \ref{fig3}C, right). The causal account shown in Fig. \ref{fig4} provides a complete causal explanation for ``what happened'' and ``what caused what'' in the transition $\{\text{(OR, AND)}_{t-1} = 10\} \prec \{\text{(OR, AND)}_t = 10\}$.

Similar to the notion of system-level integration in IIT \citep{Oizumi2014, Albantakis2015}, the principle of integration can also be applied to the causal account as a whole, not only to individual causal links (see \ref{S2}). In this way it is possible to evaluate to what extent the transition $\transition$ is irreducible to its parts, which is quantified by $\A(\transition)$.

In summary, the measures defined in this section provide the means to exhaustively assess ``what caused what'' in a transition $\transition$, and to evaluate the strength of specific causal links of interest under a particular set of background conditions, $U = u$. 

Software to analyze transitions in dynamical causal networks with binary variables is freely available within the ``PyPhi" toolbox for integrated information theory \citep{Mayner2018} at \url{https://github.com/wmayner/pyphi}, including documentation at \url{https://pyphi.readthedocs.io/en/stable/examples/actual_causation.html}.

\section{Results}
\label{R}

In the following, we will present a series of examples to illustrate the quantities and objects defined in the theory section and address several dilemmas taken from the literature on actual causation. For simplicity, we only cover examples including binary variables in the main text. Multi-variate examples which demonstrate that our proposed framework for actual causation naturally generalizes beyond the binary case can be found in the \ref{S1}. There, we also discuss in detail how our approach and the results below compare to counterfactual accounts of actual causation based on ``contingency conditions" \citep{Hitchcock2001, Halpern2001, Woodward2003, Halpern2005, Halpern2015, Weslake2015-WESAPT}\footnote{While indeterminism may play a fundamental role in physical causal models, the existing literature on actual causation largely focuses on deterministic problem cases. For ease of comparison, most causal networks analyzed in the following are thus deterministic, corresponding to prominent test cases.}.

\subsection{Same transition, different mechanism: disjunction, conjunction, biconditional, and prevention}

\begin{figure}
\includegraphics[width = \textwidth]{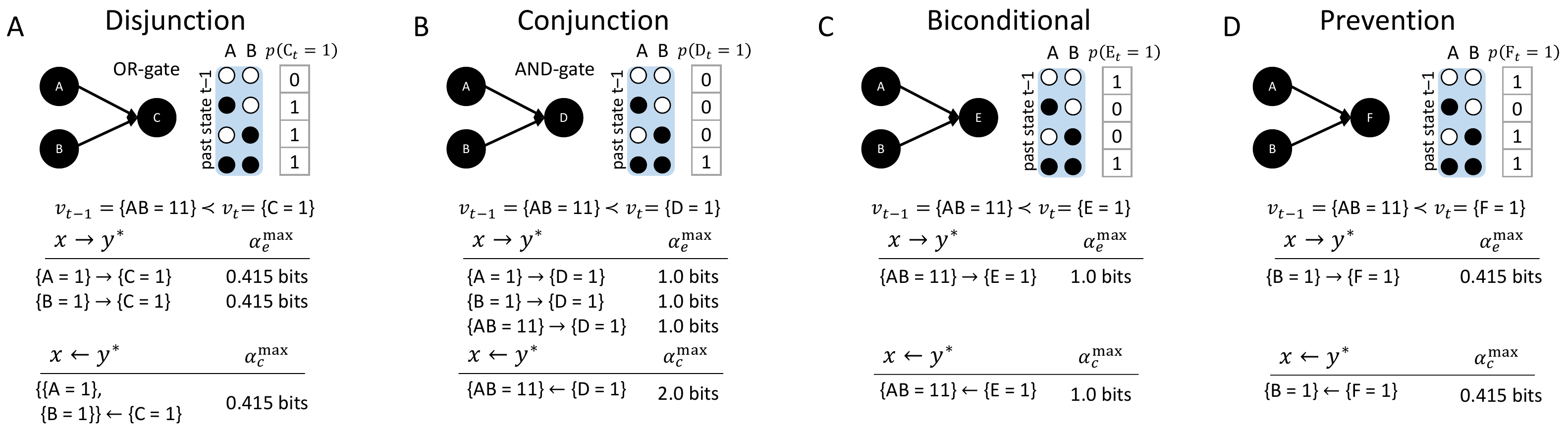}
\caption{\textbf{Four dynamically identical transitions can have different causal accounts.} Shown are the transitions (top) and their respective causal accounts (bottom).}
\label{fig5}
\end{figure}

Fig. \ref{fig5} shows 4 causal networks of different types of logic gates with two inputs each, all transitioning from the input state $v_{t-1} = \{AB = 11\}$ to the output state $v_t = \{C = 1\}$, $\{D = 1\}$, $\{E = 1\}$ or $\{F = 1\}$. From a dynamical point of view, without taking the causal structure of the mechanisms into account, the same occurrences happen in all four situations. However, analyzing the causal accounts of these transitions reveals differences in the number, type, and strength of causal links between occurrences and their actual causes or effects. 

\textbf{Disjunction:} The first example (Fig. \ref{fig5}A -- OR-gate), is a case of symmetric overdetermination (\cite{Pearl2000}, Chapter 10): each input to $C$ would have been sufficient for $\{C = 1\}$, yet both $\{A = 1\}$ and $\{B = 1\}$ occurred at $t-1$.
In this case, each of the inputs to $C$ has an actual effect, $\{A = 1\} \rightarrow \{C = 1\}$ and $\{B = 1\} \rightarrow \{C = 1\}$, as they raise the probability of $\{C = 1\}$ compared to its unconstrained probability. The high-order occurrence $\{AB = 11\}$,  however, is reducible with $\alpha_e = 0$. While both $\{A = 1\}$ and $\{B = 1\}$ have actual effects, by the causal exclusion principle, the occurrence $\{C = 1\}$ can only have one actual cause. Since both $\{A = 1\} \leftarrow \{C = 1\}$ and $\{B = 1\} \leftarrow \{C = 1\}$ have $\alpha_c = \alpha^{\m}_c = 0.415$ bits, by Definition \ref{def1}, the actual cause of $\{C = 1\}$ is either $\{A = 1\}$, or $\{B = 1\}$; which of the two inputs it is remains undetermined, since they are perfectly symmetric in this example. Note that $\{AB = 11\} \leftarrow \{C = 1\}$ also has $\alpha_c = 0.415$ bits, but $\{AB = 11\}$ is excluded from being a cause by the minimality condition.

\textbf{Conjunction:} In the second example (Fig. \ref{fig5}B -- AND-gate), both $\{A = 1\}$ and $\{B = 1\}$ are necessary for $\{D = 1\}$. In this case, each input alone has an actual effect, $\{A = 1\} \rightarrow \{C = 1\}$ and $\{B = 1\} \rightarrow \{C = 1\}$ (with higher strength than in the disjunctive case), but here also the second-order occurrence of both inputs together has an actual effect, $\{AB = 11\} \rightarrow \{D = 1\}.$ Thus, there is a composition of actual effects. Again, the occurrence $\{D = 1\}$ can only have one actual cause; here it is the second-order cause $\{AB = 11\}$, the only occurrence that satisfies the conditions in Definition \ref{def1} with $\alpha_c = \alpha_c^{\m} = 2.0$.

The two examples in Fig. \ref{fig5}A and B are often referred to as the disjunctive and conjunctive versions of the ``forest-fire'' example \citep{Halpern2005, Halpern2015, Halpern2016}, where lightning and/or a match being dropped result in a forest fire. In the case that lightning strikes and the match is dropped, $\{A = 1\}$ and $\{B = 1\}$ are typically considered two separate (first-order) causes in both the disjunctive and conjunctive version (e.g., \cite{Halpern2005}, see \ref{S1}). This result is not a valid solution within our proposed account of actual causation, as it violates the causal exclusion principle. We explicitly evaluate the high-order occurrence $\{AB = 11\}$ as a candidate cause, in addition to $\{A = 1\}$ and $\{B = 1\}$. In line with the distinct logic structure of the two examples, we identify the high-order occurrence $\{AB = 11\}$ as the actual cause of $\{D = 1\}$ in the conjunctive case, while we identify either $\{A = 1\}$ or $\{B = 1\}$ as the actual cause of $\{C = 1\}$ in the disjunctive case, but not both. By separating actual causes from actual effects, acknowledging causal composition, and respecting the causal exclusion principle, our proposed causal analysis can illuminate and distinguish all situations displayed in Fig. \ref{fig5}.

\textbf{Biconditional}: The significance of high-order occurrences is further emphasized by the third example (Fig. \ref{fig5}C), where $E$ is a ``logical biconditional'' (an XNOR) of its two inputs. In this case, the individual occurrences $\{A = 1\}$ and $\{B = 1\}$ by themselves make no difference in bringing about $\{E = 1\}$; their effect ratios are zero. For this reason, they cannot have actual effects and cannot be actual causes. Only the second-order occurrence $\{AB = 11\}$ specifies $\{E = 1\}$, which is its actual effect $\{AB = 11\} \rightarrow \{E = 1\}$. Likewise, $\{E = 1\}$ only specifies the second-order occurrence $\{AB = 11\}$, which is its actual cause $\{AB = 11\} \leftarrow \{E = 1\}$, but not its parts taken separately. Note that the causal strength in this example is lower than in the case of the AND-gate, since, everything else being equal, $\{D = 1\}$ is mechanistically a less likely output than $\{E = 1\}$.

\textbf{Prevention:} In the final example, Fig. \ref{fig5}D, all input states but $\{AB = 10\}$ lead to $\{F = 1\}$. Here, $\{B = 1\} \rightarrow \{F = 1\}$ and $\{B = 1\} \leftarrow \{F = 1\}$, whereas $\{A = 1\}$ does not have an actual effect and is not an actual cause. For this reason, the transition $\transition$ is reducible ($\A(\transition) = 0$, \ref{S2}), since $A$ could be partitioned away without loss. This example can be seen as a case of prevention: $\{B = 1\}$ causes $\{F = 1\}$, which prevents any effect of $\{A = 1\}$. In a popular narrative accompanying this example, $\{A = 1\}$ is an assassin putting poison in the King's tea, while a bodyguard administers an antidote $\{B = 1\}$, and the King survives $\{F = 1\}$ \citep{Halpern2016}. The bodyguard thus ``prevents'' the King's death\footnote{Note however that this causal model is equivalent to an OR-gate, as can be seen by switching the state labels of $A$ from `0' to `1' and vice versa. The discussed transition would correspond to the case of one input to the OR-gate being `1' and the other `0'. Since the OR-gate switches on (`1') in this case, the `0' input has no effect and is not a cause.}. 
Note that the causal account is state dependent: for a different transition, $A$ may have an actual effect or contribute to an actual cause: if the bodyguard does not administer the antidote ($\{B = 0\})$, whether the King survives depends on the assassin (the state of $A$). 

Taken together, the above examples demonstrate that the causal account and the causal strength of individual causal links within the account capture differences in sufficiency and necessity of the various occurrences in their respective transitions. Including both actual causes and effects moreover contributes to a mechanistic understanding of the transition, since not all occurrences at $t-1$ with actual effects end up being actual causes of occurrences at $t$.

\subsection{Linear threshold units}

A generalization of simple, linear logic gates, such as OR- and AND-gates, are binary linear threshold units (LTUs). Given $n$ equivalent inputs $V_{t-1} = \{V_{1, t-1}, V_{2, t-1}, \ldots, V_{n, t-1}\}$ to a single LTU $V_t$, $V_t$ will turn on (`1') if the number of inputs in state `1' exceeds a given threshold $k$, 
\begin{equation}
\label{LTU}
p(V_t = 1 \mid v_{t-1}) = \begin{cases} 1 & \text{if}~ \sum_{i = 1}^n v_{i, t-1} \geq k, \\ 0 & \text{if}~ \sum_{i=1}^n v_{i, t-1} < k. \end{cases}
\end{equation}
LTUs are of great interest, for example, in the field of neural networks, since they comprise one of the simplest model mechanisms for neurons, capturing the notion that a neuron fires if it received sufficient synaptic inputs. One example is a Majority-gate, which outputs `1' \textit{iff} more than half of its inputs are `1'.

\begin{figure}
\includegraphics[width = 4in]{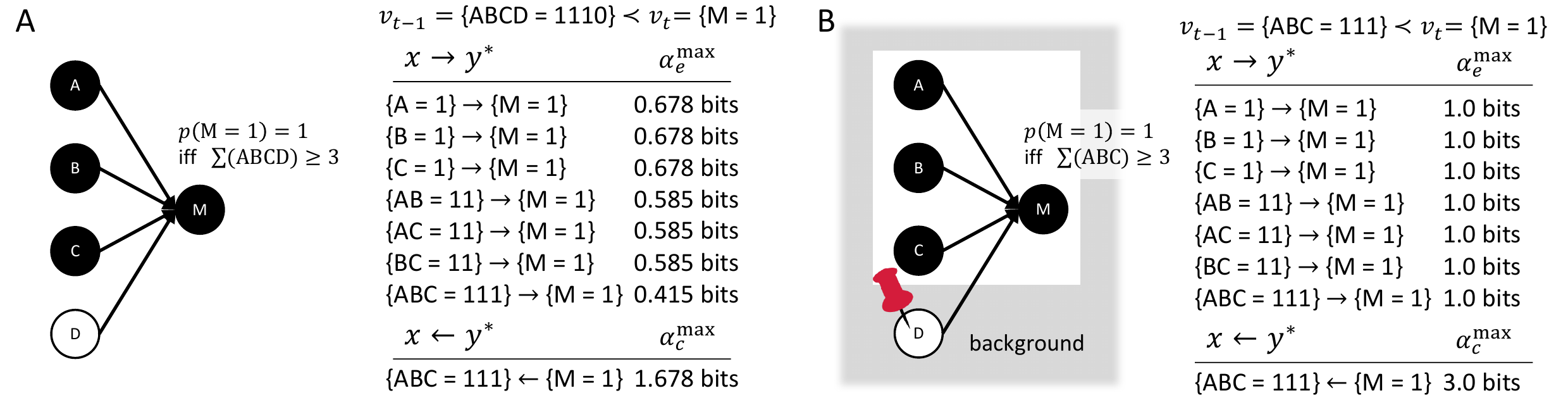}
\caption{\textbf{A linear threshold unit with four inputs and threshold $k = 3$ (Majority gate).} (A) All inputs are considered relevant variables. (B) $D = 0$ is taken as a fixed background condition (indicated by the red pin).}
\label{fig6}
\end{figure}

Fig. \ref{fig6}A displays the causal account of a Majority-gate $M$ with 4 inputs for the transition $v_{t-1} = \{ABCD = 1110\} \rightarrow v_t = \{M = 1\}$. All of the inputs in state `1', as well as their high-order occurrences, have actual effects on $\{M = 1\}$. Occurrence $\{D = 0\}$, however, does not work towards bringing about $\{M = 1\}$: it reduces the probability for $\{M = 1\}$ and thus does not contribute to any actual effects or the actual cause. As with the AND-gate in the previous section, there is a composition of actual effects in the causal account. Yet, there is only one actual cause, $\{ABC = 111\} \leftarrow \{M = 1\}$. In this case, it happens to be that the third-order occurrence $\{ABC = 111\}$ is minimally sufficient for $\{M = 1\}$---no smaller set of inputs would suffice. Note however, that the actual cause is not determined based on sufficiency, but because $\{ABC = 111\}$ is the set of nodes maximally constrained by the occurrence $\{M = 1\}$. Nevertheless, causal analysis as illustrated here will always identify a minimally sufficient set of inputs as the actual cause of an LTU $v_t = 1$, for any number of inputs $n$ and any threshold $k$. Furthermore, any occurrence of input variables $x_{t-1} \subseteq v_{t-1}$ with at most $k$ nodes, all in state `1', will be irreducible, with the LTU $v_t = 1$ as their actual effect. 

\begin{theorem}
\label{thm1}
Consider a dynamical causal network $G_u$ such that $V_t = \{Y_t\}$ is a linear threshold unit with $n$ inputs and threshold $k \leq n$, and $V_{t-1}$ is the set of $n$ inputs to $Y_t$. For a transition $v_{t-1} \prec v_{t}$, with $y_t = 1$ and $\sum v_{t-1} \geq k$, the following holds: 

\begin{enumerate}
\item  The actual cause of $\{Y_t = 1\}$ is an occurrence $\{X_{t-1} = x_{t-1}\}$ with $|x_{t-1}| = k$ and $\min(x_{t-1}) = 1$. 

\item If $\min(x_{t-1}) = 1$ and $|x_{t-1}| \leq k$ then the actual effect of $\{X_{t-1} = x_{t-1}\}$ is $\{Y_t = 1\}$; 
otherwise $\{X_{t-1} = x_{t-1}\}$ has no actual effect, it is reducible.

\end{enumerate}

\end{theorem}
Proof: See \ref{proof}.

Note that an LTU in the off (`0') state, $\{Y_t = 0\}$, has equivalent results with the role of `0' and `1' reversed, and a threshold of $n-k$. In the case of overdetermination, \textit{e.g.}, the transition $v_{t-1} = \{ABCD = 1111\} \prec v_t = \{M = 1\}$, where all inputs to the Majority-gate are `1', the actual cause will again be a subset of 3 input nodes in state `1'. However, which of the possible sets remains undetermined due to symmetry, just as in the case of the OR-gate in Fig. \ref{fig5}A.

\subsection{Distinct background conditions}
The causal network in Fig. \ref{fig6}A considers all inputs to $M$ as relevant variables. Under certain circumstance, however, we may want to consider a different set of background conditions. For example, in a voting scenario it may be a given that $D$ always votes ``no" ($D=0$). In that case we may want to analyze the causal account of the transition $v_{t-1} = \{ABC = 111\} \prec v_t = \{M=1\}$ in the alternative causal model $G_{u'}$ , where $\{D = 0\} \in \{U' = u'\}$ is treated as a background condition (Fig. \ref{fig6}B). 
Doing so results in a causal account with the same causal links but higher causal strengths. This captures the intuition that $A$, $B$, and $C$'s ``yes votes'' are more important if it is already determined that $D$ will vote ``no''. 

The difference between the causal accounts of $\transition$ in $G_u$ compared to $G_{u'}$, moreover, highlights the fact that we explicitly distinguish fixed background conditions $U = u$ from relevant variables $V$ whose counterfactual relations must be considered (see also \cite{McDermott2002}). While the background variables are fixed in their actual state $U = u$, all counterfactual states of the relevant variables $V$ are considered when evaluating the causal account of $\transition$ in $G_u$.

\subsection{Disjunction of conjunctions}
\label{R3}

Another case often considered in the actual causation literature is a disjunction of conjunctions, that is, an OR-operation over two or more AND-operations. In the general case, a disjunction of conjunctions is a variable $V_t$ that is a disjunction of $k$ conditions, each of which is a conjunction of $n_j$ input nodes $V_{t-1} = \{\{V_{i, j, t-1}\}_{i = 1}^{n_j}\}_{j=1}^{k}$, 
\begin{equation*}
p(V_t = 1 \mid v_{t-1}) = \begin{cases} 0 & \text{if} ~ \sum_{i=1}^{n_j} v_{i, j, t-1} < n_j, ~ \forall j \\ 1 & \text{otherwise} \end{cases}
\end{equation*}

\begin{figure}
\includegraphics[width = 4in]{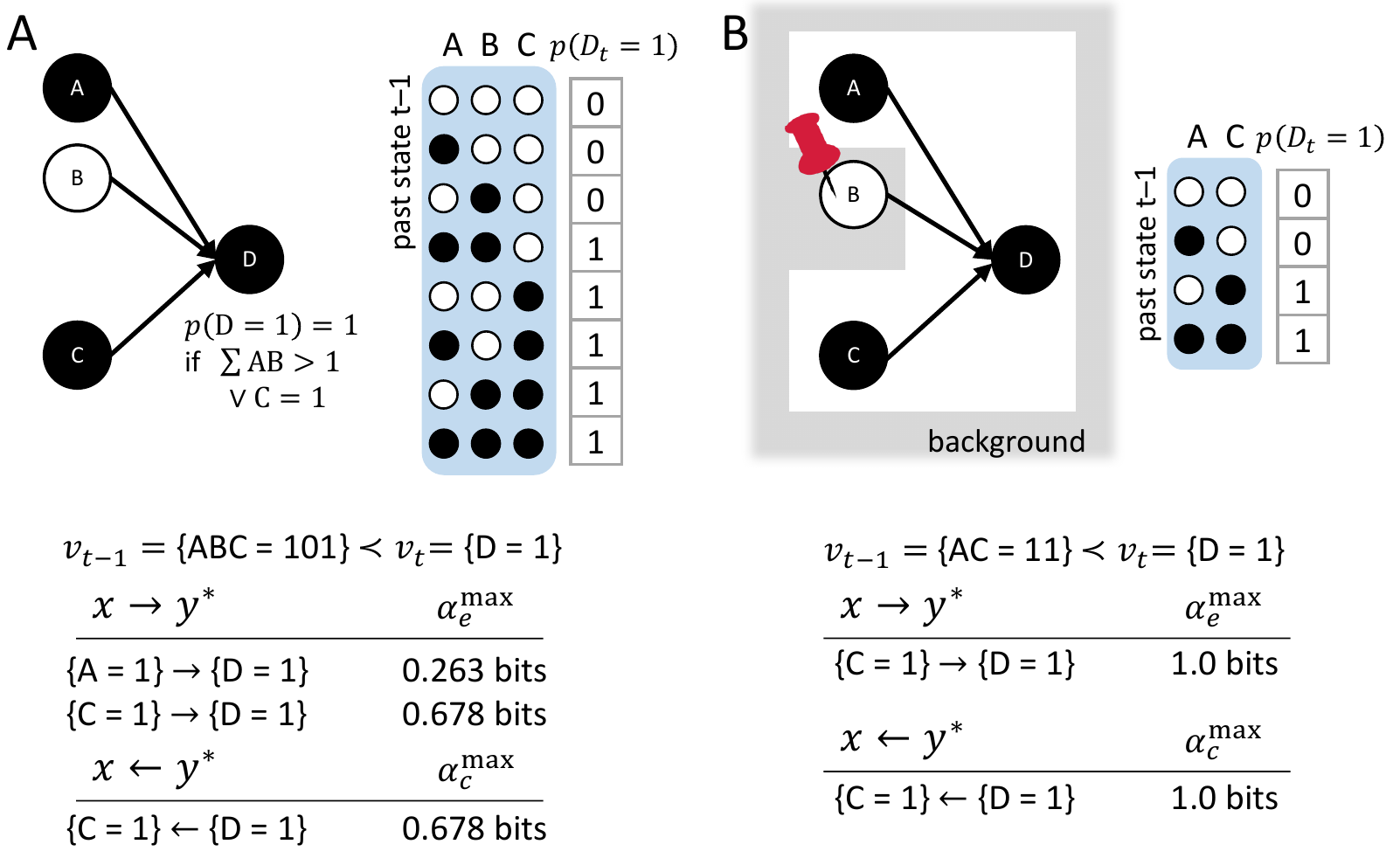}
\caption{Disjunction of two conjunctions $(A \wedge B) \vee C$. (A) All inputs to $D$ are considered relevant variables. (B) $B = 0$ is taken as a fixed background condition.}
\label{fig7}
\end{figure}

Here we consider a simple example, $(A \wedge B) \vee C$ (Fig. \ref{fig7}). The debate over this example is mostly concerned with the type of transition shown in Fig. \ref{fig7}A: $v_{t-1} = \{ABC = 101\} \prec v_t = \{D = 1\}$, and the question whether $\{A = 1\}$ is a cause of $\{D = 1\}$ even if $B = 0$.\footnote{One story accompanying this example is that ``a prisoner dies either if $A$ loads $B$'s gun and $B$ shoots, or if $C$ loads and shoots his gun, $\ldots$ $A$ loads $B$'s gun, $B$ does not shoot, but $C$ does load and shoot his gun, so that the prisoner dies'' \citep{Hopkins2003, Halpern2016}.} 

The quantitative assessment of actual causes and actual effects can help to resolve issues of actual causation in this type of example. As shown in Fig. \ref{fig7}A, with respect to actual effects, both causal links $\{A = 1\} \rightarrow \{D = 1\}$ and $\{C = 1\} \rightarrow \{D = 1\}$ are present, with $\{C = 1\}$ having a stronger actual effect. However, $\{C = 1\}$ is the one actual cause of $\{D = 1\}$, being the maximally irreducible cause with $\alpha_c^{\m}(\{D = 1\}) = 0.678$.

When judging the actual effect of $\{A = 1\}$ at $t-1$ within the transition $v_{t-1} = \{ABC = 101\} \prec v_t = \{D = 1\}$, $B$ is assumed to be undetermined. By itself, the occurrence $\{A = 1\}$ does raise the probability of occurrence $\{D = 1\}$, and thus $\{A = 1\} \rightarrow \{D = 1\}$. 

If we instead consider $\{B = 0\} \in \{U' = u'\}$ as a fixed background condition and evaluate the transition $v_{t-1} = \{AC = 11\} \prec v_t = \{D = 1\}$ in $G_{u'}$, $\{A = 1\}$ does not have an actual effect anymore (Fig. \ref{fig7}B). In this case, the background condition $\{B = 0\}$ prevents $\{A = 1\}$ from having any effect.

The results from this example extend to the general case of disjunctions of conjunctions. In the situation where $v_t=1$, the actual cause of $v_t$ is a minimally sufficient occurrence. If multiple conjunctive conditions are satisfied, the actual cause of $v_t$ remains indetermined between all minimally sufficient sets (asymmetric overdetermination). At $t-1$, any first-order occurrence in state `1', as well as any high-order occurrence of such nodes that does not overdetermine $v_t$, has an actual effect. This includes any occurrence in state all `1' that contains only variables from exactly one conjunction, as well as any high-order occurrence of nodes across conjunctions, which do not fully contain any specific conjunction. 

If instead $v_t=0$, then its actual cause is an occurrence that contains a single node in state `0' from each conjunctive condition. At $t-1$, any occurrence in state all `0' that does not overdetermine $v_t$ has an actual effect, which is any all `0' occurrence that does not contain more than one node from any conjunction.

These results are formalized by the following theorem. 
\begin{theorem}
\label{thm2}

Consider a dynamical causal network $G_u$ such that $V_t = \{Y_t\}$ is a DOC element that is a disjunction of $k$ conditions, each of which is a conjunction of $n_j$ inputs, and $V_{t-1} = \{\{V_{i, j, t-1}\}_{i = 1}^{n_j}\}_{j=1}^{k}$ is the set of its $n = \sum_j n_j$ inputs. For a transition $v_{t-1} \prec v_{t}$, the following holds: 

\begin{enumerate}
\item If $y_t = 1$,

\begin{enumerate}

\item The actual cause of $\{Y_t = 1\}$ is an occurrence $\{X_{t-1} = x_{t-1}\}$ where $x_{t-1} = \{x_{i, j, t-1}\}_{i=1}^{n_j} \subseteq v_{t-1}$ such that $\min(x_{t-1}) = 1$.

\item The actual effect of $\{X_{t-1} = x_{t-1}\}$ is $\{Y_t = 1\}$ if $\min(x_{t-1}) = 1$ and $|x_{t-1}| = c_j = n_j$; otherwise $x_{t-1}$ is reducible.

\end{enumerate}

\item If $y_t = 0$,

\begin{enumerate}

\item The actual cause of $\{Y_t = 0\}$ is an occurrence $x_{t-1} \subseteq v_{t-1}$ such that $\max(x_{t-1}) = 0$ and $c_j = 1 ~\forall~j$. 

\item If $\max(x_{t-1}) = 0$ and $c_j \leq 1~\forall~j$ then the actual effect of $\{X_{t-1} = x_{t-1}\}$ is $\{Y_t = 0\}$; otherwise $x_{t-1}$ is reducible. 

\end{enumerate}

\end{enumerate}
\end{theorem}
Proof: See \ref{proof}.

\subsection{Complicated voting}
 
\begin{figure}
\includegraphics[width = 4in]{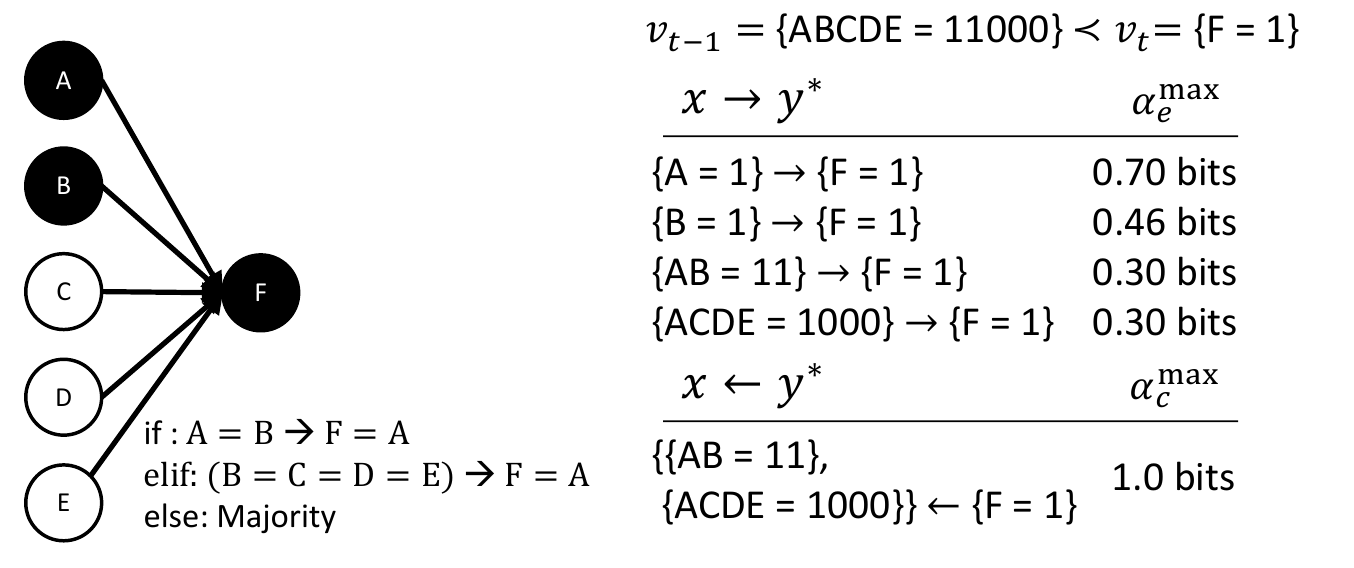}
\caption{\textbf{Complicated voting.}}
\label{fig8}
\end{figure}

As already demonstrated in the examples in Fig. \ref{fig5}C and D, the proposed causal analysis is not restricted to linear update functions or combinations thereof. Fig. \ref{fig8} depicts an example transition featuring a complicated, nonlinear update function. This specific example is taken from \citep{Halpern2015, Halpern2016}: If $A$ and $B$ agree, $F$ takes their value, if $B$, $C$, $D$, and $E$ agree, $F$ takes $A$'s value, otherwise majority decides. The transition of interest is $v_{t-1} = \{ABCDE = 11000\} \prec v_t = \{F = 1\}$.

According to \cite{Halpern2015}, intuition suggests that $\{A = 1\}$ together with $\{B = 1\}$ cause $\{F = 1\}$. Indeed, $\{AB = 11\}$ is one minimally sufficient occurrence in the transition that determines $\{F = 1\}$. The result of the present causal analysis of the transition (Fig. \ref{fig8}) is that both $\{AB = 11\}$ and $\{ACDE = 1000\}$ completely determine that $\{F = 1\}$ will occur with $\alpha_c(x_{t-1}, y_t) = \alpha_c^{\m}(y_t) = 1.0$. Thus, there is indeterminism between these two causes (see \ref{S1} for a comparison of our results with those of \cite{Halpern2015}).
In addition, the effects $\{A = 1\} \rightarrow \{F = 1\}$, $\{B = 1\} \rightarrow \{F = 1\}$, $\{AB = 11\} \rightarrow \{F = 1\}$, and $\{ACDE = 1000\} \rightarrow \{F = 1\}$ all contribute to the causal account. 

\subsection{Noise and probabilistic variables}

\begin{figure}
\includegraphics[width = 3.4in]{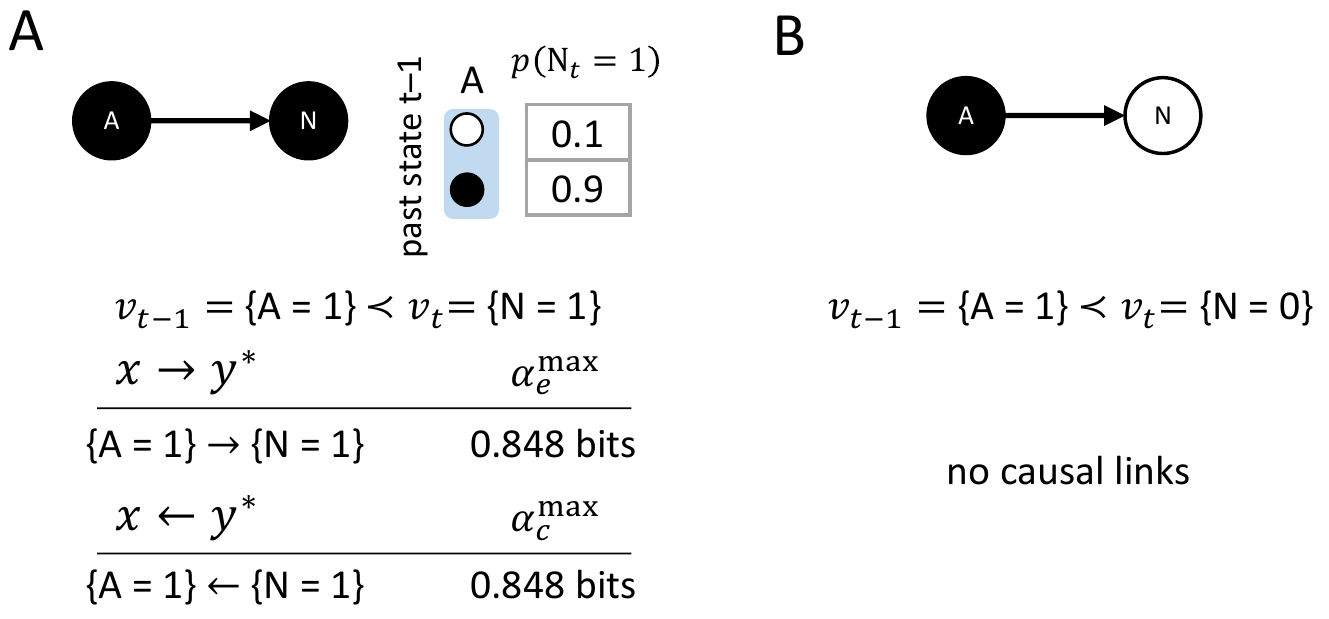}
\caption{\textbf{Probabilistic variables.} While the transition shown in (A) does have a deterministic equivalent, the transition shown in (B) would be impossible in the deterministic case.}
\label{fig9}
\end{figure}

The examples so far involved deterministic update functions. Probabilistic accounts of causation are closely related to counterfactual accounts \citep{Paul2013}. Nevertheless, certain problem cases only arise in probabilistic settings (\textit{e.g.} Fig. \ref{fig9}B). The present causal analysis can be applied equally to probabilistic and deterministic causal networks, as long as the system's transition probabilities satisfy conditional independence (Eqn. \ref{eqn1b}). No separate, probabilistic calculus for actual causation is required. 

In the simplest case, where noise is added to a deterministic transition $\transition$, the noise will generally decrease the strength of the causal links in the transition. Fig. \ref{fig9} shows the causal account of the transition $v_{t-1} = \{A = 1\} \prec v_t = \{N = 1\}$, where $N$ is the slightly noisy version of a COPY-gate. In this example, both $\{A = 1\} \rightarrow \{N = 1\}$ and $\{A = 1\} \leftarrow \{N = 1\}$. The only difference with the equivalent deterministic case is that the causal strength $\alpha_e^{\m} = \alpha_c^{\m} = 0.848$ is lower than in the deterministic case where $\alpha_e^{\m} = \alpha_c^{\m} = 1$. Note that in this probabilistic setting, the actual cause $\{A = 1\}$ by itself is not sufficient to determine $\{N = 1\}$. Nevertheless, $\{A = 1\}$ makes a positive difference in bringing about $\{N = 1\}$, and this difference is irreducible, so the causal link is present within the transition.

The transition $v_{t-1} = \{A = 1\} \prec v_t = \{N = 0\}$ has no counterpart in the deterministic case where $p(\{N = 0\}|\{A = 1\}) = 0$ (considering the transition would thus violate the realization principle). The result of the causal analysis is that there are no irreducible causal links within this transition. $\{A = 1\}$ decreases the probability of $\{N = 0\}$ and vice versa, which leads to $\alpha_{c/e}<0$. Consequently, $\alpha_{c/e}^{\m}=0$, as specified by the empty set. 
One interpretation is that the actual cause of $\{N = 0\}$ must lie outside of the system, such as a missing latent variable. Another interpretation is that the actual cause for $\{N = 0\}$ is genuine `physical noise', for example, within an element or connection. In any case, the proposed account of actual causation is sufficiently general to cover both deterministic as well as probabilistic systems. 

\subsection{Simple classifier}

As a final example, we consider a transition with a multi-variate $v_t$: the 3 variables $A$, $B$, and $C$ provide input to 3 different ``detectors", the nodes $D$, $S$, and $L$. $D$ is a ``dot-detector''; it outputs `1' if exactly one of the 3 inputs is in state `1'. $S$ is a ``segment-detector'': it outputs `1' for input states $\{ABC = 110\}$ and $\{ABC = 011\}$. $L$ detects lines, that is, $\{ABC = 111\}$.

\begin{figure}
\includegraphics[width = 4in]{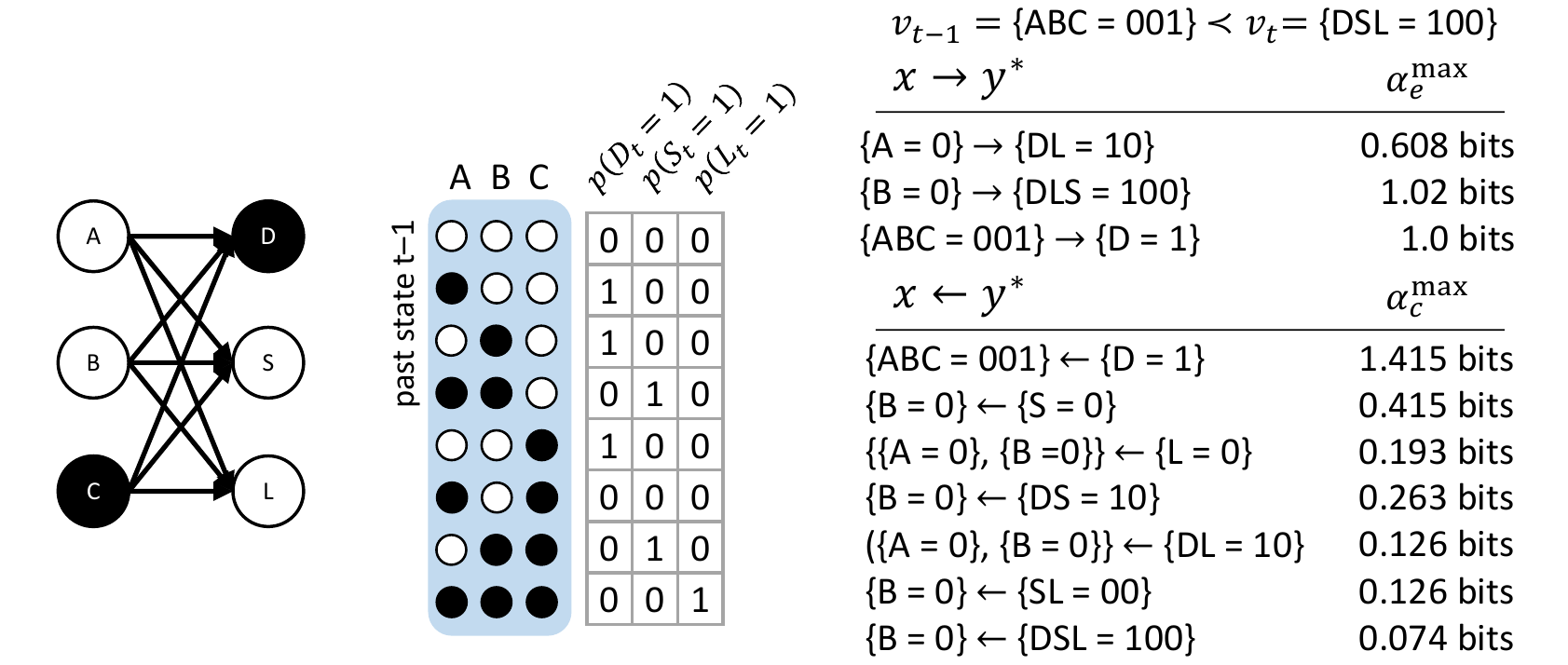}
\caption{\textbf{Simple classifier.} $D$ is a ``dot-detector'', $S$ a ``segment-detector'', and $L$ a ``line-detector'' (see text).}
\label{fig10}
\end{figure}

Fig. \ref{fig10} shows the causal account of the specific transition $v_{t-1} = \{ABC = 001\} \prec v_t = \{DSL = 100\}$. In this case, only a few occurrences $x_{t-1} \subseteq v_{t-1}$ have actual effects, but all possible occurrences $y_t \subseteq v_{t}$ are irreducible with their own actual cause. The occurrence $\{C = 1\}$ by itself, for example, has no actual effect. This may be initially surprising since $D$ is a dot detector and $\{C = 1\}$ is supposedly a dot. However, $\{C=1\}$ by itself does not raise the probability of $\{D=1\}$. The specific configuration of the entire input set is necessary to determine $\{D = 1\}$ (a dot is only a dot if the other inputs are `0'). Consequently, $\{ABC = 001\} \rightarrow \{D = 1\}$ and also $\{ABC = 001\} \leftarrow \{D = 1\}$. By contrast, the occurrence $\{A = 0\}$ is sufficient to determine $\{L = 0\}$ and raises the probability of $\{D = 1\}$; the occurrence $\{B = 0\}$ is sufficient to determine $\{S = 0\}$ and $\{L = 0\}$ and also raises the probability of $\{D = 1\}$. We thus get the following causal links: $\{A = 0\} \rightarrow \{DL = 10\}$, $\{\{A = 0\},\{B = 0\}\} \leftarrow \{L = 0\}$, $\{B = 0\} \rightarrow \{DSL = 100\}$ and $\{B = 0\} \leftarrow \{S = 0\}$. 

In addition, all high-order occurrences $y_t$ are irreducible, each having their own actual cause above those of their parts. The actual cause identified for these high-order occurrences can be interpreted as the ``strongest'' shared cause of nodes in the occurrence, for example $\{B = 0\} \leftarrow \{DS = 10\}$. While only the occurrence $\{ABC = 001\}$ is sufficient to determine $\{DS = 10\}$, this candidate causal link is reducible, because $\{DS = 10\}$ does not constrain the past state of $ABC$ any more than $\{D = 1\}$ by itself. In fact, the occurrence $\{S = 0\}$ does not constrain the past state of $AC$ at all. Thus $\{ABC = 001\}$ and all other candidate causes of $\{DS = 10\}$ that include these nodes are either reducible (because their causal link can be partitioned with $\alpha_c^{\m} = 0$) or excluded (because there is a subset of nodes whose causal strength is at least as high). In this example, $\{B = 0\}$ is the only irreducible shared cause of $\{D = 1\}$ and $\{S = 0\}$, and thus also the actual cause of $\{DS = 10\}$.

\section{Discussion}
\label{D}

In this article, we presented a principled, comprehensive formalism to assess actual causation within a given dynamical causal network $G_u$. For a transition $\transition$ in $G_u$, the proposed framework provides a complete causal account of all causal links between occurrences at $t-1$ and $t$ of the transition, based on five principles---realization, composition, information, integration, and exclusion. In what follows, we review specific features and limitations of our approach, discuss how the results relate to intuitive notions about actual causation and causal explanation, and highlight some of the main differences with previous proposals aimed at operationalizing the notion of actual causation. Specifically, our framework considers all counterfactual states rather than a single contingency, which makes it possible to assess the strength of causal links. Second, it distinguishes between actual causes and actual effects, which are considered separately. Third, it allows for causal composition, in the sense that first- and high-order occurrences can have their own causes and effects within the same transition, as long as they are irreducible. And fourth, it provides a rigorous treatment of causal overdetermination. As demonstrated in the results section and the \ref{S1}, the proposed formalism is generally applicable to a vast range of physical systems, whether deterministic or probabilistic, with binary or multi-valued variables, feedforward or recurrent architectures, as well as narrative examples, as long as they can be represented as a causal network with an explicit temporal order.


\subsection{Testing all possible counterfactuals with equal probability}
\label{D1}

\begin{figure}
\includegraphics[width = 3in]{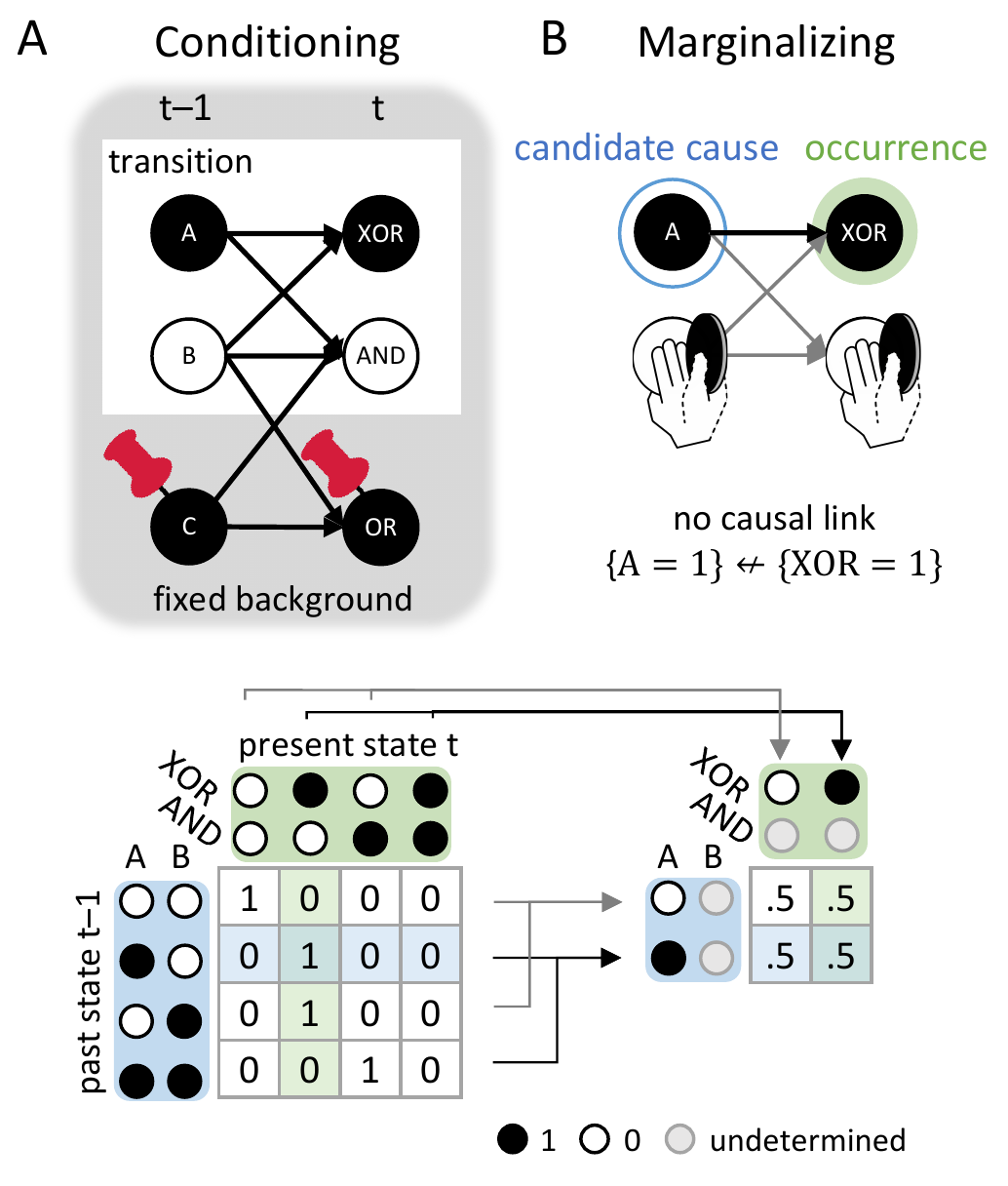}
\caption{\textbf{Causal conditioning and marginalizing.} (A) Variables outside the transition of interest are treated as fixed background conditions (indicated by the red pins). The transition probabilities $p(v_t|v_{t-1})$ are conditioned on the state of these exogenous variables. (B) When evaluating the strength of a causal link within the transition, the remaining variables in $G_u$ but outside the causal link are causally marginalized, i.e., replaced by an average across all their possible states. With $B$ marginalized, the state of $A$ by itself does not determine and is not determined by the occurrence $\{\text{XOR} = 1\}$.}
\label{fig10B}
\end{figure}


In the simplest case, counterfactual approaches to actual causation are based on the ``but-for'' test \citep{Halpern2016}: $C = c$ is a cause of $E = e$ if $C = \neg c$ implies $E= \neg e$ (``but for $c$, $e$ would not have happened''). In multi-variate causal networks this condition is typically dependent on the remaining variables $W$. What differs among current counterfactual approaches are the permissible \textit{contingencies} ($W=w$) under which the ``but-for'' test is applied
(e.g., \cite{Hitchcock2001,Yablo2002,Woodward2003,Halpern2005,Hall2007,Halpern2015, Weslake2015-WESAPT}) (see \ref{S1}). Moreover, if there is one permissible contingency (counterfactual state) $\{\neg c, w\}$ that implies $E = \neg e$, $c$ is identified as a cause of $e$ in an ``all-or-nothing" manner. In sum, current approaches test for counterfactual dependence under a fixed contingency $W = w$, evaluating a particular counterfactual state $C = \neg c$. 

Our starting point is a realization of a dynamical causal network $G_u$, which is a transition $\transition$ that is compatible with $G_u$'s transition probabilities ($p_u(v_t | v_{t-1}) > 0$) given the fixed background conditions $U = u$ (Fig. \ref{fig10B}A). However, we employ \textit{causal marginalization} instead of fixed $W = w$ and $C = \neg c$ within the transition. This means that we replace these variables with an average over \textit{all} their possible states (Eqn. \ref{cmarg}). 

Applied to variables outside of the candidate causal link (Fig. \ref{fig10B}B), causal marginalization serves to remove the influence of these variables on the causal dependency between the occurrence and its candidate cause (or effect), which is thus evaluated based on its own merits. The difference between marginalizing the variables outside the causal link of interest and treating them as fixed contingencies becomes apparent in the case of an XOR (``exclusive OR") mechanism in Fig. \ref{fig10B} (or equivalently the biconditional (XNOR), Fig. \ref{fig5}C). With input $B$ fixed in a particular state (`0' or `1') the state of the XOR will completely depend on the state of $A$. However, the state of $A$ alone does not determine the state of the XOR at all if $B$ is marginalized. The latter better captures the mechanistic nature of the XOR, which requires a difference in $A$ and $B$ to switch on (`1').

We also marginalize across all possible states of $C$ in order to determine whether $e$ counterfactually depends on $c$. Instead of identifying one particular $C = \neg c$ for which $E = \neg e$, all of $C$'s states are equally taken into account. The notion that counterfactual dependence is an ``all-or-nothing concept" \citep{Halpern2016} becomes problematic, for example, if non-binary variables are considered (see \ref{S1}) and also in non-deterministic settings. By contrast, our proposed approach, that considers all possible states of $C$, naturally extends to the case of multi-valued variables and probabilistic causal networks. Moreover, it has the additional benefit that we can quantify the strength of the causal link between an occurrence and its actual cause (effect). In the present framework, having a positive effect ratio $\rho_e(x_{t-1}, y_t) > 0$ is necessary but not sufficient for $x_{t-1} \rightarrow y_t$, and the same for a positive cause ratio $\rho_c(x_{t-1}, y_t) > 0$. 

Taken together, we argue that causal marginalization, that is, averaging over contingencies and all possible counterfactuals of an occurrence, reveals the mechanisms underlying the transition. By contrast, fixing relevant variables to any one specific state largely ignores them. 
This is because a mechanism is only fully described by all its transition probabilities, for all possible input states (Eqn. \ref{eqn1b}). For example, the biconditional $E$ in Fig. \ref{fig5}C, only differs from the conjunction $D$ in Fig. \ref{fig5}B, for the input state {AB = 00}. 
Once the underlying mechanisms are specified based on all possible transition probabilities, causal interactions can be quantified in probabilistic terms \citep{Ay2008, Oizumi2014} even within a single transition $\transition$, \textit{i.e.} in the context of actual causation \citep{Glennan2011, Korb2011}. However, this also means that all transition probabilities have to be known for the proposed causal analysis, even for states that are not typically observed (see also \cite{Ay2008,Balduzzi2008,Janzing2013,Oizumi2014}). 

Finally, in our analysis all possible past states are weighted equally in the causal marginalization. Related measures of information flow in causal networks \citep{Ay2008} and causal information \citep{Korb2011} consider weights based on a distribution of $p(v_{t-1})$, for example, the stationary distribution, or observed probabilities, or also a maximum entropy distribution (equivalent to weighting all states equally). 
However, in the context of actual causation, the prior probabilities of occurrences at $t-1$ are extraneous to the question ``what caused what?" 
All that matters is what actually happened, the transition $\transition$, and the underlying mechanisms. How likely $v_{t-1}$ was to occur should not influence the causes and effects within the transition, nor how strong the causal links are between actual occurrences at $t-1$ and $t$. In other words, the same transition, involving the same mechanisms and background conditions should always result in the same causal account. Take, for instance, a set of nodes $AB$ that output to $C$, which is a deterministic OR-gate. If $C$ receives no further inputs from other nodes, then whenever $\{AB = 11\}$ and $\{C = 1\}$, the causal links, their strength, and the causal account of the transition $\{AB = 11\} \prec \{C = 1\}$ should be the same as in Fig. \ref{fig5}A (``Disjunction"). Which larger system the set of nodes was embedded in, or what the probability was for the transition to happen in the first place, according to the equilibrium, observed, or any other distribution is not relevant in this context. Let us assume, for example, that $\{A = 1\}$ was much more likely to occur than $\{B = 1\}$. This bias in prior probability does not change the fact that, mechanistically, $\{A = 1\}$ and $\{B = 1\}$ have the same effect on $\{C = 1\}$ and are equivalent causes.

\subsection{Distinguishing actual effects and actual causes}
\label{D2}

An implicit assumption commonly made about (actual) causation is that the relation between cause and effect is bidirectional: if occurrence $C=c$ had an effect on occurrence $E=e$, then $c$ is assumed to be a cause of $e$ \citep{Hitchcock2001, Yablo2002, Woodward2003, Halpern2005, Hall2007, Halpern2015, Weslake2015-WESAPT, Twardy2011, Fenton-Glynn2017}. 
As demonstrated throughout the Results section, however, this conflation of causes and effects is untenable once multi-variate transitions $\transition$ are considered (see also next, \ref{D3}). There, an asymmetry between causes and effects simply arises due to the fact that the set of variables that is affected by an occurrence $x_{t-1} \subseteq v_{t-1}$ typically differs from the set of variables that affects an occurrence $y_t \subseteq v_t$. Take the toy classifier example in Fig. \ref{fig10}: while $\{B = 0\}$ is the actual cause of $\{S = 0\}$, $\{B = 0\}$'s actual effect is $\{DLS = 100\}$. 

Accordingly, we propose that a comprehensive causal understanding of a given transition is provided by its complete causal account $\C$ (Definition \ref{def4}), including both actual effects and actual causes. Actual effects are identified from the perspective of occurrences at $t-1$, whereas actual causes are identified from the perspective of occurrences at $t$. This means that also the causal principles of composition, integration, and exclusion are applied from these two perspectives. When we evaluate causal links of the form $x_{t-1} \rightarrow y_t$, any occurrence $x_{t-1}$ may have one actual effect $y_{t} \subseteq v_{t}$ if $x_{t-1}$ is irreducible ($\alpha^{\m}_e(x_{t-1}) > 0$) (Definition \ref{def2}). When we evaluate causal links of the form $x_{t-1} \leftarrow y_t$, any occurrence $y_t$ may have one actual cause $x_t \subseteq v_{t-1}$ if $y_t$ is irreducible ($\alpha^{\m}_c(y_t) > 0$) (Definition \ref{def1}).
As seen in the first example (Fig. \ref{fig4}), there may be a high-order causal link in one direction, but the reverse link may be reducible.

As mentioned in the Introduction and exemplified in the \ref{S1}, our approach has a more general scope but is still compatible with the traditional view of actual causation, concerned only with actual causes of singleton occurrences. Nevertheless, even in the limited setting of singleton $v_t$, considering both causes and effects may be illuminating. Consider, for example, the transition shown in Fig. \ref{fig7}A: by itself, the occurrence $\{A = 1\}$ raises the probability of $\{D = 1\}$ ($\rho_e(x_{t-1}, y_t) = \alpha_e(x_{t-1}, y_t) > 0$), which is a common determinant of being a cause in probabilistic accounts of (actual) causation \citep{Good1961, Suppes1970, Eells1983, Pearl2009}. However, even in deterministic systems with multi-variate dependencies, the fact that an occurrence $c$, by itself, raises the probability of an occurrence $e$, does not necessarily determine that $E=e$ will actually occur \citep{Paul2013}. In the example of Fig. \ref{fig7}, $\{A = 1\}$ is neither necessary nor sufficient for $\{D = 1\}$. Here, this issue is resolved by acknowledging that both $\{A = 1\}$ and $\{C = 1\}$ have an actual effect on $\{D = 1\}$, whereas $\{C=1\}$ is identified as the (one) actual cause of $\{D=1\}$,\footnote{Note that Pearl initially proposed maximizing the posterior probability $p(c \mid e)$ as a means of identifying the best (``most probable'') explanation for an occurrence $e$ (\cite{Pearl1988}; Chapter 5). This approach has later been criticized, among others, by Pearl himself (\cite{Pearl2000}; Chapter 7), as it had been formalized in purely probabilistic terms, lacking the notion of system interventions. Moreover, without a notion of irreducibility, as applied in the present framework, explanations based on $p(c \mid e)$ tend to include irrelevant variables \citep{Shimony1991, Chajewska1997}.} in line with intuition \citep{Halpern2015}. 


In sum, an actual effect $x_{t-1} \rightarrow y_t$ does not imply the corresponding actual cause $x_{t-1} \leftarrow y_t$ and vice versa. Including both directions in the causal account may thus provide a more comprehensive explanation of ``what happened'' in terms of ``what caused what''.

\begin{figure}
\includegraphics[width = 3.8in]{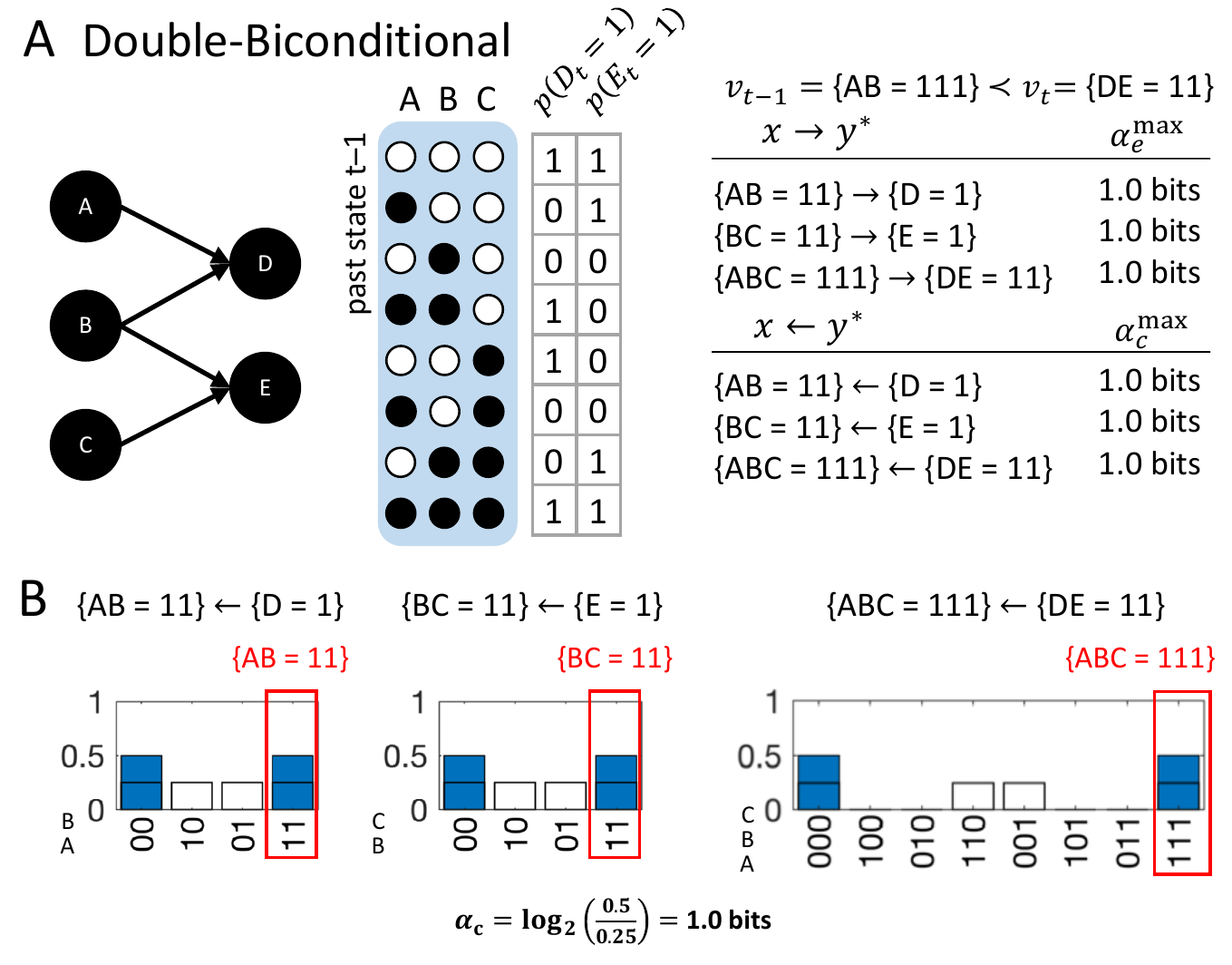}
\caption{\textbf{Composition: high-order occurrences.} (A) Double Biconditional: Transition and causal account. (B) Cause repertoires corresponding to the two first-order and one second-order occurrences with actual causes (see main text).}
\label{fig11}
\end{figure}

\subsection{Composition}
\label{D3}

The proposed framework of actual causation explicitly acknowledges that there may be high-order occurrences, which have genuine actual causes or actual effects. 
While multi-variate dependencies play an important role in complex distributed systems \citep{Mitchell1998, Sporns2000, Wolff2018}, they are largely ignored in the actual causation literature.

From a strictly informational perspective focused on predicting $y_t$ from $x_{t-1}$, one might be tempted to disregard such compositional occurrences and their actual effects, since they do not add predictive power. For instance, the actual effect of $\{AB = 11\}$ in the conjunction example of Fig. \ref{fig5}B is informationally redundant, since $\{D = 1\}$ can be inferred (predicted) from $\{A = 1\}$ and $\{B = 1\}$ alone. From a causal perspective, however, such compositional causal links specify mechanistic constraints that would not be captured otherwise. It is these mechanistic constraints, and not predictive powers, that provide an explanation for ``what happened'' in the various transitions shown in Fig. \ref{fig5} by revealing ``what caused what''.
In Fig. \ref{fig5}C for example, the individual nodes $A$ and $B$ do not fulfill the most basic criterion for having an effect on the XNOR node $\{E = 1\}$ as $\rho_e(x_{t-1}, y_t) = 0$, whereas the second-order occurrence $\{AB = 11\}$ has the actual effect $\{E = 1\}$. In the conjunction example (Fig. \ref{fig5}B), $\{A = 1\}$ and $\{B = 1\}$ both constrain the AND-gate $D$ in the same way, but the occurrence $\{AB = 11\}$ further raises the probability of $\{D = 1\}$ compared to the effect of each individual input. The presence of causal links specified by first-order occurrences does not exclude the second-order occurrence $\{AB = 11\}$ from having an additional effect on $\{D = 1\}$.

To illustrate this with respect to both actual causes and actual effects, we can extend the XNOR example to a ``double-biconditional'' and consider the transition $v_{t-1} = \{ABC = 111\} \prec v_t = \{DE = 11\}$ (Fig. \ref{fig11}). In the figure, both $D$ and $E$ are XNOR nodes that share one of their inputs (node $B$), and $\{AB = 11\} \leftarrow \{D = 1\}$ and $\{BC = 11\} \leftarrow \{E = 1\}$. As illustrated by the cause-repertoires shown in Fig. \ref{fig11}B, and in accordance with $D$'s and $E$'s logic function (mechanism), the actual cause of $\{D = 1\}$ can be described as the fact that $A$ and $B$ were in the same state, and the actual cause of $\{E = 1\}$ as the fact that $B$ and $C$ were in the same state. In addition to these first-order occurrences, also the second-order occurrence $\{DE = 11\}$ has an actual cause $\{ABC = 111\}$, which can be described as the fact that all three nodes $A$, $B$, and $C$ were in the same state. Crucially, this fact is not captured by either the actual cause of $\{D = 1\}$, or by the actual cause of $\{E = 1\}$, but only by the constraints of the second-order occurrence $\{DE = 11\}$. On the other hand, the causal link $\{ABC = 111\} \leftarrow \{DE = 11\}$ cannot capture the fact that $\{AB = 11\}$ was the actual cause of $\{D = 1\}$ and $\{BC = 11\}$ was the actual cause of $\{E = 1\}$. Of note, in this example the same reasoning applies to the composition of high-order occurrences at $t-1$ and their actual effects.

In sum, high-order occurrences capture multi-variate mechanistic dependencies between the occurrence's variables that are not revealed by the actual causes and effects of their parts. Moreover, a high-order occurrence does not exclude lower-order occurrences over their parts, which specify their own actual causes and effects. In this way, the composition principle makes explicit that high-order and first-order occurrences all contribute to the explanatory power of the causal account.

\begin{figure}
\includegraphics[width = 4in]{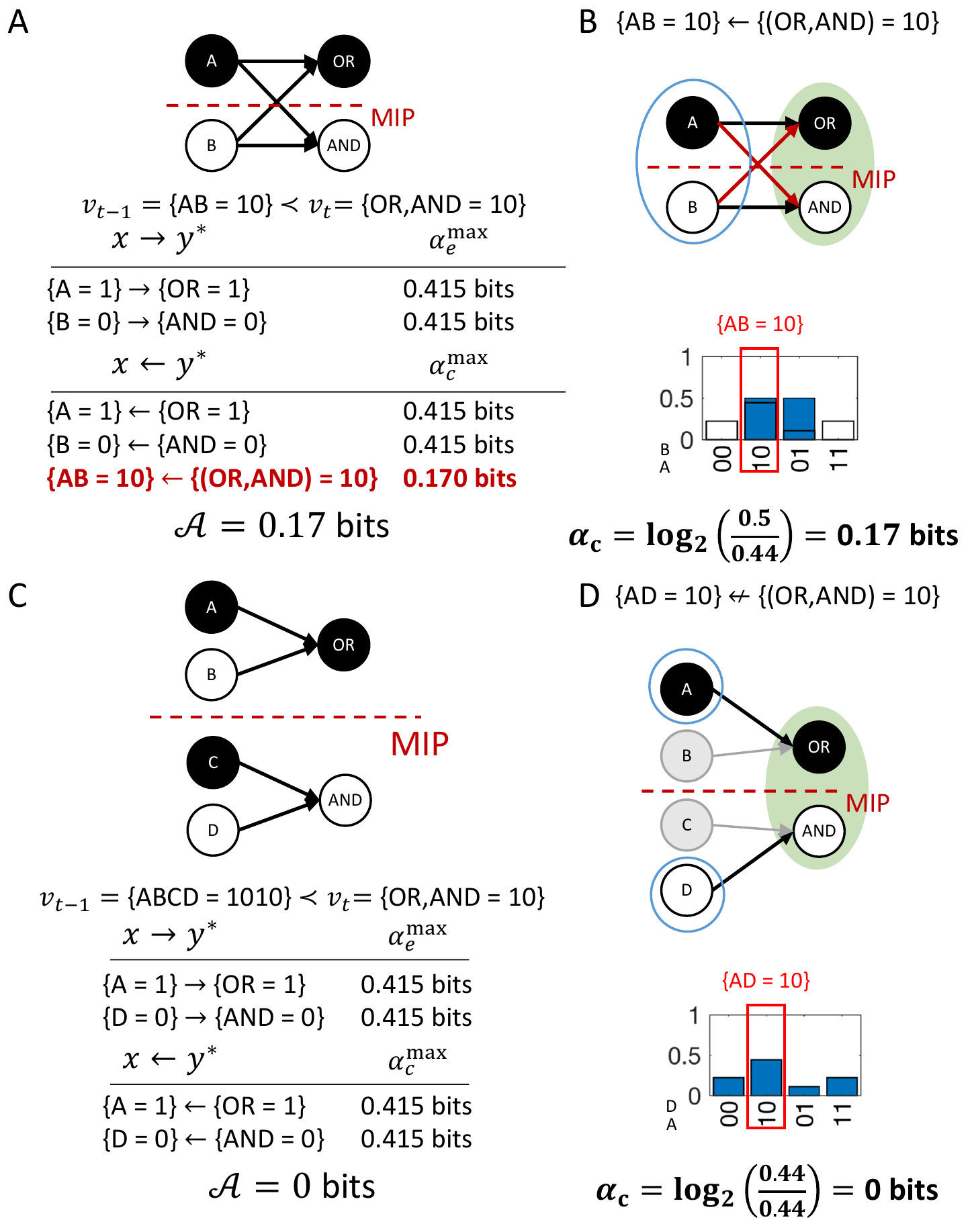}
\caption{\textbf{Integration: irreducible vs. reducible occurrences.} (A) Transition and causal account of Fig. \ref{fig4}. (B) The second--order occurrence $\{(OR, AND) = 10\}$ with actual cause $\{AB = 10\}$ is irreducible under the MIP. (C) Reducible transition with equivalent first-order causal links, but missing the second--order causal link present in (A). (D) The constraints specified by the second--order occurrence $\{(OR, AND) = 10\}$ here are the same, and thus reducible, to those under the MIP.}
\label{fig12}
\end{figure}

\subsection{Integration}
\label{D4}

As discussed above, high-order occurrences can have actual causes and effects, but only if they are irreducible to their parts. This is illustrated in Fig. \ref{fig12}, in which a transition equivalent to our initial example in Fig. \ref{fig4} (Fig. \ref{fig12}A) is compared against a similar, but reducible transition (Fig. \ref{fig12}C) in a different causal network. The two situations differ mechanistically: the OR and AND gate in Fig. \ref{fig12}A receive common inputs from the same two nodes, while the OR and AND in Fig. \ref{fig12}C have independent sets of inputs. Nevertheless, the actual causes and effects of all single-variable occurrences are identical in the two cases. In both transitions, $\{\text{OR} = 1\}$ is caused by its one input in state `1', and $\{\text{AND} = 0\}$ is caused by its one input in state `0'. What distinguishes the two causal accounts is the additional causal link in Fig. \ref{fig12}A, between the second-order occurrence $\{\text{(OR,AND)} = 10\}$ and its actual cause $\{AB = 10\}$. $\{\text{(OR,AND)} = 10\}$ raises the probability of both $\{AB = 10\}$ (in Fig. \ref{fig12}A) and $\{AD = 10\}$ (in Fig. \ref{fig12}C) compared to their unconstrained probability $\pi = 0.25$, and thus $\rho_c(x_{t-1}, y_t) > 0$ in both cases. Yet, only $\{AB = 10\} \leftarrow \{\text{(OR,AND)} = 10\}$ in Fig. \ref{fig12}A is irreducible to its parts. This is shown by partitioning across the $\mip$ with $\alpha_c(x_{t-1}, y_t) = 0.17$. This second-order occurrence thus specifies that the OR and AND gate in Fig. \ref{fig12}A receive common inputs---a fact that would otherwise remain undetected. 

As described in the \ref{S2}, using the measure $\A(\transition)$ we can also quantify the extent to which the entire causal account $\C$ of a transition $\transition$ is irreducible. $\A(\transition) = 0$ indicates that $\transition$ can either be decomposed into multiple transitions without causal links between them (Fig. \ref{fig12}C), or includes variables without any causal role in the transition (e.g., Fig. \ref{fig5}D).

\begin{figure}
\includegraphics[width = \textwidth]{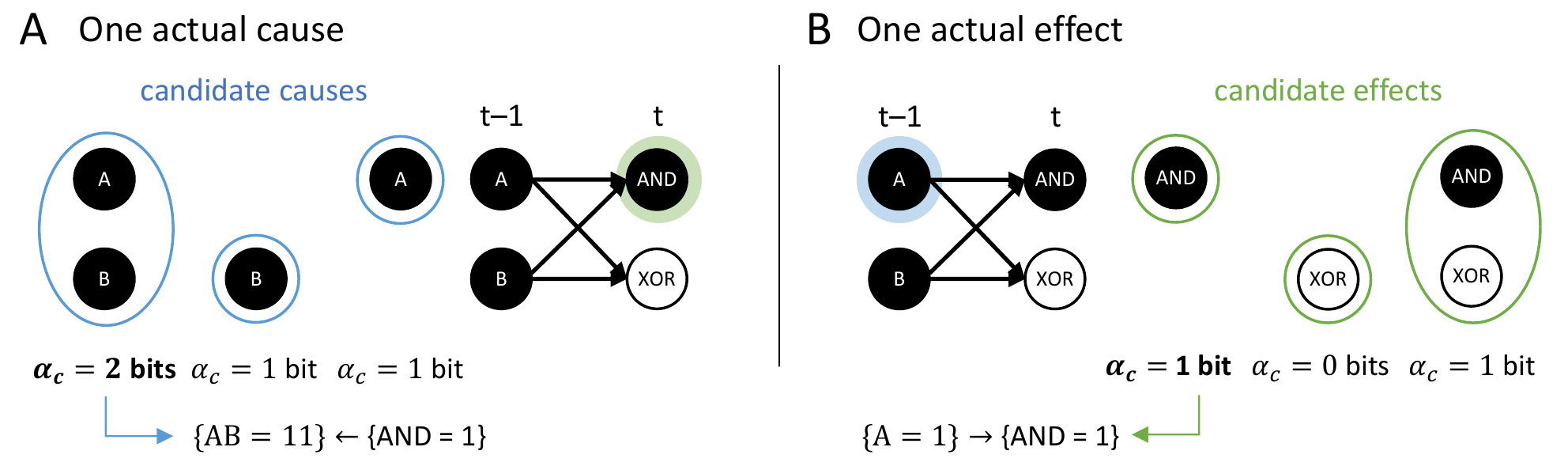}
\caption{\textbf{Exclusion: any occurrence can at most have one actual cause or effect.} (A) Out of the three candidate causes $\{A = 1\}$, $\{B = 1\}$, and $\{AB = 11\}$, the actual cause of $\{\text{AND} = 1\}$ is the high-order occurrence $\{AB = 11\}$ with $\alpha_c = \alpha^{\m}_c = 2.0$ bits. (B) Out of the three candidate effects, $\{\text{AND} = 1\}$, $\{\text{XOR} = 1\}$, and $\{\text{(AND, XOR)} = 11\}$, the actual effect of $\{A = 1\}$ is the first-order occurrence $\{AND = 1\}$ with $\alpha_e = \alpha^{\m}_e = 1.0$ bit. $\{\text{(AND, XOR)} = 11\}$ is excluded by the minimality condition (Definition \ref{def2}).}
\label{fig13}
\end{figure}

\subsection{Exclusion}
\label{D4.5}

That an occurrence can affect several variables (high-order effect), and that the cause of an occurrence can involve several variables (high-order cause) is uncontroversial \citep{Woodward2010}. Nevertheless, the possibility of multi-variate causes and effects is rarely addressed in a rigorous manner. Instead of one high-order occurrence, contingency-based approaches to actual causation typically identify multiple first-order occurrences as separate causes in these cases (see also \ref{S1}). This is because some approaches only allow for first-order causes by definition (e.g., \cite{Weslake2015-WESAPT}), while other accounts include a minimality clause that does not consider causal strength and thus excludes virtually all high-order occurrences in practice (e.g., \cite{Halpern2005}, but see \cite{Halpern2015}). Take the example of a simple conjunction $\text{AND} = A \land B$ in transition $\{AB = 11\} \prec \{\text{AND}=1\}$  (Fig. \ref{fig5}B, and Fig. \ref{fig13}). To our knowledge, all contingency-based approaches regard the first-order occurrences $\{A = 1\}$ and $\{B = 1\}$ as two separate causes of $\{\text{AND}=1\}$ in this case (but see \cite{Datta2016}), while we identify the second-order occurrence $\{AB = 11\}$ (the conjunction) as the one actual cause with $\alpha^{\m}_c$. 

Given a particular occurrence $x_{t-1}$ in the transition $\transition$, we explicitly consider the whole power set of $v_t$ as candidate effects of $x_{t-1}$, and the whole power set of $v_{t-1}$ as candidate causes of a particular occurrence $y_t$ (Fig. \ref{fig13}). However, the possibility of genuine multi-variate actual causes and effects requires a principled treatment of causal overdetermination. While most approaches to actual causation generally allow for both $\{A = 1\}$ and $\{B = 1\}$ to be actual causes of $\{\text{AND} = 1\}$, this seemingly innocent violation of the causal exclusion principle becomes prohibitive once $\{A = 1\}$, $\{B = 1\}$, and $\{AB = 11\}$ are recognized as candidate causes. In this case, either $\{AB = 11\}$ was the actual cause, or $\{A = 1\}$, or $\{B = 1\}$. Allowing for any combination of these occurrences, however, would be illogical. Within our framework, any occurrence can thus have at most one actual cause (or effect) within a transition---the minimal occurrence with $\alpha^{\m}$ (Fig. \ref{fig13}). Finally, cases of true, mechanistic overdetermination due to symmetries in the causal network are resolved by leaving the actual cause (effect) indetermined between all $x^*(y_t)$ with $\alpha^{\m}_c$ (see Definitions \ref{def1} and \ref{def2}). In this way, the causal account provides a complete picture of the actual mechanistic constraints within a given transition.

\subsection{Intended scope and limitations}
\label{D5}

The objective of many existing approaches to actual causation is to provide an account of people's intuitive causal judgments \citep{Halpern2016}. For this reason, the literature on actual causation is largely rooted in examples involving situational narratives, such as ``Billy and Suzy throw rocks at a bottle'' \citep{Pearl2000, Halpern2016}, which are then compressed into a causal model to be investigated. Such narratives can serve as intuition pumps, but can also lead to confusion if important aspects of the story are omitted in the causal model applied to the example \citep{Hitchcock2007, Paul2013, Weslake2015-WESAPT} (see \ref{S1}). 

Our objective is to provide a principled, quantitative causal account of ``what caused what" within a fully specified (complete) model of a physical systems of interacting elements. We purposely set aside issues regarding model selection or incomplete causal knowledge in order to formulate a rigorous theoretical framework applicable to any predetermined, dynamical causal network \citep{Pearl2010, Halpern2016}. This puts the explanatory burden on the formal framework of actual causation, rather than on the adequacy of the model. In this setting, causal models should always be interpreted mechanistically and time is explicitly taken into account. Rather than on capturing people's intuitions, an emphasis is put on explanatory power and consistency (see also \cite{Paul2013}). With a proper formalism in place, future work should address to what extent and under which conditions the identified actual causes and effects generalize across possible levels of description (macro vs. micro causes and effects), or under incomplete knowledge (see also \cite{Rubenstein2017, Marshall2018}).


In addition, the examples examined in this study have been limited to direct causes and effects within transitions $\transition$ across a single system update. The explanatory power of the proposed framework was illustrated in several examples, which included paradigmatic problem cases involving overdetermination and prevention. Yet, some prominent examples that raise issues of ``preemption'' or ``causation by omission'' have no direct equivalent in these basic types of physical causal models (see \ref{S1}). While the approach can, in principle, identify and quantify counterfactual dependencies across $k > 1$ time steps by replacing $p_u(v_t \mid v_{t-1})$ with $p_u(v_t \mid v_{t-k})$ in Eqn. \ref{eqn1b}, for the purpose of tracing a causal chain back in time \citep{Datta2016}, the role of intermediary occurrences remains to be investigated. Nevertheless, the present framework is unique in providing a general, quantitative, and principled approach to actual causation that naturally extends beyond simple, binary, and deterministic example cases to all mechanistic systems that can be represented by a set of transition probabilities as specified in Eqn. \ref{eqn1b}. 


\subsection{Accountability and causal responsibility}

This work presents a step towards a quantitative causal understanding of ``what is happening" in systems such as natural or artificial neural networks, computers, and other discrete, distributed dynamical systems. Such causal knowledge can be invaluable, for example, to identify the reasons for an erroneous classification by a convolutional neural network \citep{Szegedy2013}, or the source of a protocol violation in a computer network \citep{Datta2015}. A notion of multi-variate actual causes and effects, in particular, is crucial for addressing questions of accountability, or sources of network failures \citep{Halpern2016} in distributed systems. A better understanding of the actual causal links that govern a system's transitions should also improve our ability to effectively control the dynamical evolution of such systems and to identify adverse system states that would lead to unwanted system behaviors. 

Finally, a principled approach to actual causation in neural networks may illuminate the causes of an agent's actions or decisions (biological or artificial) \citep{Economist2018, WillKnight, damasio2012neurobiology}, including the causal origin of voluntary actions \citep{Haggard2008}. However, addressing the question ``who caused what?'', as opposed to ``what caused what'', implies modeling an agent with intrinsic causal power and intention \citep{Tononi2013, Datta2015}. Future work will combine the present mechanistic framework for actual causation with a mechanistic account of autonomous, causal agents, based on the same set of principles \citep{Marshall, Oizumi2014}.

\begin{supplement}
  \sname{Supplementary Discussion}\label{S1}
  \stitle{}
  \slink[]{}
  \sdatatype{.pdf} 
  \sdescription{Comparison to current counterfactual approaches to actual causation.}
\end{supplement}

\begin{supplement}
  \sname{Supplementary Methods}\label{S2}
  \stitle{}
  \slink[]{}
  \sdatatype{.pdf} 
  \sdescription{Irreducibility of the causal account.}
\end{supplement}

\begin{supplement}
  \sname{Supplementary Proofs}\label{proof}
  \stitle{}
  \slink[]{}
  \sdatatype{.pdf} 
  \sdescription{Proof of Theorem \ref{thm1} and \ref{thm2}.}
\end{supplement}

\bibliographystyle{imsart-nameyear}
\bibliography{AC_bibtex}

\end{document}


\begin{frontmatter}

\runauthor{Albantakis et al.}
\runtitle{Supplementary Discussion}
\end{frontmatter}

\title{Comparison to current counterfactual approaches to actual causation}
\vspace{0.2in}

Current counterfactual approaches to actual causation all regard $C = c$ as an actual cause of $E = e$ if at least one permissible setting of the remaining variables exists under which $C = \neg c$ leads to $E = \neg e$. What distinguishes the various approaches is their (increasingly elaborate) prescribed sets of conditions for permissible settings, or ``contingencies" (e.g., \cite{Hitchcock2001,Yablo2002,Woodward2003,Halpern2005,Hall2007,Halpern2015,Weslake2015-WESAPT}). The notion of counterfactual dependence as a basis for causal judgments can be traced back to \cite{Hume1758} and was elaborated by \cite{Lewis1973a}. 
\cite{Pearl2000} provided a theoretical formulation of counterfactual dependence within the framework of causal models paired with structural equations. The subsequently developed Halpern-Pearl (``HP") definition of actual causation introduced in \citep{Halpern2001} and updated in \citep{Halpern2005} remains one of the most influential proposals to date and is frequently used as a standard for comparison. Nevertheless, counterexamples exist that raise concerns about the generality of the HP-definition as an account of actual causation (e.g., \cite{Weslake2015-WESAPT, Halpern2015}). \cite{Halpern2015, Halpern2016} describes the original and updated HP-definition of actual causation, as well as a modified version introduced in \citep{Halpern2015}, and compares the different proposals based on a set of frequently cited problem cases. \cite{Weslake2015-WESAPT} provides an overview of counterfactual approaches to actual causation highlighting specific example cases that pose problems for the various accounts\footnote{Note, however, that Weslake only allows for singleton causes (individual variables), which is unnecessarily restrictive and goes against the causal composition principle (main text). Halpern's modified HP-definition \citep{Halpern2015}, for example, is equivalent to ``PRE" \citep{Weslake2015-WESAPT}, but allowing for multi-variate causes, which resolves several problem cases.}. While \cite{Weslake2015-WESAPT} demonstrates how refining the contingency conditions may address an increasing number of test cases, he also points out that the resulting sets of conditions lack a principled justification beyond their successes in singled-out examples. The same point has recently been made by \cite{Beckers2018}\footnote{\cite{Beckers2018} arrive at a definition of actual causation based on contrastive production within a valid timing, following a set of four ``principles" derived from typical problem cases. Their proposed account bears some resemblance to that of \cite{Weslake2015-WESAPT}.}. 

In the following, we will exemplify similarities and differences between our proposed account of actual causation (``IIT-account") and the various HP-definitions, with reference to several other proposed contingency conditions \citep{Hitchcock2001,Woodward2003, Weslake2015-WESAPT}, as well as probabilistic accounts of actual causation \citep{Twardy2011, Fenton-Glynn2017}.
Since these comparison accounts do not consider actual effects or causal composition, here, we will only discuss actual causes of singleton occurrences.
Our objective is to demonstrate that the IIT-account has the potential to provide a principled approach, rather than sets of contingency conditions. First, however, we begin with a few remarks on differences in motivation and scope between the IIT-account and most current approaches to actual causation. For completeness and clarity, this supplementary discussion repeats several points already addressed in the main text.

\section{Differences in motivation and scope}

A main motivation behind most accounts of actual causation is to accurately capture people's intuitive causal judgments. For this reason, much value is placed on how a given account fares when applied to causal models that represent situational narratives. As \cite{Halpern2005} put it: ``The best way to judge the adequacy of an approach are the intuitive appeal of the definitions and how well it deals with examples." While a useful definition of actual causation should certainly match the natural language usage of ``cause and effect" up to a point \citep{Paul2013, Halpern2016}, people's causal judgments are often influenced by contextual factors such as typicality and normality, as well as moral factors \citep{Hitchcock2009, HalpernHitch2015}.
This can lead to contradictory intuitions for the same causal model when placed within two different narratives \citep{HalpernHitch2015, Weslake2015-WESAPT}.

By contrast, our interest lies primarily in the principled analysis of actual causation in distributed dynamical systems, such as artificial neural networks, computers made of logic gates, or cellular automata, but also biological brain circuits, or gene regulatory networks (see Introduction, main text). The objective is to provide a complete, quantitative causal account of ``what caused what" within a transition between two consecutive system states (time $t-1$ and $t$). In this setting, causal models should always be interpreted mechanistically and an emphasis is put on explanatory power and consistency, rather than capturing people's intuitions (see also \cite{Paul2013}). 

This difference in focus also leads to a difference in scope: while the IIT-account can, in principle, evaluate causal links across any time interval $\Delta t$, i.e. within a system transition from time $t-\Delta t$ to time $t$, it does not, at present, address issues regarding causal transitivity, or causal links in non-Markovian systems. In a follow-up study, we plan to extend the proposed formalism based on system interventions and partitions to transitions across multiple time steps. For the moment, however, we restrict ourselves to examples without intermediary events between causes and effects. Doing so admittedly sidesteps many example cases from the philosophical literature, such as the late-preemption example of Billy and Suzy throwing rocks at a bottle with Suzy's rock hitting first \citep{Lewis2000}.
However, some prominent problem cases that are based on abstract narratives might simply not have an equivalent in physical systems of interacting mechanisms that take time into account explicitly.

For example, cases of ``preemption" that are frequently discussed in the philosophical literature often do not have an explicit temporal order embedded in the accompanying causal model  (see also Discussion section 4.5, main text). Take, for instance, the Billy-Suzy rock-throwing problem (see, e.g., \cite{Halpern2005}). If time is taken into account explicitly, Billy's rock either arrives at the bottle at the same time as Suzy's or later (see also \cite{Pearl2000}, Section 10.3 ``Temporal Preemption"). In the first case, the preemption example reduces to a case of symmetric overdetermination (main text, Fig. 5A). In the second case, Billy's rock could not have been a cause of the bottle shattering \textit{at time $t$} in the first place. A time-indexed solution to this example is provided in \citep{Halpern2005} in addition to the ``simple" solution that introduces the additional variables ``Suzy's throw hits" and ``Billy's throw hits" into the causal model. As noted by Halpern and Pearl, the time-indexed model avoids an objection to the ``simple" solution, where one could argue that the additional variables adjust the causal model in precisely the way to produce the desirable result.

Given the present focus on discrete dynamical systems, ``causation by omission" is also not an issue, since an element's state labels (e.g., ``0" or ``1", ``a" or ``b", etc.) have no implicit meaning and should thus play no role in the causal analysis. 

In the following, we provide some examples showing that the IIT-account provides a general, consistent, and principled approach to actual causation across a wide range of cases of direct, mechanistic influences. 


\section{Examples highlighting similarities and differences}
\label{examples}
This section summarizes the results of various accounts of actual causation (\cite{Halpern2005, Halpern2015, Weslake2015-WESAPT} and references therein) applied to several examples, including:

\begin{itemize}
    \item ``Majority Voting"
    \item ``Forest-Fire'' (also called ``Window"): $FF := L \land M$ (conjunctive) and $FF := L \lor M$ (disjunctive) 
    \item ``Loader": $D := (A \land B) \lor C$
    \item ``Complicated Voting": see Fig. 8 (main text)
    \item ``Command": $C := (M = 1) \lor (S = 1 \land M \neq 2)$
    \item ``Combination Lamp": $L:= (A=B) \lor (B=C) \lor (A=C)$
    \item ``Majority Voting with three choices"
\end{itemize}
taken from \citep{Schaffer2000,Livengood2013,Weslake2015-WESAPT, Halpern2015, Halpern2016}. 

For formal definitions we refer the reader to the original publications. Here we briefly summarize the algorithms behind the various HP-definitions within the present scope of direct influences without intermediary variables, i.e., given a transition $\transition$\footnote{ This reduces the subset $\Vec{Z} \subset \Vec{V}$ to $\Vec{X}$ in the formal HP-definitions given, e.g., in \citep{Halpern2015}.}. 

All HP-definitions provide three conditions AC1-3. AC1 states that both the occurrence $y_{t} \subseteq {v}_{t}$ and its candidate cause $x_{t-1} \subseteq {v}_{t-1}$ must actually happen. This is equivalent to our first causal principle of realization (main text). AC3 is a minimality condition stating that there must not be a subset of $x_{t-1}$ that also satisfies AC1 and AC2. While the same condition holds in the IIT-account, the HP-definitions generally allow for multiple causes of an occurrence, the modified HP-definition even for multiple causes with overlapping subsets of variables. By contrast, the IIT-account only allows for one occurrence $x_{t-1} \subseteq {v}_{t-1}$ to be an actual cause of $y_t$ according to its causal exclusion principle; $x_{t-1}$ can, however, be multi-variate. Finally, AC2 specifies the permissible contingencies under which the counterfactual dependence of $y_t$ on $x_{t-1}$ is tested by setting $X_{t-1}$ to a different state $x'_{t-1}\in \Omega_X$. The original, updated, and modified HP-definition are thus distinguished by their particular AC2. 

The original HP-definition allows setting the contingency $W_{t-1} \subseteq {V}_{t-1}\setminus X_{t-1}$ into any state $w_{t-1} \in \Omega_W$, as long as doing so would ensure that $Y_t$ still takes its actual state $y_t$ whenever $X_{t-1} = x_{t-1}$\footnote{Under the restricted scope of direct causal influences, the original HP-definition is equivalent to that of \cite{Hitchcock2001} and AC* of \cite{Woodward2003}, which, however do not explicitly include a minimality condition akin to AC3.}. 
The updated HP-definition, in addition, restricts the permissible contingency conditions $w_t$ to be such that $Y_t$ takes its actual state $y_t$ when $X_{t-1} = x_{t-1}$, even if any subset of $W_{t-1}$ is switched back to its actual state\footnote{Note that the formal definitions for the original and updated HP-definitions include more details, which, however, do not matter in the present context.}. This additional requirement allows the updated HP-definition to resolve the ``Loader" example (see below).  

\cite{Halpern2015} introduced a third variant, the ``modified" HP-definition. AC2 according to the modified HP-definition requires all variables in $W_{t-1} \subseteq {V}_{t-1}\setminus X_{t-1}$ to take their actual values. Causes identified by the modified HP-definition thus correspond to all minimal sets of variables that, if they were altered, would change the state of $Y_t$ from its actual state $y_t$ to another state $y'_t \in \Omega_Y$.

In section \ref{comp}, we will discuss how these proposed contingency conditions relate to the IIT-account. First, however, let us compare their respective results in the various problem cases listed above:

\subsection{Majority voting}
\label{Maj}
We start with the following simple voting scenario taken from \cite{Halpern2016} (Example 2.3.2) as it highlights the differences between the various HP-definitions and our approach: consider 11 voters deciding between two candidates ($A$ and $B$) based on a simple majority count. The causal model to this scenario corresponds to a linear threshold unit with $k = 6$ (Eqn. 17) in the main text, Fig. 6), that is, a Majority-gate with 11 inputs. As discussed in \citep{Halpern2016}, in the case of a 6-5 win for $A$ all HP-definitions agree that each of the voters in favor of $A$ counts as an actual cause of $A$'s win, meaning that there are 6 first-order actual causes. Note that each of these causes is necessary for $A$'s win, but only their conjunction is sufficient. As stated in Theorem 3.1 (main text), our proposed causal analysis identifies the entire set of 6 voters in favor of $A$ as the one higher-order actual cause, which arguably is more intuitive because it is the only occurrence at $t-1$ that is both necessary and sufficient for the outcome.

In the second situation all 11 voters decide in favor of candiate $A$---a case of overdetermination. Here the updated and original HP-definitions still declare each voter a first-order actual cause, while none is necessary or sufficient. Halpern's modified definition \citep{Halpern2016}, however, finds every subset of six voters a higher-order cause of $A$'s win (each minimally sufficient set). This seems like our proposed solution: one of the subsets of six voters is the actual cause of $A$'s win; which one, however, remains undetermined. This distinction drawn by the IIT-account is due to the causal exclusion principle (every occurrence can have at most one actual cause). To illustrate, declaring \textit{every} subset of six voters a cause would mean that each voter contributed to 252 causes, which would multiply their causal power beyond necessity and should thus be avoided. 

Note, moreover, that the modified HP account does not generally identify minimally sufficient sets as the actual cause (see \cite{Halpern2016}, section 2.6), even in similar majority voting examples. This becomes apparent if we consider other possible scenarios that could lead to $A$'s win, such as (i) a 10-1 win, or (ii) a 8-3 win, etc. The IIT-account consistently identifies a minimally sufficient set (in this case a subset of 6 voters) as the actual cause in any linear threshold example, as proven in Theorem 3.1 (main text). 
By contrast, the modified HP-definition gives different results in all cases. In (i) it would identify all sets of 5 $A$-voters and in (ii) all sets of 3 $A$-voters as causes of $A$'s win (corresponding to the number of votes that would have to be switched in order to change the outcome). This is despite the fact that, mechanistically, 6 votes are required for $A$'s win no matter the actual number.

\subsection{``Forest-Fire'': conjunctive and disjunctive}
\label{ANDOR}
In the ``Forest-Fire" example, a forest catches fire if lightning strikes ($L$) and/or an arsonist drops a match ($M$). 
The causal models pertaining to the ``Forest-Fire'' examples, both the conjunctive ($FF := L \land M$) and disjunctive ($FF := L \lor M$) cases, are in fact also linear threshold units with two inputs. Intuitively, the conjunctive case corresponds to an ``AND" logic-gate, while the disjunctive case corresponds to an ``OR" logic-gate (see Fig. 5A,B, main text).
Let us assume the standard setting where lightning strikes ($L=1$), the match is dropped ($M$ = 1) and, subsequently, the forest catches fire ($FF = 1$).
As shown in Fig. 5A and B, in the conjunctive case, our approach identifies the high-order occurrence $\{LM = 11\}$ as the actual cause, while in the disjunctive case it declares either of the first-order occurrences $L=1$ or $M=1$ as the actual cause with indeterminacy between the two. Note that these solutions correspond to the mechanistic logic structure of the two cases. 

By contrast, the various HP-definitions take both $L=1$ and $M=1$ as separate first-order causes in the conjunctive case, but explicitly deny that the higher-order conjunctive occurrence $\{LM = 11\}$ is a cause \citep{Halpern2016}. The original and updated definition give the same result in the disjunctive case, while the modified definition declares $\{LM = 11\}$ to be a high-order cause in the disjunctive version, arguably reversing the logical structures of the examples. In accordance with \cite{Halpern2015}, we see it as a feature of our account that it can distinguish the two scenarios. By contrast to the modified HP-definition, however, the IIT solution fits the logical setup (see also \cite{Datta2016}).


\subsection{``Loader"}
\label{Loader}
This example corresponds to the disjunction of conjunctions shown in Fig. 7: $D := (A \land B) \lor C$. One story accompanying this example is that ``a prisoner dies either if $A$ loads $B$'s gun and $B$ shoots, or if $C$ loads and shoots his gun. (\dots) $A$ loads $B$'s gun, $B$ does not shoot, but $C$ does load and shoot his gun, so that the prisoner dies" \citep{Hopkins2003, Halpern2015}. Intuition suggests that $A = 1$ should not be regarded as a cause of death $D = 1$ in the case that $B = 0$. Nevertheless, $A = 1$ does raise the probability for $D = 1$ by itself (when $B$ and $C$ are left undetermined, i.e., are causally marginalized)  (Fig. 7, main text). However, the one actual cause of $D=1$ according to the IIT-account is $C = 1$, in line with the updated and modified HP-definition, as well as \citep{Weslake2015-WESAPT}. The original HP-definition, however, would count $A = 1$ as a cause, which motivated the updated HP-definition in the first place \citep{Weslake2015-WESAPT, Halpern2015, Halpern2016}. Note that, in the overdetermined case with $A = 1$, $B = 1$, and $C = 1$, the IIT-account identifies either the higher-order occurrence $\{AB = 11\}$ or the first-order occurrence $\{C = 1\}$ as the actual cause with indeterminism between the two minimally sufficient possibilities. The original and updated HP-definition, as well as \citep{Weslake2015-WESAPT}, would call the first-order occurrences $\{A = 1\}$, $\{B = 1\}$, and $\{C = 1\}$ causes. The modified HP-definition would, in fact, declare the higher-order occurrences $\{AC = 11\}$ and $\{BC = 11\}$ as causes, although neither is necessary nor minimally sufficient for $\{D = 1\}$.

\subsection{``Complicated Voting"}
See Fig. 8, main text. This example is discussed in \cite{Halpern2015, Halpern2016}. If $A$ and $B$ agree, $F$ takes their value, if $B$, $C$, $D$, and $E$ agree, $F$ takes $A$'s value, otherwise majority decides. The actual transition is $\{ABCDE = 11000\} \prec \{F = 1\}$. As described in the main text (section 3.4), the IIT-account declares the occurrence $\{AB = 11\}$ the actual cause of $\{F = 1\}$, in line with intuition. None of the HP-definitions arrive at this result without introducing additional variables to the example. The original and updated HP-definitions would declare each of the inputs a first-order cause of $\{F = 1\}$, while the modified HP-definition identifies $\{A = 1\}$ as a first-order cause, but also the higher-order occurrences $\{BC = 10\}$, $\{BD = 10\}$, and $\{BE = 10\}$ (see \cite{Halpern2016}, which corrects the result in \cite{Halpern2015}). The IIT-account captures the mechanistic structure of the example and demonstrates that it is possible to resolve this problem case without the need of additional variables. Arguably, the additional variables introduced by \cite{Halpern2015, Halpern2016} only serve to build the desired result into the causal model (see also section \ref{addV} below). 

\subsection{``Command"}
An example in which the original \textit{and} updated HP-definition arguably do not align with intuition is the non-binary ``Command" example, a case of ``trumping-preemption" \citep{Schaffer2000,Weslake2015-WESAPT, Halpern2015, Halpern2016}. In this example, a major ($M$) and a sergeant ($S$) give orders to a corporal ($C$). $M$ and $S$ can take three values depending on what the major and sergeant do. They can each either do nothing ($0$), order the corporal to ``shoot" ($1$) or to ``halt" ($2$). The major's orders trump those of the sergeant: $C := (M = 1) \lor (S = 1 \land M \neq 2)$. The situation considered is that in which the major and the sergeant both order the corporal to shoot ($MS = 11$ leading to $C=1$). The original and updated HP-definition consider $S = 1$ a cause of $C = 1$ even though the major's orders ($M = 1$) trump those of the sergeant, which means that the corporal would have shot ($C = 1$) given $M = 1$ no matter what the sergeant would have done. By contrast, Halpern's modified HP-definition \citep{Halpern2015} and also \cite{Weslake2015-WESAPT} do not count $S=1$ as a cause of $C=1$. Our account agrees with the latter and declares $M = 1$ the sole actual cause of $C = 1$ with $\alpha^{max}_c = 1.17$. This example, moreover, demonstrates that the IIT-account can indeed be applied to causal networks with discrete, non-binary variables, identifying again the minimally sufficient set of variables as the actual cause. 

\subsection{``Combination Lamp"}
Another example that shows that the original and updated HP-definition have trouble with causal models including non-binary variables is the ``Combination Lamp" example proposed by \citep{Weslake2015-WESAPT}. A lamp ($L$) has three switches ($A$, $B$, $C$), which can be in one of three states ($-1$, $0$, or $1$). The lamp switches on iff at least two of the switches are in the same position: $L:= (A=B) \lor (B=C) \lor (A=C)$. The setting considered is $A = 1$, $B = -1$, and $C = -1$ leading to $L = 1$. As in the previous example, the original and updated HP-definition declare all switches causes of $L = 1$, while the modified HP-definition and Weslake's proposed account only regard $B = 1$ and $C = 1$ as (separate) causes. Here, our account specifically determines the joint occurrence $\{BC = 11\}$ as the actual cause of $L = 1$ with $\alpha^{max}_c = 1.95$. By comparison, the entire set of inputs $\{ABC = (-1)11\}$ has a lower $\alpha_c = 0.363$ and is thus excluded from being a cause. Notably, the individual elements all have $\alpha_c = 0$, as their individual state does not matter at all for the state of $L$. An intervention on only one input variable would not grant any control over the state of $L$; whether $L$ switches on or not would be entirely left to uncontrolled influences. The actual cause specified by the IIT-account, by contrast, identifies an ideal (minially sufficient) controller for $L = 1$.

\subsection{Majority voting given three options}
\label{maj3}
As a final non-binary example we consider a voting scenario suggested by \citep{Livengood2013}. For computational reasons, we will consider 15 voters that vote 13-2 in favor of candidate $A$ instead of the original example of 19 voters that vote 17-2. First, let us consider the binary case in which the voters choose between candidate $A$ and $B$. In line with the first example and Theorem 3.1 (main text), our account would consider one out of all minimally sufficient sets of 8 A-voters the actual cause of $A$'s win--- which one remains undetermined. By contrast, the original and updated HP-definition would consider each of the 15 voters for $A$ a cause of $A$'s win. 

We now add a third candidate $C$ and consider the vote 13-2-0 for $A$, $B$, and $C$, respectively. Throughout we assume that none of the candidates wins in case of a tie for the maximum number of votes. In this three-candidate case, the original and updated HP-definition also count the two voters who voted for $B$ as causes of $A$'s win. Note that the number of votes for $B$ may matter for $A$'s win in other configurations, for example, if the vote was 7-4-4, or 7-5-3. However, in the actual case of 13-2-0 the $A$-voters surpassed a simple majority and, by themselves, the $B$-votes each decrease $A$'s probability to win. Here, as in the binary version, the IIT-account identifies an undetermined set of 8 of the 13 A-voters as the actual cause with $\alpha^{max}_c = 1.82$, while $\alpha_c = 0$ for the B-voters. In the specific case of a 7-4-4 vote, the IIT-account identifies a set of $\{7 \times A,2 \times B,2 \times C\}$ voters as the actual cause ($\alpha^{max}_c = 1.82$). If any of the votes had been different, A might not have won, so only the whole set is sufficient here. To see why this set is minimally sufficient, consider 7-2-1, in this case, if the remaining 5 voters cast their ballots for $B$, then the result would be 7-7-1, a tie, in which case $A$ is not considered the winner. Note again, however, that the IIT-account of actual causation does not presuppose sufficiency as a requirement for causation. Instead, the minimally sufficient sets emerge as actual causes in these voting scenarios as a consequence of applying the five principles of causation.

Finally, how the modified HP-account would fair in the above example is not entirely clear. Halpern revised the original verdict from \citep{Halpern2015} in \citep{Halpern2016}, stating that the $B$ votes are parts of causes. Yet, it seems to us that the correct answer according to the modified HP-definition should be that any set of 6 voters for $A$ is a cause of $A's$ win. This is because changing 6 votes from $A$ to $B$ would change the outcome to a win for $B$ instead of $A$ (satisfying requirement AC2$^m$). Since the 6 A-voters are a subset of the cause specified in \citep{Halpern2016}, it excludes the larger set from being a cause according to AC3.

\subsection{A note on introducing additional variables and the value of a mechanistic scope}
\label{addV}
Frequently in the philosophical literature, solutions to causal dilemmas are proposed in which the original causal model is modified. To disambiguate the model, additional variables are introduced. This is warranted if the original model did not capture all relevant features of the accompanying narrative (e.g., modeling the Billy-Suzy rock-throwing example as a simple case of symmetric overdetermination ignores an important part of the narrative: that Suzy's rock actually hits the bottle while Billy's does not \citep{Halpern2005, Halpern2016}). In other cases, however, the practice of adding variables just deflects from the original problem. \cite{Halpern2015}, for example, argues in favor of a ``richer" model to the ``Combination Lamp" example for which the results of the original and updated HP-definitions seem more intuitive. The ``richer" model is interesting by itself, but it is a different physical system. If the mechanism $L$ is actually the one described in ``Combination Lamp" with direct inputs from $A$, $B$, and $C$, changing this causal model to include intermediary variables does not resolve the original problem. See also the proposal to introduce an additional variable to ``Loader" (\cite{Halpern2016}, section 2.8.1) to redeem the original HP-definition, or ``Complicated Voting" (\cite{Halpern2016}, section 4.1.4).
In the realm of discrete dynamical systems, the causal model corresponds to a physical system of interacting mechanisms and thus is what it is. Either an account can provide a satisfactory answer or it cannot. Restricting one's scope to physical systems of interacting mechanisms can thus shed light on the consistency and explanatory power of an actual causation account per se.

\section{Conceptual differences}
\label{comp}

In this section, we provide some additional discussion of three important conceptual differences between the IIT-account and existing accounts of actual causation. 

\subsection{Marginalizing across contingencies}
Accounts of actual causation typically identify actual causes by testing counterfactual dependence under a set of contingency conditions \citep{Hitchcock2001,Yablo2002,Woodward2003,Halpern2005,Hall2007,Halpern2015,Weslake2015-WESAPT}. Here we argue that testing the counterfactual dependence between $y_{t}$ and $x_{t-1}$ directly, by employing causal marginalization, is more general and eliminates the need for elaborate sets of contingency conditions.

A single variable $Y_{t}$ often depends on multiple input variables $\{X_{j,t-1}\}$, as specified by the causal network or, equivalently, $Y_{t}$'s structural equation. In such a setting, the counterfactual dependence of $y_{t}$ on a single-variable occurrence $x_{j,t-1}$ may only appear for specific contingencies (states of $W_{t-1} = V_{t-1}\setminus X_{j, t-1}$ other than its actual state). A simple ``but-for" test given the actual state $W_{t-1} = w_{t-1}$ would thus fail in this case. There are certain examples of this situation where intuition suggests that $X_{j, t-1}$ should be the actual cause (or at least part of the actual cause) of $Y_{t}$ (e.g., cases of overdetermination as in \ref{Maj} and \ref{ANDOR}). This motivates the assessment of counterfactual dependence under contingencies. However, there are other examples in which the intuition is that $X_{j, t-1}$ should not be a cause of $Y_{t}$ (e.g., ``Loader" \ref{Loader}), and these examples further motivate putting restrictions on the set of permissible contingencies under which one evaluates counterfactual dependence \citep{Weslake2015-WESAPT, Halpern2016}.  

For any account of actual causation based on contingency conditions, whether or not there is a counterfactual dependence between $y_{t}$ and $x_{t-1}$ is assessed based on a specific state of $W_{t-1} = V_{t-1}\setminus X_{t-1}$. Consequently, $x_{t-1}$ is not evaluated based on its own merits. This leads to counter-intuitive results that go against the logical structure of the example mechanisms (see section \ref{examples}). The modified HP-definition \citep{Halpern2015}, for instance, reverses the logical structure in the conjunctive and disjunctive ``Forest-Fire" example (identifying two separate causes in the conjunctive ``AND" case, and a joint cause in the disjunctive ``OR" case), while the original and updated HP-definitions cannot distinguish between the two cases (see also \cite{Datta2015}).



In the framework proposed here, we determine the actual cause of an occurrence $y_t$ starting from $y_t$'s cause-repertoire $\pi({V}_{t-1}|y_{t})$ (Eqn. 4, main text) over all variables $V_{t-1}$. For a single element occurrence $y_{t}$ in a deterministic causal model this interventional conditional probability distribution simply identifies all possible states of ${V}_{t-1}$ out of $\Omega_{{V_{t-1}}}$ that are consistent with $Y_{t} = y_{t}$. The cause-repertoire is akin to the ``redundancy range" of $y_{t}$ specified by \cite{Hitchcock2001}\footnote{Technically, the ``redundancy range" corresponds to the counterfactual states of all off-path variables (here ${V}_{t-1}\setminus X_{t-1}$) under which $Y_{t} = y_{t}$ holds, when $X_{t-1} = x_{t-1}$. The cause-repertoire also includes other possible states of $X_{t-1}$.}. The common idea is to fix $Y_{t}$ in its actual state $y_{t}$ and then look backward to identify the range of possible causes in ${V}_{t-1}$. 


From the cause-repertoire over $V_{t-1}$, we can obtain the cause-repertoire $\pi({X}_{t-1}|y_{t})$ of a particular subset of variables $X_{t-1} \subseteq {V}_{t-1}$ by causally marginalizing over all variables in ${V}_{t-1}\setminus X_{t-1}$. 
The cause-repertoire $\pi({X}_{t-1}|y_{t})$ corresponds to the contingency under which the (strength of) the counterfactual dependence between $y_{t}$ and $x_{t-1}$ is evaluated. However, instead of being fixed in any particular state, the variables $V_{t-1} \setminus X_{t-1}$ are causally marginalized.\footnote{The causal network $G_u$ itself is still conditioned on the state of the exogenous variables $U=u$. We thus explicitly distinguish between fixed background conditions ($U=u$) and other relevant variables (${V}_{t-1}\setminus X_{t-1}$) whose counterfactuals must be considered (see also \cite{McDermott2002}).} In this way, we ensure that the counterfactual dependence between $y_{t}$ and $x_{t-1}$ is independent of the state of ${V}_{t-1}\setminus X_{t-1}$.\footnote{Marginalizing across all possible states of ${V}_{t-1}\setminus X_{t-1}$ bears some resemblance to sufficiency accounts of actual causation that require that $Y_{t} = y_{t}$ follows $X_{t-1} = x_{t-1}$, under every possible state of ${V}_{t-1}\setminus X_{t-1}$ \citep{McDermott2002, Woodward2003, Weslake2015-WESAPT}. Such sufficiency accounts, however, do not test for counterfactual dependence between $y_{t}$ and $x_{t-1}$ or have to be extended by a set of contingency conditions \citep{Weslake2015-WESAPT} akin to those of \citep{Hitchcock2001, Halpern2001, Halpern2005}. This again is a consequence of assuming fixed states for ${V}_{t-1}\setminus X_{t-1}$. The IIT-account typically identifies a minimally sufficient set of variables as the actual cause of a singleton occurrence $y_{t}$ in deterministic causal models. However, the approach of causally marginalizing across the states of ${V}_{t-1}\setminus X_{t-1}$ naturally extends to the probabilistic case and provides a measure of the causal strength of the counterfactual dependence between $y_{t}$ and $x_{t-1}$.} 

\subsection{Quantifying counterfactual dependence} In the IIT-account of actual causation, we quantify the counterfactual dependence between ${x}_{t-1}$ and $y_{t}$ by comparing how an occurrence $y_t$ increases the likelihood of the past state $x_{t-1}$, compared to all other possible states of $Y_t$. This is captured by the cause ratio $\rho_c(x_{t-1}, y_{t})$ (Eqn. 12, main text)\footnote{Note that $\rho_c(x_{t-1}, y_{t})$ is identical to $\rho_e(x_{t-1}, y_{t})$, if $y_t$  is a single-variable occurrence. While we technically take the perspective of $y_{t}$ to identify its actual cause and compare $\pi({x}_{t-1}|y_{t})$ to $\pi({x}_{t-1})$, this is equivalent to comparing $\pi({y}_{t}|x_{t-1})$ to $\pi({y}_{t})$.}. For a single-variable occurrence $\rho_c(x_{t-1}, y_{t}) = \alpha_c(x_{t-1}, y_{t})$, which determines the strength of the counterfactual causal link between ${x}_{t-1}$ and $y_{t}$. By contrast, most accounts of actual causation take counterfactual dependence to be an ``all-or-nothing concept" \citep{Halpern2016}. An occurrence $x_{t-1}$ is typically considered a cause of ${y}_{t}$ if there exists one specific counterfactual $x'_{t-1} \in \Omega_{X_t \setminus x_t}$ for which $Y_{t} \neq y_{t}$. In the case of binary variables, this condition implies that $\rho_c(x_{t-1}, y_{t}) > 0$. However, this does not have to be the case for non-binary variables, where the advantages of comparing $x_{t-1}$ to an average of all its possible states become apparent (see e.g., \ref{maj3}).





There are also probabilistic, counterfactual accounts of actual causation, which are typically based on the notion that causes should raise the probability of their effects, such as the requirement that $p(e \mid c) > p(e)$ \citep{Good1961, Suppes1970, Eells1983, Lewis1986}. In the present context of causal networks, or equivalently, structural equation models, the same condition is required to hold under interventions, that is, $p(e \mid \Do(C = c)) > p(e)$ \citep{Pearl2000, Pearl2009}. However, just as in the case of deterministic counterfactual dependence, $p(e \mid \Do(C = c)) > p(e)$ might only hold under certain contingencies. In fact, it has been argued that probability raising per se might neither be necessary nor sufficient for actual causation \citep{Hitchcock2004, Fenton-Glynn2017}.

\cite{Twardy2011} and \cite{Fenton-Glynn2017} recently proposed extensions of contingency-based accounts of actual causation to probabilistic causal models. While \cite{Twardy2011} and \cite{Fenton-Glynn2017} employ probabilities, they still treat actual causation as an ``all-or-nothing" concept. Causes are generally considered to be single variables and identified based on contingency conditions. Therefore, these probabilistic accounts inherit the merits and problems of the deterministic accounts of actual causation they are based on \citep{Hitchcock2001, Halpern2001}. All reservations about accounts of actual causation based on contingency conditions raised in the previous section still apply. In addition, the claim that a cause does not necessarily raise the probability of its effect only holds true if the remaining variables $W_{t-1} = {V}_{t-1}\setminus X_{t-1}$ are fixed in a particular state. The IIT-account instead leaves the state of $W_{t-1}$ undetermined to avoid any dependencies on $w_{t-1}$ while evaluating $\rho_c(x_{t-1}, y_{t})$ (and thus $p({y}_{t}|\Do(X_{t-1} = x_{t-1}))$). In this case, $\rho_c(x_{t-1}, y_{t}) > 0$ is a necessary, but not sufficient, requirement for $x_{t-1}$ to be a cause of $y_t$, meaning that $x_{t-1}$, by itself, must raise the probability of $y_t$. 

Another advantage of the IIT-account is that $\rho_c(x_{t-1}, y_{t})$ already provides an initial measure of causal strength that quantifies the causal link between $x_{t-1}$ and $y_{t}$ and can be compared across all possible candidate causes\footnote{For multi-variable occurrences $y_{t}$, it is moreover necessary to test for irreducibility. The actual strength of the causal link is thus quantified by $\alpha_c(x_{t-1}, y_{t})$. See main text.}. Probabilistic accounts that identify actual causes based on specific contingencies $w'_{t-1}$, such as \citep{Twardy2011} and \citep{Fenton-Glynn2017}, do not provide a unique quantifier for the causal link between $x_{t-1}$ and $y_{t}$, since the difference between $p({y}_{t}|\Do(X_{t-1} = x_{t-1}))$ and $p({y}_{t}|\Do(X_{t-1} = x'_{t-1}))$ depends on $w'_{t-1}$ and there may be multiple permissible $w'_{t-1}$.

\subsection{Identifying the actual cause}
In general, it is possible that $y_t$ is counterfactually dependent on more than one variable (or even more than one set of variables). Some accounts of actual causation only allow for singleton causes and effects \citep{Weslake2015-WESAPT}, excluding multi-variate causes by definition. Under the HP-definition, supersets of actual causes are excluded from being causes themselves by a minimality clause (``AC3") \citep{Halpern2005, Halpern2015}. Nevertheless, these accounts still allow for $y_{t}$ to have potentially many different actual causes, as long as they fulfill the imposed contingency conditions. Note that this limits the explanatory power of individual causes considerably, as it remains unclear whether an identified (singleton) cause was sufficient or necessary, both or neither. Even the complete set of causes often does not illuminate the mechanistic connection between causes and effect (see, e.g., \ref{ANDOR}).

The IIT-account follows a causal exclusion principle, allowing only one cause for a given occurrence $y_{t}$ out of all possible occurrences $x_{t-1}\subseteq {v}_{t-1}$. 
To identify the \textit{actual} cause of $y_{t}$, we measure and compare the strength $\alpha_c(x_{t-1}, y_{t})$ of all possible causal links with $y_{t}$. The actual cause $x_{t-1}^*$ of $y_t$ is the occurrence with maximal causal strength ($\alpha_c(x_{t-1}^*, y_t) = \alpha^\m_c(y_{t})$). Ties in causal strength can occur for two reasons: first, adding inessential elements to $x^*_{t-1}$ has no effect on $\alpha^\m_c(y_{t})$. For this reason, we include a minimality condition similar to AC3: any superset of $x^*_{t-1}$ with the same $\alpha^\m_c(y_{t})$ is excluded from being a possible cause. Second, ties can also occur under symmetries in the causal model, in cases of true causal overdetermination. Upholding the causal exclusion principle, such degenerate cases are resolved by noting that the \textit{one} actual cause $x^*_{t-1}$ remains undetermined between all minimal $x_{t-1}$ with $\alpha_c(x_{t-1}, y_{t}) = \alpha^\m_c(y_{t})$. As demonstrated in the above example cases (see Section \ref{examples}), the actual cause identified by the IIT-account consistently captures the mechanistic connection between cause and effect.

\bibliographystyle{imsart-nameyear}
\bibliography{AC_bibtex}


\begin{frontmatter}

\runauthor{Albantakis et al.}
\runtitle{Supplementary Methods}
\end{frontmatter}

\title{Irreducibility of the causal account}
\vspace{0.2in}

Similar to the notion of system-level integration in integrated information theory (IIT) \citep{Oizumi2014, Albantakis2015}, the principle of integration can also be applied to the causal account as a whole, not only to individual causal links. The causal account of a particular transition $\transition$ of the dynamical causal network $G_u$ is defined as the set of all causal links within the transition (Definition 2.4, main text). 


In the following we define the quantity $\A(\transition)$, which measures to what extent the transition $\transition$ is irreducible to its parts. Moreover, we introduce $\A_e(\transition)$, which measures the irreducibility of $v_{t-1}$ and its set of ``effect" causal links $\{x_{t-1} \rightarrow y_t\} \in \C(\transition)$, and $\A_c(\transition)$, which measures the irreducibility of $v_{t}$ and its set of ``cause" causal links $\{x_{t-1} \leftarrow y_t\} \in \C(\transition)$.
In this way, we can

\begin{figure}
\includegraphics[width = 5in]{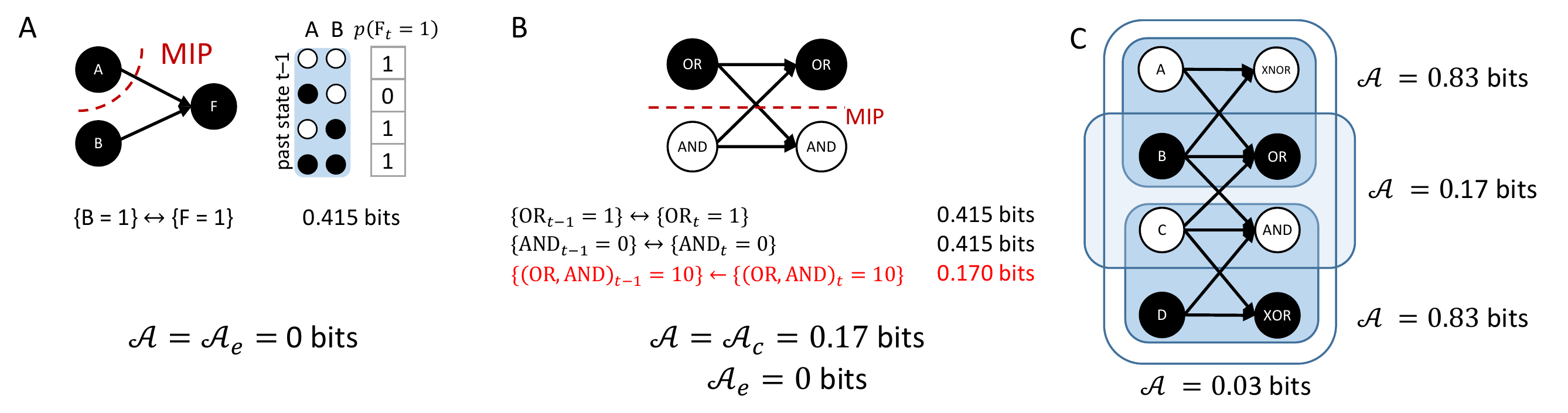}
\caption{\textbf{Reducible and irreducible causal accounts.} (A) ``Prevention" example (see Fig. 7D, main text). $\A = 0$ bits as $\{A = 1\}$ does not contribute to any causal links. (B) Irreducible transition (see Fig. 6, main text). A partition of the transition along the $\mip$ destroys the 2\textsuperscript{nd}-order causal link, leading to $\A = 0.17$ bits. (C) In larger systems, $\A$ can be used to identify (sub)transitions with highly entangled causes and effect. While the causes and effects in the full transition are only weakly entangled with $\A = 0.03$ bits, the top and bottom (sub)transitions are irreducible with $\A = 0.83$ bits.}
\label{figS1}
\end{figure}

\begin{itemize}
    \item identify irrelevant variables within a causal account that do not contribute to any causal link (Fig. \ref{figS1}A),
    \item evaluate how entangled the sets of causes and effects are within a transition $\transition$ (Fig. \ref{figS1}B), and
    \item compare $\A$ values between (sub)transitions, in order to identify clusters of variables whose causes and effects are highly entangled, or only minimally connected (Fig. \ref{figS1}C).
\end{itemize}

We can assess the irreducibility of $v_{t-1}$ and its set of ``effect" causal links $\{x_{t-1} \rightarrow y_t\} \in \C(\transition)$ in parallel to $\alpha_e(x_{t-1}, y_t)$, by testing all possible partitions $\Psi(v_{t-1}, V_t)$  (Eqn. 7, main text).  This means that, the transition $\transition$ is partitioned into independent parts in the same manner that an occurrence $x_{t-1}$ is partitioned when assessing $\alpha_e(x_{t-1}, y_t)$. We then define the irreducibility of $v_{t-1}$ as the difference in the total strength of actual effects (causal links of the form $x_{t-1}\rightarrow y_t$) in the complete causal account $\C$ compared to the causal account under the $\mip$, which again denotes the partition in $\Psi(v_{t-1}, V_t)$ that makes the least difference to $\C$:
\begin{equation}
\label{eqn10}
\A_e(\transition) = \sum_{x \rightarrow y \in \C} \left(\alpha_e^{\m}(x)\right) - \sum_{x \rightarrow y \in \C_{\mip}} \left(\alpha_e^{\m}(x)_{\mip}\right) 
\end{equation}
In the same way, the irreducibility of $v_t$ and its set of causal links $\{x_{t-1} \leftarrow y_t\} \in \C(\transition)$ is defined as the difference in the total strength of actual causes (causal links of the form $x_{t-1}\leftarrow y_t$) in the causal account $\C$ compared to the causal account under the $\mip$:
\begin{equation}
\label{eqn11}
\A_c(\transition) = \sum_{x \leftarrow y \in \C} \left(\alpha_c^{\m}(y)\right) - \sum_{x \leftarrow y \in \C_{\mip}} \left(\alpha_c^{\m}(y)_{\mip}\right) 
\end{equation}
where the $\mip$ is again the partition that makes the least difference out of all possible partitions $\Psi(V_{t-1}, v_t)$ (Eqn. 9, main text). This means that the transition $\transition$ is partitioned into independent parts in the same manner that an occurrence $y_{t}$ is partitioned when assessing $\alpha_c(x_{t-1}, y_t)$. The irreducibility of a single-variable $v_{t-1}$ or $v_t$ reduces to $\alpha_e^{\m}$ of its one actual effect $y_t$, or $\alpha_c^{\m}$ of its one actual cause $x_{t-1}$, respectively.

By considering the union of possible partitions, $\Psi(\transition) = \Psi(v_{t-1}, V_t) \cup \Psi(V_{t-1}, v_t)$, we can moreover assess the overall irreducibility of the transition $\transition$. A transition $\transition$ is reducible if there is a partition $\psi \in \Psi(\transition)$ such that the total strength of causal links in $\C(\transition)$ is unaffected by the partition. Based on this notion we define the irreducibility of a transition $\transition$ as:
\begin{equation}
\A(\transition) = \sum \alpha^{\m}(\C) - \sum \alpha^{\m}(\C_{\mip}),
\end{equation}
where
\begin{equation*}
\sum \alpha^{\m}(\C) = \sum_{x \rightarrow y \in \C} \left(\alpha_e^{\m}(x)\right) + \sum_{x \leftarrow y \in \C} \left(\alpha_c^{\m}(y)\right)
\end{equation*}
is a summation over the strength of all causal links in the causal account $\C(\transition)$, and the same for the partitioned causal account $\C_{\mip}$.

Fig. \ref{figS1}A shows the ``Prevention" example of Fig. 7D, main text. $\{A = 1\}$ has no effect and is no cause in this transition. Replacing $\{A = 1\}$ with an average over all its possible states does not make a difference to the causal account and thus $\A(\transition) = 0$ in this case. 
Fig. \ref{figS1}B shows the
causal account $\C_{\mip}$ of the transition $\transition$ with $v_{t-1} = v_t = \{\text{OR}, \text{AND} = 10\}$ under its $\mip$ into $m = 2$ parts
with $\A(\transition) = 0.17$. This is the causal strength that would be lost if we treated $\transition$ as two separate transitions $\{\text{OR}_{t-1} = 1\} \prec \{\text{OR}_t = 1\}$ and $\{\text{AND}_{t-1} = 0\} \prec \{\text{AND}_t = 0\}$ instead of a single one within $G_u$.

The irreducibility $\A(\transition)$ provides a measure of how causally ``entangled" the variables $V$ are during the transition $\transition$. In a larger system, we can measure and compare the $\A$ values of multiple (sub)transitions. In Fig. \ref{figS1}C, for example, the causes and effects of the full transition are only weakly entangled ($\A = 0.03$ bits), while the transitions involving the four upper or lower variables, respectively, are much more irreducible ($\A = 0.83$ bits). In this way, 
$\A(\transition)$ may be a useful quantity when evaluating more parsimonious causal explanations against the complete causal account of the full transition.

\bibliographystyle{imsart-nameyear}
\bibliography{AC_bibtex}


\begin{frontmatter}

\runauthor{Albantakis et al.}
\runtitle{Supplementary Proofs}
\end{frontmatter}

\section*{Supplementary Proof 1} 
\stepcounter{section}

The first theorem describes the actual causes and effects for an observation of a linear threshold unit (LTU) $V_t = \{Y_t\}$ with $n$ inputs and threshold $k$, and its inputs $V_{t-1}$. First, a series of lemmas are demonstrated based on the transition probabilities $q_{c,j}$: if $X_{t-1} = x_{t-1} \subseteq V_{t-1} =v_{t-1}$ is an occurrence with size $|X_{t-1}| = c$ and $j$ of the $c$ elements in $X_{t-1}$ are in the `ON' state $(\sum_{x \in x_{t-1}} x = j)$, then 

\[ q_{c, j} = p(Y_t = 1|X_{t-1} = x_{t-1}) = \begin{cases} 
\sum\limits_{i = k - j}^{n - c} \frac{1}{2^{n-c}}\binom{n-c}{i} ~ &\text{if} ~ j \leq k ~ \text{and} ~ j > k - (n - c) \\
1 \quad &\text{if} ~ j \geq k \\
0 \quad &\text{if} ~ j < k - (n - c) \\
\end{cases} \]
First we demonstrate that the probabilities $q_{c, j}$ are non-decreasing as the number of `ON' inputs $j$ increases for a fixed size of occurrence $c$, and that there is a specific range of values of $j$ and $c$ such that the probabilities are strictly increasing. 
\begin{lemma}
\label{lemma1}
$q_{c, j} \geq q_{c, j-1}$ with $q_{c, j} = q_{c, j-1}$ iff $j > k$ or $j < k - (n - c)$. 
\end{lemma}

\begin{proof}

If $j > k$ then 
\[ q_{c, j} = q_{c, j-1} = 1. \]
If $j < k - (n - c)$ then
\[ q_{c, j} = q_{c, j-1} = 0. \]
If $k - (n - c) \leq j \leq k$ then
\[ q_{c, j} = \frac{1}{2^{n-c}}\sum_{i = k - j}^{n - c}\binom{n-c}{i} = \frac{1}{2^{n-c}}\sum_{i = k - (j - 1)}^{n - c}\binom{n-c}{i} - \frac{1}{2^{n-c}} = q_{c, j-1} + \frac{1}{2^{n-c}} > q_{c, j-1} \]

\end{proof}
Next we demonstrate two results relating the transition probabilities between occurrences of different sizes, 
\begin{lemma}
\label{lemma2}
$q_{c, j} = \frac{1}{2}\left(q_{c+1, j} + q_{c+1, j+1}\right)$ for $1 \leq c < n$ and $0 \leq j \leq c$.
\end{lemma}

\begin{proof}

If $j \geq k$ then
\[q_{c, j} = q_{c + 1, j} = q_{c+1, j+1} = 1, \]
so 
\[q_{c, j} = \frac{1}{2}\left(q_{c+1, j} + q_{c+1, j+1}\right) = 1. \] 
If $j = k - 1$ then
\[q_{c + 1, j + 1} = 1, \]
and 
\begin{eqnarray*}
q_{c, j} = q_{c, k - 1} &=& \frac{1}{2^{n - c}} \sum_{i = 1}^{n-c} \binom{n - c}{i} \\
&=& \frac{1}{2^{n-c}}\left(1 + \sum_{i = 1}^{n - (c + 1)}\binom{n - (c + 1)}{i - 1} + \binom{n - (c + 1)}{i}\right) \\
&=& \frac{1}{2}\left(\frac{1}{2^{n - (c+1)}}\sum_{i = 1}^{n - (c + 1)} \binom{n - (c+1)}{i} + \frac{1}{2^{n - (c+1)}}\sum_{i=0}^{n - (c+1)} \binom{n - (c+1)}{i}\right) \\
&=& \frac{1}{2}(q_{c+1, j} + 1) \\
&=& \frac{1}{2}(q_{c+1, j} + q_{c+1, j+1}) \\
\end{eqnarray*} 
If $j < k - (n - c)$ then 
\[q_{c, j} = q_{c + 1, j} = q_{c+1, j+1} = 0, \]
so 
\[q_{c, j} = \frac{1}{2}\left(q_{c+1, j} + q_{c+1, j+1}\right) = 0. \] 
If $j = k - (n - c)$ then 
\[q_{c + 1, j} = 0, \]
and
\[q_{c, j} = \frac{1}{2^{n - c}} \sum_{i = n - c}^{n-c} \binom{n - c}{i} = \frac{1}{2^{n-c}} \sum_{i = n - (c + 1)}^{n-(c + 1)} \binom{n - (c + 1)}{i} = \frac{1}{2}(q_{c+1, j+1} + q_{c+1, j}). \] 
Finally, if $k - (n - c) + 1 < j < k - 1$ then
\begin{eqnarray*}
q_{c, j} &=& \frac{1}{2^{n - c}} \sum_{i = k-j}^{n - c}\binom{n-c}{i} \\
 &=& \frac{1}{2^{n-c}} \left( 1 + \sum_{i = k-j}^{n - (c + 1)}\binom{n-c}{i} \right) \\
 &=& \frac{1}{2^{n-c}} \left( 1 + \sum_{i = k-j}^{n - (c + 1)}\binom{n-(c+1)}{i} + \sum_{i = k-j}^{n - (c + 1)}\binom{n-(c+1)}{i-1} \right) \\
 &=& \frac{1}{2^{n-c}} \left( \sum_{i = k-j}^{n - (c + 1)}\binom{n-(c+1)}{i} + \left(1 + \sum_{i = k-j}^{n - (c + 1)}\binom{n-(c+1)}{i-1} \right)\right) \\
 &=& \frac{1}{2^{n - c}} \left( \sum_{i = k-j}^{n - (c + 1)}\binom{n-(c+1)}{i} + \sum_{i = k- (j + 1)}^{n - (c + 1)}\binom{n-(c+1)}{i} \right) \\
 &=& \frac{1}{2}\left(q_{c + 1, j} + q_{c + 1, j + 1}\right) 
\end{eqnarray*}
\end{proof}

\begin{lemma}
\label{lemma3}
If $c < k$ then $q_{c, c} < q_{c+1, c+1}$
\end{lemma}

\begin{proof}

\begin{eqnarray*}
q_{c, c} &=& \frac{1}{2}\left(q_{c+1, c} + q_{c+1, c+1}\right) ~~ \text{(Lemma 1.2)} \\
 &<& \frac{1}{2}\left(q_{c+1, c+1} + q_{c+1, c+1}\right) ~~ \text{(Lemma 1.1)} \\
 &=& q_{c+1, c+1}
\end{eqnarray*}
\end{proof}

Finally, we consider a quantity $Q(c)$, the sum of $q$ over all possible states for an occurrence of size $c$. The value $Q(c)$ acts as a normalization term when calculating the cause repertoire of occurrence $\{Y_t = 1\}$. Here we demonstrate a relationship between these normalization terms across occurrences of different sizes,
\begin{lemma}
\label{lemma4}
Let $Q(c) = \sum\limits_{j = 0}^c\binom{c}{j}q_{c, j}$ then $Q(c) = \frac{1}{2}Q(c + 1)$
\end{lemma}

\begin{proof}
\begin{eqnarray*}
Q(c) &=& \sum_{j = 0}^c\binom{c}{j}q_{c, j} \\
&=& \frac{1}{2}\sum_{j = 0}^c \binom{c}{j}\left(q_{c + 1, j} + q_{c+1, j+1}\right) \\
&=& \frac{1}{2}\left( \sum_{j = 1}^c \binom{c}{j-1}q_{c+1, j} + \sum_{j = 0}^c \binom{c}{j}q_{c+1,j}\right) \\
&=& \frac{1}{2}\left(q_{c+1, c+1} + q_{c+1, 0} + \sum_{j = 1}^c \binom{c+1}{j}q_{c+1, j}\right) \\
&=& \frac{1}{2}\left(\sum_{j=0}^{c+1} \binom{c+1}{j}q_{c+1,j} \right) \\
&=& \frac{1}{2}Q(c+1) 
\end{eqnarray*}
\end{proof}

Using the above lemmas, we are now in a position to prove the actual causes and actual effects in the causal account of a single LTU in the `ON' state. The causal account for a LTU in the `OFF' state follows by symmetry. 
\begin{theorem}
\label{thm1}

Consider a dynamical causal network $G_u$ such that $V_t = \{Y_t\}$ is a linear threshold unit with $n$ inputs and threshold $k \leq n$, and $V_{t-1}$ is the set of $n$ inputs to $Y_t$. For a transition $v_{t-1} \prec v_{t-1}$, with $y_t = 1$ and $\sum v_{t-1} \geq k$, the following holds: 

\begin{enumerate}
\item  The actual cause of $\{Y_t = 1\}$ is an occurrence $\{X_{t-1} = x_{t-1}\}$ with $|x_{t-1}| = k$ and $\min(x_{t-1}) = 1$. Furthermore, the causal strength of the link is

\[\alpha_c^{\max}(y_t) = k - \log_2\left(\sum_{j = 0}^k q_{k, j}\right) > 0.\]

\item If $\min(x_{t-1}) = 1$ and $|x_{t-1}| \leq k$ then the actual effect of $\{X_{t-1} = x_{t-1}\}$ is $\{Y_t = 1\}$ with causal strength 
\[ \alpha_e(x_{t-1}, y_t) = \log_2\left(\frac{q_{c, c}}{q_{c-1, c-1}}\right) > 0, \] 
otherwise $\{X_{t-1} = x_{t-1}\}$ is reducible ($\alpha^{\max}_e(x_{t-1}) = 0$).

\end{enumerate}

\end{theorem}

\begin{proof}
\label{proof1}
 
\textbf{Part 1:} Consider an occurrence $\{X_{t-1} = x_{t-1}\}$ such that $|x_{t-1}| = c \leq n$ and $\sum\limits_{x \in x_{t-1}} x = j$, the probability of $x_{t-1}$ in the cause-repertoire of $y_t$ is

\[ \pi(x_{t-1}|y_t) = \frac{q_{c, j}}{Q(c)}. \]
Since $Y_t$ is a first-order occurrence, there is only one possible partition, and the causal strength of a potential link is thus

\[ \alpha_c(x_{t-1}, y_t) = \rho_c(x_{t-1}, y_t) = \log_2\left(\frac{\pi(x_{t-1}|y_t)}{\pi(x_{t-1})}\right) = \log_2\left(\frac{2^cq_{c, j}}{Q(c)}\right). \]
For a fixed value of $c$, the maximum value of causal strength occurs at $j = c$ (since adding `ON' elements can only increase q(c, j), Lemma \ref{lemma1}), 

\[\max_{|x_{t-1}|=c} \alpha_c(x_{t-1}, y_t) = \max_j \log_2\left(\frac{2^c q_{c, j}}{Q(c)}\right) = \log_2\left(\frac{2^c q_{c, c}}{Q(c)}\right) \]
Applying Lemma \ref{lemma3} and Lemma \ref{lemma4}, we see that across different values of $c$, this maximum is increasing for $0 < c < k$, 

\begin{eqnarray*}
\max_{|x_{t-1}|=c + 1} \alpha_c(x_{t-1}, y_t) - \max_{|x_{t-1}|=c} \alpha_c(x_{t-1}, y_t) &=& \log_2\left(\frac{2^{c+1} q_{c+1, c+1}}{Q(c+1)}\right) - \log_2\left(\frac{2^c q_{c, c}}{Q(c)}\right) \\
&=& \log_2\left(\frac{2^{c+1} q_{c+1, c+1} Q(c)}{2^c q_{c, c} Q(c+1)}\right) \\
&=& \log_2\left(\frac{q_{c+1, c+1}}{q_{c, c}}\right) \\
&>& 0, 
\end{eqnarray*}
and that for $k \leq c$ the causal strength is constant, 

\begin{eqnarray*}
\max_{|x_{t-1}|=c + 1} \alpha_c(x_{t-1}, y_t) - \max_{|x_{t-1}|=c} \alpha_c(x_{t-1}, y_t) &=& \log_2\left(\frac{2^{c+1} q_{c+1, c+1}}{Q(c+1)}\right) - \log_2\left(\frac{2^c q_{c, c}}{Q(c)}\right) \\
&=& \log_2\left(\frac{q_{c+1, c+1}}{q_{c, c}}\right) \\
&=& \log_2\left(\frac{1}{1}\right) = 0. 
\end{eqnarray*}
By setting $c = j \geq k$ we find the maximum causal strength is 
\[ \alpha_c^{\max}(y_t) = \log_2\left(\frac{2^c q_{c, c}}{Q(c)}\right) = \log_2\left(\frac{2^k}{Q(k)}\right) = k - \log_2\left(\sum_{j = 0}^k q_{k, j}\right) > 0. \]

Any occurrence $x_{t-1}$ with $j \geq k$ has maximal causal strength and satisfies condition (1) of being an actual cause, 
\[ \alpha_c(x_{t-1}, y_t) = \log_2\left(\frac{2^{c} q_{c, j}}{Q(c)}\right) = \log_2\left(\frac{2^{k}}{Q(k)}\right) = \alpha_c^{\max}(y_t). \] 
If $c \geq k$, then there exists a subset $x'_{t-1} \subset x_{t-1}$ with $j' \geq k$ and $c' < c$ such that $x'_{t-1}$ also satisfies condition (1) and thus $x_{t-1}$ does not satisfy condition (2). However, if $j = c = k$, then any subset $x'_{t-1}$ of $x_{t-1}$ has $j' < k$, so 
\[ \alpha_c(x'_{t-1}, y_t) = \log_2\left(\frac{2^{c'} q_{c', j'}}{Q(c')}\right) < \log_2\left(\frac{2^c}{Q(c)}\right) = \alpha(x_{t-1}, y_t),\] 
and thus $x_{t-1}$ satisfies condition (2). Therefore, we have that the actual cause of $y_t$ is an occurrence $x_{t-1}$ such that $|x_{t-1}| = k$ and $\min x_{t-1} = 1$,
\[x^*(y_t) = \{x_{t-1} \subset v_{t-1} \mid \min x_{t-1} = 1 \text{, and } |x_{t-1}| = k\}. \]

\textbf{Part 2:} Again, consider occurrences $X_{t-1} = x_{t-1}$ with $|x_{t-1}| = c$ and $\sum\limits_{x \in x_{t-1}} x = j$. The probability of $y_t$ in the effect repertoire of such an occurrence is 

\[ \pi(y_t|x_{t-1}) = q_{c, j} = \begin{cases} 
\sum\limits_{i = k - j}^{n - c} \frac{1}{2^{n-c}}\binom{n-c}{i} ~ &\text{if} ~ j \leq k ~ \text{and} ~ j > k - (n - c) \\
1 \quad &\text{if} ~ j \geq k \\
0 \quad &\text{if} ~ j < k - (n - c) \\
\end{cases} \]
Since there is only one element in $v_t$, the only question is whether or not $x_{t-1}$ is reducible. If it is reducible, it has no actual effect, otherwise its actual effect must be $y_t$. First, if $j < c$, then $\exists ~ x = 0 \in x_{t-1}$ and we can define a partition $\psi = \left\{ \{(x_{t-1} - x), y_t\}, \{x, \varnothing\}\right\}$ such that

\[ \pi(y_t|x_{t-1})_\psi = \pi(y_t | (x_{t-1} - x)) \times \pi(\varnothing | x) = \pi(y_t | (x_{t-1} - x)) = q_{c - 1, j} \]
and 

\[ \alpha_e(x_{t-1}, y_t) \leq \log_2\left(\frac{\pi(y_t|x_{t-1})}{\pi(y_t|x_{t-1})_\psi}\right) = \log_2\left(\frac{q_{c, j}}{q_{c - 1, j}}\right) \leq 0 \quad \text{(Lemma1.1/1.2)}, \]
so $x_{t-1}$ is reducible. Next we consider the case where $j = c$ but $c > k$. In this case we define a partition $\psi = \left\{ \{(x_{t-1} - x), y_t\}, \{x, \varnothing\}\right\}$ (where $x \in x_{t-1}$ is any element), such that 

\[ \pi(y_t|x_{t-1})_\psi = \pi(y_t | (x_{t-1} - x)) \times \pi(\varnothing | x) = \pi(y_t | (x_{t-1} - x)) = q_{c - 1, c - 1}, \]
and since $c > k$,

\[ \alpha_e(x_{t-1}, y_t) \leq \log_2\left(\frac{\pi(y_t|x_{t-1})}{\pi(y_t|x_{t-1})_\psi}\right) = \log_2\left(\frac{q_{c, c}}{q_{c - 1, c - 1}}\right) = \log_2\left(\frac{1}{1}\right) = 0, \]
so $x_{t-1}$ is again reducible. Lastly, we show that for $j = c$ and $c \leq k$, that $x_{t-1}$ is irreducible with actual effect $\{Y_t = 1\}$. All possible partitions of the pair of occurrences can be formulated as $\psi = \left\{ \{(x_{t-1} - x), y_t\}, \{x, \varnothing\}\right\}$ (where $x \subseteq x_{t-1}$ with $|x| = d > 0$), such that 

\[ \pi(y_t|x_{t-1})_\psi = \pi(y_t | (x_{t-1} - x)) \times \pi(\varnothing | x) = \pi(y_t | (x_{t-1} - x)) = q_{c - d, c - d}, \]
and

\[ \alpha_e(x_{t-1}, y_t) = \min_{\psi} \log_2\left(\frac{\pi(y_t|x_{t-1})}{\pi(y_t|x_{t-1})_\psi}\right) = \min_d \log_2\left(\frac{q_{c, c}}{q_{c - d, c - d}}\right). \]
The minimum information partition occurs when $d = 1$ (Lemma \ref{lemma3}) and thus $\{X_{t-1} = x_{t-1}\}$ is irreducible with actual effect $\{Y_t = 1\}$ and causal strength 

\[ \alpha_e(x_{t-1}, y_t) = \log_2\left(\frac{q_{c, c}}{q_{c-1, c-1}}\right). \]

\end{proof}

\section*{Supplementary Proof 2}
\stepcounter{section}

The second theorem describes the actual causes and effects for an observation of a disjunction of conjunctions (DOC) $V_t = \{Y_t\}$ that is a disjunction of $k$ conjunctions, each over $n_j$ elements, and its inputs $V_{t-1} = \{\{V_{i, j, t-1}\}_{i=1}^{n_j}\}_{j=1}^{k}$. The total number of inputs to the DOC element is $n = \sum_{j=1}^k n_j$. We consider occurrences $x_{t-1}$ that contain $c_j \leq n_j$ elements from each of the $k$ conjunctions, and the total number of elements is $|x_{t-1}| = c = \sum_{j=1}^k c_j$. To simplify notation, we further define $\bar{x}_{j, t-1} = \{v_{i, j, t-1}\}_{i = 1}^{n_j}$, an occurrence with $c_j = n_j$ and $c_{j'} = 0$ if $j' \neq j$. In other words, $\bar{x}_{j, t-1}$ is the set of elements that make up the $j^{th}$ conjunction. First, a series of lemmas are demonstrated based on the transition probabilities $q(s)$:

\[ q(s) = p(Y_t = 1|x_{t-1} = s) \]

To isolate the specific conjunctions, we define $s_j \subset x_{t-1}$ to be the state of $X_{t-1}$ within the $j^{th}$ conjunction, and $\bar{s}_{j} = \cup_{i=1}^{j} s_i \subseteq x_{t-1}$ be the state $X_{t-1}$ within the first $j$ conjunctions. For a DOC with $k$ conjunctions, we consider occurrences with $c_j$ elements from each conjunction, $X_{t-1} = \{\{x_{i, j, t-1}\}_{i = 1}^{c_j}\}_{j=1}^{k}$. In the specific case of a disjunction of two conjunctions,

\[ q(s_1, s_2) = \begin{cases} 
0 ~ &\text{if} ~ \min(s_1) = \min(s_2) = 0 \\
\frac{1}{2^{n_1 - c_1}} ~ &\text{if} ~ \min(s_1) = 1, ~ \min(s_2) = 0 \\
\frac{1}{2^{n_2 - c_2}} ~ &\text{if} ~ \min(s_1) = 0, ~ \min(s_2) = 1 \\
\frac{2^{n_1 - c_1} + 2^{n_2 - c_2} - 1}{2^{n_1 + n_2 - c_1 - c_2}} &\text{if} ~ \min(s_1) = \min(s_2) = 1, \\
\end{cases} \]
and in the case of $k > 2$ conjunctions, we define the probability recursively

\[ q(\bar{s}_{k-1}, s_{k}) =  \begin{cases} 
q(\bar{s}_{k-1}) ~ &\text{if} ~ \min(s_k) = 0 \\
q(\bar{s}_{k-1}) + \frac{(1 - q(\bar{s}_{k-1}))}{2^{n_k - c_k}} ~ &\text{if} ~ \min(s_k) = 1 \\
\end{cases} \]

The first two lemmas demonstrate the effect of adding an additional element to an occurrence. Adding an `ON' input to an occurrence $x_{t-1}$ can never decrease the probability of $\{Y_t = 1\}$, while adding an `OFF' input to an occurrence $x_{t-1}$ can never increase the probability of $\{Y_t = 1\}$.  
\begin{lemma}
\label{lemma5}
 If $\{x_{t-1} = s\} = \{x_{t-1}' = s', x_{i, j, t-1} = 1\}$, then $q(s') \leq q(s)$.
\end{lemma}

\begin{proof}
The proof is given by induction. We first consider the case where $k = 2$. Assume w.l.g. that the additional element $x_{i, j, t-1}$ is from the first conjunction ($c_1 = c_1' + 1$, $c_2 = c_2'$). If $\min(s_1') = 0$ then $q(s') = q(s)$. If $\min(s_2') = 0$ and $\min(s_1') = 1$ then

\[ \frac{q(s')}{q(s)} = \frac{2^{n_1 - (c_1' + 1)}}{2^{n_1 - c_1'}} = \frac{1}{2} < 1, \]
so $q(s') < q(s)$. Lastly, if $\min(s_1') = \min(s_2') = 1$ then

\[  \frac{q(s')}{q(s)} = \frac{2^{n_1 + n_2 - (c_1' + 1) - c_2'}(2^{n_1 - c_1'} + 2^{n_2 - c_2'} - 1)}{2^{n_1 + n_2 - c_1' - c_2'}(2^{n_1 - (c_1' + 1)} + 2^{n_2 - c_2'} - 1)} = \frac{2^{n_1 - c_1'} + 2^{n_2 - c_2'} - 1}{2^{n_1 - c_1'} + 2(2^{n_2 - c_2'} - 1)} < 1. \]
Therefore, when $k = 2$ we have that $q(s') \leq q(s)$. Next, we assume the result holds for $k-1$, $q(\bar{s}_{k-1}') \leq q(\bar{s}_{k-1})$ and demonstrate the result for general $k$. Again, assume the additional element is from the first conjunction ($c_1 = c_1' + 1, c_j = c_j'$ for $ j > 1$). If $\min(s_k) = 0$ then 

\[ \frac{q(\bar{s}_k')}{q(\bar{s}_k)} = \frac{q(\bar{s}_{k-1}')}{q(\bar{s}_{k-1})} \leq 1, \]
and if $\min(s_k) = 1$ then

\begin{eqnarray*}
\frac{q(\bar{s}_k')}{q(\bar{s}_k)} &=& \frac{q(\bar{s}_{k-1}') + (1 - q(\bar{s}_{k-1}')) / 2^{n_k - c_k}}{q(\bar{s}_{k-1}) + (1 - q(\bar{s}_{k-1})) / 2^{n_k - c_k}} \\
&=& \frac{(2^{n_k - c_k} - 1)q(\bar{s}_{k-1}') + 1}{(2^{n_k - c_k} - 1)q(\bar{s}_{k-1}) + 1} \leq 1. \\
\end{eqnarray*}
\end{proof}

\begin{lemma}
\label{lemma6}
 If $\{x_{t-1} = s\} = \{x_{t-1}' = s', x_{i, j, t-1} = 0\}$, then $q(s') \geq q(s)$.
\end{lemma}

\begin{proof}
The proof is given by induction. We first consider the case where $k = 2$. Assume w.l.g. that the additional element is from the first conjunction ($c_1 = c_1' + 1$, $c_2 = c_2'$). If $\min(s_1') = 0$ then $q(s') = q(s)$. If $\min(s_2') = 0$ and $\min(s_1') = 1$ then

\[ q(s') = \frac{1}{2^{n_1 - c_1'}} > 0 = q(s) \]
Lastly, if $\min(s_1') = \min(s_2') = 1$ then

\[  \frac{q(s')}{q(s)} = \frac{2^{n_2 - c_2'}(2^{n_1 - c_1'} + 2^{n_2 - c_2'} - 1)}{2^{n_1 + n_2 - c_1' - c_2'}} = \frac{2^{n_1 - c_1'} + 2^{n_2-c_2'} -1}{2^{n_1 - c_1'}} \geq 1 \]
Therefore, when $k = 2$ we have that $q(s') \geq q(s)$. Next, we assume the result holds for $k-1$, $q(\bar{s}_{k-1}') \geq q(\bar{s}_{k-1})$ and demonstrate the result for general $k$. Again, assume the additional element is from the first conjunction ($c_1 = c_1' + 1, c_j = c_j'$ for $ j > 1$). If $\min(s_k) = 0$ then 

\[ \frac{q(\bar{s}_k')}{q(\bar{s}_k)} = \frac{q(\bar{s}_{k-1}')}{q(\bar{s}_{k-1})} \leq 1, \]
and if $\min(s_k) = 1$ then

\begin{eqnarray*}
\frac{q(\bar{s}_k')}{q(\bar{s}_k)} &=& \frac{q(\bar{s}_{k-1}') + (1 - q(\bar{s}_{k-1}')) / 2^{n_k - c_k}}{q(\bar{s}_{k-1}) + (1 - q(\bar{s}_{k-1})) / 2^{n_k - c_k}} \\
&=& \frac{(2^{n_k - c_k} - 1)q(\bar{s}_{k-1}') + 1}{(2^{n_k - c_k} - 1)q(\bar{s}_{k-1}) + 1} \geq 1. \\
\end{eqnarray*}
\end{proof}

Next, we again consider a normalization term $Q(c)$, which is the sum of $q(s)$ over all states of the occurrence. Here we demonstrate the effect on $Q(c)$ of adding an additional element to an occurrence.
\begin{lemma}
\label{lemma7}
For an occurrence $\{X_{t-1} = x_{t-1}\}$ with $|x_{t-1}| = c > 0$, define $Q(c) = \sum_s q(s)$. Now consider adding a single element to an occurrence, $x_{t-1}' = \{x_{t-1}, x_{i, j_1, t-1}\}$, ($x_{i, j_1, t-1} \notin x_{t-1}$) such that $c'_{j_1} = c_{j_1} + 1$ and $c'_j = c_j$ for $j \neq j_1$, so that $c' = c+1$. Then $\frac{Q(c')}{Q(c)} = 2$. 
\end{lemma}

\begin{proof}
The proof is again given by induction. We first consider the case where $k = 2$, 

\begin{eqnarray*}
Q(c) &=& \sum_s q(s) \\
&=& \frac{2^{c_1} - 1}{2^{n_2 - c_2}} + \frac{2^{c_2} - 1}{2^{n_1 - c_1}} + \frac{2^{n_1 - c_1} + 2^{n_2 - c_2} - 1}{2^{n_1 + n_2 - c_1 - c_2}} \\ 
&=& \frac{2^{n_1} + 2^{n_2} - 1}{2^{n_1 + n_2 - c_1 - c_2}} \\
\end{eqnarray*}
Assume w.l.g. that the additional element to the first conjunction ($c_1' = c_1 + 1$). Then we have that 

\[ \frac{Q(c')}{Q(c)} = \frac{2^{n_1 + n_2 - c_1 - c_2}(2^{n_1} + 2^{n_2} - 1)}{2^{n_1 + n_2 - c_1' - c_2'}(2^{n_1}+2^{n_2} - 1)} = \frac{2^{n_1 + n_2 - c_1 - c_2}}{2^{n_1 + n_2 - (c_1 + 1) - c_2}} = 2 \]
Therefore, when $k = 2$ we have that $\frac{Q(c')}{Q(c)} = 2$. Next, we assume the result holds for $k-1$ and demonstrate the result for general $k$. Using the recursive relationship for $q$, we get

\begin{eqnarray*}
Q_k(c) &=& \sum_{\bar{s}_k} q(\bar{s}_k) \\
&=& \sum_{s_k} \sum_{\bar{s}_{k-1}} q(\bar{s}_{k-1}, s_k) \\
&=& (2^{c_k} - 1)\sum_{\bar{s}_{k-1}} q(\bar{s}_{k-1}) + \sum_{\bar{s}_{k-1}} \left(q(\bar{s}_{k-1}) + \frac{(1 - q(\bar{s}_{k-1}))}{2^{n_k - c_k}}\right) \\
&=& \frac{(2^{n_k} - 1)Q_{k-1}(c-c_k) + 2^{c - c_k}}{2^{n_k - c_k}}, \\
\end{eqnarray*}
Again, assume the additional element is from the first conjunction $c_1' = c_1 + 1$, for the ratio we have

\begin{eqnarray*} 
\frac{Q_k(c')}{Q_k(c)} &=& \frac{(2^{n_k} - 1)Q_{k-1}(c' - c_k') + 2^{c' - c_k'}}{(2^{n_k} - 1)Q_{k-1}(c - c_k) + 2^{c - c_k}} \\
&=&  \frac{(2^{n_k} - 1)2Q_{k-1}(c- c_k) + 2^{(c - c_k) + 1}}{(2^{n_k} - 1)Q_{k-1}(c - c_k) + 2^{c - c_k}} \\
&=& 2 \left(\frac{(2^{n_k} - 1)Q_{k-1}(x_{t-1}') + 2^{c - c_k}}{(2^{n_k} - 1)Q_{k-1}(x_{t-1}') + 2^{c - c_k}}\right) \\ 
&=& 2
\end{eqnarray*}
\end{proof}

The final two Lemmas demonstrate conditions under which the probability of $\{Y_t = 1\}$ is either strictly increasing or strictly decreasing. 
\begin{lemma}
\label{lemma8}
 If $\min(x_{t-1}) = 1$, $c_j < n_j~\forall~j$ and $x_{t-1}' \subset x_{t-1}$ then $q(s') < q(s)$.
\end{lemma}

\begin{proof}
The proof is given by induction. We first consider the case where $k = 2$. Assume w.l.g. that $x_{t-1}$ has an additional element in the first conjunction relative to $x_{t-1}'$ ($c_1 = c_1' + 1$, $c_2 = c_2'$). The result can be applied recursively for differences of more than one element. 

\begin{eqnarray*}
\frac{q(s)}{q(s')} &=& \left(\frac{2^{n_1 - c_1} + 2^{n_2-c_2} - 1}{2^{n_1 - c_1'} + 2^{n_2-c_2'} - 1}\right)\left(\frac{2^{n_1+n_2-c_1'-c_2'}}{2^{n_1+n_2-c_1-c_2}}\right) \\
&=& 2\left(\frac{2^{n_1 - c_1} + 2^{n_2-c_2} - 1}{2^{n_1 - c_1 + 1} + 2^{n_2-c_2} - 1}\right) \\
&>& 1 \quad \text{(since $c_2 < n_2$)}
\end{eqnarray*}
Therefore, when $k = 2$ we have that $q(s') < q(s)$. Next, we assume the result holds for $k-1$, $q(\bar{s}_{k-1}') < q(\bar{s}_{k-1})$ and demonstrate the result for general $k$. Again, assume that $x_{t-1}$ and $x_{t-1}'$ differ by a single element in the first conjunction ($c_1 = c_1' + 1, c_j = c_j'$ for $ j > 1$). Since $\min(s_k) = 1$, 

\begin{eqnarray*}
\frac{q(\bar{s}_k)}{q(\bar{s}_k')} &=& \frac{q(\bar{s}_{k-1}') + (1 - q(\bar{s}_{k-1}')) / 2^{n_k - c_k}}{q(\bar{s}_{k-1}) + (1 - q(\bar{s}_{k-1})) / 2^{n_k - c_k}} \\
&=& \frac{(2^{n_k - c_k} - 1)q(\bar{s}_{k-1}') + 1}{(2^{n_k - c_k} - 1)q(\bar{s}_{k-1}) + 1} \\ 
&>& 1. \\
\end{eqnarray*}
\end{proof}

\begin{lemma}
\label{lemma9}
If $\max(x_{t-1}) = 0$ , $c_j \leq 1~\forall~j$ and $x_{t-1}' \subset x_{t-1}$ then $q(s) < q(s')$.
\end{lemma}

\begin{proof}
The proof is given by induction. We first consider the case where $k = 2$. Assume w.l.g. that $x_{t-1}$ has an additional element in the first conjunction relative to $x_{t-1}'$ ($c_1 = c_1' + 1 = 1$, $c_2 = c_2'$). The result can be applied recursively for differences of more than one element. First, consider the case where $c_2 = 1$. Then we have 

\[ q(s') = \frac{1}{2^{n_2 - c_2}} > 0 = q(s). \]
Next consider the case where $c_2 = 0$:

\[ q(s') = \frac{2^{n_1} + 2^{n_2} - 1}{2^{n_1+n_2}} = \frac{1}{2^{n_2}}\left(\frac{2^{n_1}+2^{n_2} - 1}{2^{n_1}}\right) = q(s)\left(\frac{2^{n_1}+2^{n_2} - 1}{2^{n_1}}\right) > q(s). \]
Therefore, when $k = 2$ we have that $q(s) < q(s')$. Next, we assume the result holds for $k-1$, $q(\bar{s}_{k-1}) < q(\bar{s}_{k-1}')$, and demonstrate the result for general $k$. Again, assume that $x_{t-1}$ and $x_{t-1}'$ differ by a single element in the first conjunction ($c_1 = c_1' + 1, c_j = c_j'$ for $ j > 1$). Since $\min(s_k) = 0$, 

\[ \frac{q(\bar{s}_k)}{q(\bar{s}_k')} = \frac{q(\bar{s}_{k-1})}{q(\bar{s}_{k-1}')} < 1. \]

\end{proof}

Using the above Lemmas, we are now in a position to prove the actual causes and actual effects in the causal account of a single DOC and its inputs. We separately consider the case where the DOC is in the `ON' and the `OFF' state.
\begin{theorem}
\label{thm2}

Consider a dynamical causal network $G_u$ such that $V_t = \{Y_t\}$ is a DOC element that is a disjunction of $k$ conditions, each of which is a conjunction of $n_j$ inputs, and $V_{t-1} = \{\{V_{i, j, t-1}\}_{i = 1}^{n_j}\}_{j=1}^{k}$ is the set of its $n = \sum_j n_j$ inputs. For a transition $v_{t-1} \prec v_{t}$, the following holds: 

\begin{enumerate}
\item If $y_t = 1$,

\begin{enumerate}

\item The actual cause of $\{Y_t = 1\}$ is an occurrence $\{X_{t-1} = x_{t-1}\}$ where $x_{t-1} = \{x_{i, j, t-1}\}_{i=1}^{n_j} \subseteq v_{t-1}$ such that $\min(x_{t-1}) = 1$.

\item The actual effect of $\{X_{t-1} = x_{t-1}\}$ is $\{Y_t = 1\}$ if $\min(x_{t-1}) = 1$ and $|x_{t-1}| = c_j = n_j$; otherwise $x_{t-1}$ is reducible.

\end{enumerate}

\item If $y_t = 0$,

\begin{enumerate}

\item The actual cause of $\{Y_t = 0\}$ is an occurrence $x_{t-1} \subseteq v_{t-1}$ such that $\max(x_{t-1}) = 0$ and $c_j = 1 ~\forall~j$. 

\item If $\max(x_{t-1}) = 0$ and $c_j \leq 1~\forall~j$ then the actual effect of $\{X_{t-1} = x_{t-1}\}$ is $\{Y_t = 0\}$; otherwise $x_{t-1}$ is reducible. 

\end{enumerate}

\end{enumerate}
\end{theorem}

\begin{proof}
\label{proof2}

\textbf{Part 1a:} The actual cause of $\{Y_t = 1\}$. For an occurrence $\{X_{t-1} = x_{t-1}\}$, the probability of $x_{t-1}$ in the cause repertoire of $y_t$ is 

\[ \pi(x_{t-1}\mid y_t) = \frac{q(s)}{Q(c)}. \]
Since $Y_t$ is a first-order occurrence, there is only one possible partition, and the causal strength of a potential link is thus

\[ \alpha_c(x_{t-1}, y_t) = \log_2\left(\frac{\pi(x_{t-1}\mid y_t)}{\pi(x_{t-1})}\right) = \log_2\left(\frac{2^cq(s)}{Q(c)}\right) = \log_2\left(Q_1q(s)\right), \]
where $Q_1 = \frac{2^c}{Q(c)} ~\forall~c$ (Lemma \ref{lemma7}). If we then consider adding a single element to the occurrence $x_{t-1}' = \{x_{t-1}, x_{i, j, t-1}'\}$ ($x_{i, j, t-1}' \notin x_{t-1}$) then the difference in causal strength is

\[ \alpha_c(x_{t-1}, y_t) - \alpha_c(x_{t-1}', y_t) = \log_2\left(\frac{Q_1\,q(s)}{Q_1\,q(s')}\right) = \log_2\left(\frac{q(s)}{q(s')} \right) \]
Combining the above with Lemma \ref{lemma6}, adding an element $x_{i, j, t-1} = 0$ to an occurrence cannot increase the causal strength, and thus occurrences that include elements in state `OFF' cannot be the actual cause of $y_t$. By Lemma \ref{lemma5}, adding an element $x_{i, j, t-1} = 1$ to an occurrence cannot decrease the causal strength. Furthermore, if $c_j = n_j$ and $\min(\bar{x}_{j, t-1}) = 1$, then $q(s) = 1$ and 

\[\alpha_c(y_t, x_{t-1}) = \log_2\left(Q_1q(s)\right) = \log_2(Q_1), \]
independent of the number of elements in the occurrence from other conjunctions $c_{j'}$ and their states $s_{j'}$ ($j' \neq j$). Since the value $Q_1$ does not depend on the specific value of $j$, it must be the case that this is the maximum value of causal strength, $\alpha^{\m}(y_t)$. Furthermore, if $c_j < n_j ~\forall~ j$ then 

\[\alpha_c(y_t, x_{t-1}) = \log_2\left(Q_1q(s)\right) < \log_2\left(Q_1\right). \]
Therefore, the maximum value of causal strength is 
\[\log_2\left(Q_1\right), \]
and an occurrence $x_{t-1}$ achieves this value (satisfying condition (1) of being an actual cause) if and only if there exists $j$ such that $c_j = n_j$ and $\min (\bar{x}_{j, t-1}) = 1$, \textit{i.e.} the occurrence includes a conjunction whose elements are all `ON'. Consider an occurrence that satisfies condition (1), such that there exists $j_1$ with $c_{j_1} = n_{j_1}$. If there exists $j_2 \neq j_1$ such that $c_{j_2} > 0$, then we can define a subset $x'_{t-1} \subset x_{t-1}$ with $c'_{j_1} = n_{j_1}$ and $c'_{j_2} = 0$ that also satisfies condition (1), and thus $x_{t-1}$ does not satisfy condition (2). Finally, if no such $j_2$ exists ($x_{t-1} = \bar{x}_{j, t-1}$) then any subset $x'_{t-1} \subset x_{t-1}$ has $c_j < n_j \, \forall j$ and does not satisfy condition (1), so $x_{t-1}$ satisfies condition (2). Therefore, we have that the actual cause of $y_t$ is an occurrence $x_{t-1} = \bar{x}_{j, t-1}$ such that $\min x_{t-1} = 1$,
\[x^*(y_t) = \{\bar{x}_{j, t-1} \subset v_{t-1} \mid \min \bar{x}_{j, t-1} = 1\}. \]

\textbf{Part 1b:} Actual effect of $x_{t-1}$ when $y_t = 1$. Again, consider occurrences $X_{t-1} = x_{t-1}$ with $c_j$ elements from each of the $k$ conjunctions. The effect repertoire of a DOC with $k$ conjunctions over such occurrences is  

\[ \pi(y_t\mid x_{t-1} = s) = q(s) \]
Since there is only one element in $v_t$, the only question is whether or not $x_{t-1}$ is reducible. If it is reducible, it has no actual effect, otherwise its actual effect must be $y_t$. First, if there exists $x \in x_{t-1}$ with $x = 0$ then we can define $x_{t-1}'$ such that $x_{t-1} = \{x_{t-1}', x\}$, and a partition $\psi = \left\{ \{x_{t-1}', y_t\}, \{x, \varnothing\}\right\}$, \textit{i.e.} cutting away $x$, such that

\[ \pi(y_t\mid x_{t-1})_\psi = \pi(y_t \mid x_{t-1}') \times \pi(\varnothing \mid x) = \pi(y_t \mid x_{t-1}') = q(s'), \]
By Lemma \ref{lemma6}, $q(s') \geq q(s)$, and thus 

\[ \alpha_e(x_{t-1}, y_t) \leq \log_2\left(\frac{\pi(y_t  \mid x_{t-1})}{\pi(y_t \mid x_{t-1})_\psi}\right) = \log_2\left(\frac{q(s)}{q(s')}\right) \leq 0, \]
so $x_{t-1}$ is reducible. Next we consider the case where $\min(x_{t-1}) = 1$, but there exists $j_1, j_2$ such that $c_{j_1} = n_{j_1}$ and $c_{j_2} > 0$. We define $x_{t-1}' = \bar{x}_{j_1, t-1}$ and a partition $\psi = \left\{ \{x_{t-1}', y_t\}, (x_{t-1} \setminus x_{t-1}'), \varnothing\}\right\}$, such that 

\[ \pi(y_t \mid x_{t-1})_\psi = \pi(y_t \mid x_{t-1}') \times \pi(\varnothing \mid (x_{t-1} - x_{t-1}')) = \pi(y_t \mid x_{t-1}') = q(s'), \]
and thus 

\[ \alpha_e(x_{t-1}, y_t) \leq \log_2\left(\frac{\pi(y_t \mid x_{t-1})}{\pi(y_t \mid x_{t-1})_\psi}\right) = \log_2\left(\frac{q(s)}{q(s')}\right) = \log_2\left(\frac{1}{1}\right) = 0, \]
so $x_{t-1}$ is again reducible. We now split the irreducible occurrences into two cases. First, we consider $\min(x_{t-1}) = 1$ and all $c_j < n_j$. All possible partitions of the pair of occurrences can be formulated as $\psi = \left\{ \{x_{t-1}', y_t\}, \{ (x_{t-1} \setminus x_{t-1}'), \varnothing\}\right\}$ (where $x_{t-1}' \subset x_{t-1}$), such that 

\[ \pi(y_t \mid x_{t-1})_\psi = \pi(y_t \mid x_{t-1}')) \times \pi(\varnothing \mid (x_{t-1} - x_{t-1}') = \pi(y_t \mid x_{t-1}') = q(s'), \]
and by Lemma \ref{lemma8},

\[ \alpha_e(x_{t-1}, y_t) = \min_{\psi} \left(\log_2\left(\frac{\pi(y_t \mid x_{t-1})}{\pi(y_t \mid x_{t-1})_\psi}\right)\right) = \min_{\psi} \left(\log_2\left(\frac{q(s)}{q(s')}\right)\right) > 0. \]
So $x_{t-1}$ is irreducible, and its actual effect is $\{Y_1 = 1\}$. Next we consider occurrences such that $\min(x_{t-1}) = 1$, $c_{j_1} = n_{j_1}$, and $c_j = 0$ for $j \neq j_1$ (\textit{i.e.} $x_{t-1} = \bar{x}_{j_1, t-1}$). All possible partitions of the pair of occurrences can be formulated as $\psi = \left\{ \{x_{t-1}', y_t\}, \{(x_{t-1} - x_{t-1}', \varnothing\}\right\}$ (where $x_{t-1}' \subset x_{t-1}$), such that 

\[ \pi(y_t \mid x_{t-1})_\psi = \pi(y_t \mid x_{t-1}') \times \pi(\varnothing \mid (x_{t-1} - x_{t-1}') = \pi(y_t \mid x_{t-1}') = q(s'), \]

\[ \alpha_e(x_{t-1}, y_t) \leq \log_2\left(\frac{\pi(y_t \mid x_{t-1})}{\pi(y_t \mid x_{t-1})_\psi}\right) = \log_2\left(\frac{q(s)}{q(s')}\right) = \log_2\left(\frac{1}{q(s')}\right) > 0, \]
and $x_{t-1}$ is again irreducible with actual effect $\{Y_t = 1\}$. \\ 

\vspace{0.2in}

\textbf{Part 2a:} The actual cause of $\{Y_t = 0\}$. For an occurrence $\{X_{t-1} = x_{t-1}\}$ the cause repertoire of $y_t$ is 

\[ \pi(x_{t-1} \mid y_t) = \frac{1 - q(s)}{2^c - Q(c)}. \]
Since $Y_t$ is a first-order occurrence, there is only one possible partition, and the causal strength of a potential link is thus

\[ \alpha_c(x_{t-1}, y_t) = \log_2\left(\frac{\pi(x_{t-1} \mid y_t)}{\pi(x_{t-1})}\right) = \log_2 \left(\frac{2^c(1 - q(s))}{2^c - Q(c)}\right) = \log_2\left(Q_0\,q(s)\right), \]
where $Q_0 = \frac{2^c}{2^c - Q(c)}~\forall~c$ (Lemma \ref{lemma7}). If we then consider adding a single element to the occurrence $x_{t-1}' = \{x_{t-1}, x_{i, j, t-1}'\}$ ($x_{i, j, t-1}' \notin x_{t-1}$), then the difference in causal strength is

\[ \alpha_c(x_{t-1}, y_t) - \alpha_c(x_{t-1}', y_t) = \log_2\left(\frac{Q_0(1 - q(s))}{Q_0(1 - q(s'))}\right) = \log_2\left(\frac{1-q(s)}{1-q(s')}\right) \]
By Lemma \ref{lemma6}, adding an element $x = 1$ to an occurrence cannot increase the causal strength, and thus occurrences that include elements in state `ON' cannot be the actual cause of $y_t$. By Lemma \ref{lemma5}, adding an element $x = 0$ to an occurrence cannot decrease the causal strength. If $c_j > 0 ~\forall~ j$ and $\max(x_{t-1}) = 0$, then

\[\alpha_c(y_t, x_{t-1}) = \log_2\left(Q_0(1-q(s))\right) = \log_2\left(Q_0\right), \]
independent of the actual values of $c_j$. Since this holds for any set of $c_j$ that satisfies the conditions, it must be the case that this value is $\alpha^{\m}(y_t)$. Furthermore, if there exists $j$ such that $c_j = 0$ then 

\[\alpha_c(y_t, x_{t-1}) = \log_2\left(Q_0(1- q(s))\right) < \log_2\left(Q_0\right). \]
Therefore, the maximum value of causal strength is 
\[\log_2\left(Q_0\right), \]
and an occurrence $x_{t-1}$ achieves this value (satisfying condition (1) of being an actual cause) if and only if $c_j > 0\,\forall\,j$ and $\max(x_{t-1}) = 0$, \textit{i.e.} the occurrence contains elements from every conjunction, and only elements whose state is `OFF'. 

Consider an occurrence $x_{t-1}$ that satisfies condition (1). If there exists $j_1$ such that $c_{j_1} > 1$, then we can define a subset $x'_{t-1} \subset x_{t-1}$ with $c'_{j_1} = 1$ that also satisfies condition (1), and thus $x_{t-1}$ does not satisfy condition (2). Finally, if $c_j = 1\,\forall\, j$ then for any subset $x'_{t-1} \subset x_{t-1}$ there exists $j$ such that  $c'_j = 0$, so $x'_{t-1}$ does not satisfy condition (1), and thus $x_{t-1}$ satisfies condition (2). Therefore, we have that the actual cause of $y_t$ is an occurrence $x_{t-1}$ such that $\max x_{t-1} = 0$ and $c_j = 1\,\forall\, j$,
\[x^*(y_t) = \{x_{t-1} \subseteq v_{t-1} \mid \max(x_{t-1}) = 0 \text{ and } c_j = 1\,\forall\, j \}. \] 

\textbf{Part 2b:} Actual effect of $x_{t-1}$ when $y_t = 0$. Again, consider occurrences $X_{t-1} = x_{t-1}$ with $c_j$ elements from each of $k$ conjunctions. The probability of $y_t$ in the effect repertoire of $x_{t-1}$ is   

\[ \pi(y_t \mid x_{t-1} = s) = 1 - q(s). \]
Since there is only one element in $v_t$, the only question is whether or not $x_{t-1}$ is reducible. If it is reducible, it has no actual effect, otherwise its actual effect must be $y_t$. First, if there exists $x_{i, j, t-1} \in x_{t-1}$ such that $x_{i, j, t-1} = 1$ then we can define $x_{t-1}'$ such that $x_{t-1} = \{x_{t-1}', x_{i, j, t-1}\}$ and a partition $\psi = \left\{ \{x_{t-1}', y_t\}, \{x_{i, j, t-1}, \varnothing\}\right\}$ such that

\[ \pi(y_t \mid x_{t-1})_\psi = \pi(y_t \mid x_{t-1}') \times \pi(\varnothing \mid x_i, j, t-1) = \pi(y_t \mid x_{t-1}') = 1 - q(s'). \]
By Lemma \ref{lemma5}, we have $1 - q(s) \leq 1 - q(s')$, and thus 

\[ \alpha_e(x_{t-1}, y_t) \leq \log_2\left(\frac{\pi(y_t \mid x_{t-1})}{\pi(y_t \mid x_{t-1})_\psi}\right) = \log_2\left(\frac{1 - q(s)}{1 - q(s')}\right) \leq 0, \]
so $x_{t-1}$ is reducible. Next we consider the case where $\max(x_{t-1}) = 0$, but there exists $j$ such that $c_{j} > 1$. We define $x_{t-1}'$ with $c'_j = 1 \, \forall \, j$ such that $x_{t-1} = \{x_{t-1}', x_{i, j, t-1}\}$, and a partition $\psi = \left\{ \{x_{t-1}', y_t\}, \{x_{i, j, t-1}, \varnothing\}\right\}$, such that  

\[ \pi(y_t \mid x_{t-1})_\psi = \pi(y_t \mid x_{t-1}') \times \pi(\varnothing \mid x_{i, j, t-1}) = \pi(y_t \mid x_{t-1}') = q(s') = 1 \]
and

\[ \alpha_e(x_{t-1}, y_t) \leq \log_2\left(\frac{\pi(y_t \mid x_{t-1})}{\pi(y_t \mid x_{t-1})_\psi}\right) = \log_2\left(\frac{1 - q(s)}{1 - q(s')}\right) = \log_2\left(\frac{1}{1}\right) = 0, \]
so $x_{t-1}$ is again reducible. Finally, we show that occurrences $x_{t-1}$ are irreducible if $\max(x_{t-1}) = 0$ and all $c_j \leq 1$. All possible partitions of the pair of occurrences can be formulated as $\psi = \left\{ \{x_{t-1}', y_t\}, \{ (x_{t-1} \setminus x_{t-1}'), \varnothing\}\right\}$ (where $x_{t-1}' \subset x_{t-1}$), such that $c_j' \leq c_j~\forall~j$, and $c' < c$.  Then  

\[ \pi(y_t \mid x_{t-1})_\psi = \pi(y_t \mid x_{t-1}') \times \pi(\varnothing \mid (x_{t-1} - x_{t-1}')) = \pi(y_t \mid x_{t-1}') = 1 - q(s'), \]
and by Lemma \ref{lemma9},

\[ \alpha_e(x_{t-1}, y_t) = \min_{\psi} \left(\log_2\left(\frac{\pi(y_t \mid x_{t-1})}{\pi(y_t \mid x_{t-1})_\psi}\right)\right) = \min_{\psi} \left(\log_2\left(\frac{1 - q(s)}{1 - q(s')}\right)\right) > 0. \]
So $\{X_{t-1} = x_{t-1}\}$ is irreducible, and its actual effect is $\{Y_1 = 1\}$.

\end{proof}